\newif\ifJOURNAL
\JOURNALfalse
\newif\ifWP
\WPfalse
\newif\ifarXiv
\arXivfalse

\arXivtrue

\newif\ifnotJOURNAL	% derivative
\notJOURNALtrue
\ifJOURNAL\notJOURNALfalse\fi

\ifJOURNAL
  \documentclass{article}
  \usepackage{amsmath,amssymb,amsfonts,latexsym,epsfig,stmaryrd}
  \usepackage[sort&compress,numbers,comma]{natbib}
\fi

\ifWP
  \documentclass[toc]{kpnsarticle}
  \usepackage{amsmath,amssymb,amsfonts,latexsym,epsfig,stmaryrd}
  \usepackage[sort&compress,numbers,comma]{natbib}

  \input{OT2enc.def}
  
\fi

\ifarXiv
  \documentclass{article}
  \usepackage{amsmath,amssymb,amsfonts,latexsym,epsfig,stmaryrd}
  \usepackage[sort&compress,numbers,comma]{natbib}
\fi

\newcommand{\K}{\mathcal{K}}

\newcommand{\st}{\mathop{|}}
\newcommand{\bbbn}{\mathbb{N}}
\newcommand{\bbbr}{\mathbb{R}}

\newcommand{\UP}{\overline{{\rm P}}}

\newcommand{\Prob}{{\rm Pr}}

\newcommand{\givn}{\mathbin{|}}

\newtheorem{lemma}{Lemma}
\newtheorem{proposition}{Proposition}
\newtheorem{inproposition}{Informal Proposition}

\newtheorem{Remark}{Remark}

\newenvironment{proof}
  {\trivlist \item[\hskip\labelsep\textbf{Proof}]}
  {\endtrivlist}

\newcommand{\boxforqed}{\rule{.3em}{1.5ex}}
\newcommand{\qedtext}{\unskip\nobreak\hfil
  \penalty50\hskip1em\null\nobreak\hfil\boxforqed
  \parfillskip=0pt\finalhyphendemerits=0\endgraf}
\newcommand{\qedmath}{\eqno\boxforqed}

\newenvironment{Equation*}
  {$$\begin{array}{c}\displaystyle}
  {\end{array}$$}
\newenvironment{EquationQED*}
  {$$\begin{array}{c}\displaystyle}
  {\end{array}\qedmath$$}
\newlength{\IndentI}
\newlength{\IndentII}
\newlength{\IndentIII}
\newlength{\WidthI}
\newlength{\WidthII}
\newlength{\WidthIII}
\ifJOURNAL
  \title{A Tutorial on Conformal Prediction}
  \author{Glenn Shafer and Vladimir Vovk\\
  \texttt{gshafer{\rm@}rutgers.edu} \qquad
  \texttt{vovk{\rm@}cs.rhul.ac.uk}\\
  \texttt{http://glennshafer.com} \qquad
  \phantom{\hspace{-1cm}}\texttt{http://vovk.net}}
\fi

\ifWP
  \title{A tutorial on conformal prediction}
  \author{Glenn Shafer and Vladimir Vovk}
  
  \twodatestrue
  
\fi

\ifarXiv
  \title{A Tutorial on Conformal Prediction}
  \author{Glenn Shafer \qquad and \qquad Vladimir Vovk\\
  \texttt{gshafer{\rm@}rutgers.edu} \qquad
  \texttt{vovk{\rm@}cs.rhul.ac.uk}\\
  \texttt{http://glennshafer.com} \qquad
  \texttt{http://vovk.net}\qquad~}
\fi

\begin{document}
\maketitle

\begin{abstract}
   Conformal prediction uses past experience
   to determine precise levels of confidence in new predictions.
   Given an error probability $\epsilon$, together with 
   a method that makes a prediction $\hat{y}$ of a label $y$, 
   it produces a set of labels, typically containing $\hat{y}$,
   that also contains $y$ with probability $1-\epsilon$.
   Conformal prediction can be applied to any method for producing
   $\hat{y}$: a nearest-neighbor method, a support-vector machine,
   ridge regression, etc.
      
   Conformal prediction is designed for an on-line setting
   in which labels are predicted successively, each one being revealed
   before the next is predicted.  
   The most novel and valuable feature of conformal prediction is that
   if the successive examples are sampled 
   independently from the same distribution, then the successive
   predictions will be right $1-\epsilon$ of the time, even
   though they are based on an accumulating dataset rather than on 
   independent datasets.

   In addition to the model under which successive examples are 
   sampled independently, other 
   on-line compression models can also use conformal prediction.
   The widely used Gaussian linear model is one of these.

   This tutorial presents a self-contained account of the theory of 
   conformal prediction and works through several numerical examples.
   A more comprehensive treatment of the topic is provided in 
   \textit{Algorithmic Learning in a Random World}, by Vladimir
   Vovk, Alex Gammerman, and Glenn Shafer (Springer, 2005).
   \ifWP
     \newpage
   \fi
\end{abstract}

\ifJOURNAL
  \pagebreak

  \tableofcontents

  \pagebreak
  \normalsize
\fi

\section{Introduction}

How good is your prediction $\hat{y}$?  If you are predicting the label $y$ of a new object,
how confident are you that $y = \hat{y}$?
If the label $y$ is a number, how close do you think it is to $\hat{y}$?
In machine learning, these questions are usually answered in a fairly rough way 
from past experience.  We expect new predictions to fare about as well
as past predictions.

Conformal prediction uses past experience to determine precise 
levels of confidence in 
predictions.  Given a method for making a 
prediction $\hat{y}$, conformal prediction produces
a \textit{$95\%$ prediction region}---a set 
$\Gamma^{0.05}$ that contains $y$ with probability at least $95\%$.  
Typically $\Gamma^{0.05}$ also contains the prediction $\hat{y}$.
We call $\hat{y}$ the \textit{point prediction}, and we call
$\Gamma^{0.05}$ the \textit{region prediction}.  In the case of 
regression, where $y$ is a number, $\Gamma^{0.05}$ is typically an interval 
around $\hat{y}$.  In the case of classification, where $y$ has 
a limited number of possible values, $\Gamma^{0.05}$ may consist of a few
of these values or, in the ideal case, just one.

Conformal prediction can be used with any method of point prediction for
classification or regression, including support-vector 
machines, decision trees, boosting, neural networks,
and Bayesian prediction.  Starting from the method for 
point prediction, we construct a \textit{nonconformity
measure}, which measures how unusual an example looks relative
to previous examples, and the \textit{conformal algorithm}
turns this nonconformity measure into prediction regions.

Given a nonconformity measure, the conformal algorithm produces a prediction
region $\Gamma^{\epsilon}$ for every probability of error
$\epsilon$.  The region $\Gamma^{\epsilon}$ is a $(1-\epsilon)$-\textit{prediction 
region}; it contains $y$ with probability at least $1-\epsilon$.
The regions
for different $\epsilon$ are nested: 
when $\epsilon_1 \geq \epsilon_2$, so that 
$1-\epsilon_1$ is a lower level of confidence
than $1-\epsilon_2$, we have
$\Gamma^{\epsilon_1} \subseteq \Gamma^{\epsilon_2}$.
If $\Gamma^{\epsilon}$ contains only a single label (the ideal outcome in the 
case of classification), we may ask how small $\epsilon$ can be made
before we must enlarge $\Gamma^{\epsilon}$ by adding a second label;
the corresponding value of $1-\epsilon$
is the confidence we assert in the predicted label.

As we explain in~\S\ref{sec:works}, the conformal algorithm 
is designed for an on-line setting,
in which we predict the labels of objects 
successively, seeing each label after we have 
predicted it and before we predict the next one.  Our 
prediction $\hat{y}_n$ of the $n$th label $y_n$ may use
observed features $x_n$ of the $n$th object and the preceding
examples $(x_1,y_1),\dots,(x_{n-1},y_{n-1})$.
The size of the prediction region $\Gamma^{\epsilon}$
may also depend on these details.
Readers most interested in implementing the conformal
algorithm may wish to turn directly to 
the elementary examples in \S\ref{subsec:oldalone} 
and~\S\ref{subsec:features} and then turn
back to the earlier more general material as needed.

As we explain in \S\ref{sec:history}, the
on-line picture leads to a new concept of validity for 
prediction with confidence.  Classically, a method for finding
$95\%$ prediction regions
was considered valid if it had a $95\%$ probability of containing 
the label predicted, because by the law of the large numbers
it would then be correct $95\%$ of the time when repeatedly
applied to independent datasets.  But in the on-line picture,
we repeatedly apply a method not to independent datasets but to
an accumulating dataset.  After using $(x_1,y_1),\dots,(x_{n-1},y_{n-1})$
and $x_n$ to predict $y_n$, we use 
$(x_1,y_1),\dots,(x_{n-1},y_{n-1}),(x_n,y_n)$
and $x_{n+1}$ to predict $y_{n+1}$, and so on.  For a $95\%$ on-line
method to be valid, $95\%$ of these predictions must be correct.
Under minimal assumptions,
conformal prediction is valid in this new and powerful sense.

One setting where conformal prediction is valid in
the new on-line sense is the one in which the 
examples $(x_i,y_i)$ are sampled independently from a constant 
population---i.e., from a fixed but unknown probability 
distribution $Q$.  It is also valid under
the slightly weaker assumption that the examples
are probabilistically \textit{exchangeable}
(see \S\ref{sec:exchangeable}) and under other
on-line compression models, including the
widely used Gaussian linear model (see \S\ref{sec:compression}).
The validity of conformal prediction under these models is 
demonstrated in Appendix~\ref{app:proof}.

In addition to the validity of a method for producing $95\%$
prediction regions, we are also interested in its efficiency.
It is efficient if the prediction region is usually 
relatively small and therefore informative.
In classification, we would like to see a 95\% prediction region 
so small that it contains
only the single predicted label $\hat{y}_n$.  
In regression, we would like to see a very narrow interval around
the predicted number $\hat{y}_n$.  

The claim of 95\% confidence for a 95\% conformal prediction region
is valid under exchangeability, no matter what the probability
distribution $Q$ the examples follow and no matter
what nonconformity measure is used to construct the conformal
prediction region.  But the efficiency of conformal prediction 
will depend on $Q$ and
the nonconformity measure.  If we think we know $Q$, we may choose
a nonconformity measure that will be efficient if we are right.
If we have prior probabilities for $Q$, 
we may use these prior probabilities
to construct a point predictor $\hat{y}_n$ and a nonconformity measure.  
In the regression case, we might use as $\hat{y}_n$
the mean of the posterior distribution for $y_n$ given the first $n-1$ examples and $x_n$;
in the classification case, we might use the label with the greatest 
posterior probability.  
This strategy of first guaranteeing validity under a relatively weak assumption
and then seeking efficiency under stronger assumptions conforms to advice
long given by John Tukey and others \cite{tukey:1986,small/etal:2006}.  

Conformal prediction is studied in detail
in \textit{Algorithmic Learning in a Random
World}, by Vovk, Gammerman, and Shafer \cite{vovk/gammerman/shafer:2005}.
A recent exposition by Gammerman and Vovk \cite{gammerman/vovk:2007}
emphasizes connections with the theory of randomness, Bayesian methods, and
induction.  In this article we emphasize 
the on-line concept of validity, the meaning of exchangeability,
and the generalization to other on-line compression models.
We leave aside many important topics 
that are treated in \textit{Algorithmic Learning in a Random
World}, including extensions beyond the on-line picture.

\section{Valid prediction regions}\label{sec:history}

Our concept of validity is consistent with a tradition that can be traced back
to Jerzy Neyman's introduction of confidence intervals for parameters
in 1937 \cite{neyman:1937}
and even to work by Laplace and others in 
the late 18th century.  But the shift of emphasis to prediction
(from estimation of parameters) and to the on-line setting 
(where our prediction rule is repeatedly updated) involves some rearrangement 
of the furniture.

The most important novelty in conformal prediction is that its successive
errors are probabilistically independent.
This allows us to interpret ``being right 95\% of the time''
in an unusually direct way.
In~\S\ref{subsec:normaldistribution}, 
we illustrate this point with a well-worn example, 
normally distributed random variables.

In~\S\ref{subsec:less}, we contrast confidence 
with full-fledged conditional probability.  This contrast has been the topic of endless
debate between those who find confidence methods informative
(classical statisticians)
and those who insist that full-fledged probabilities 
based on all one's information are
always preferable, even if the only available probabilities are
very subjective (Bayesians).  Because the debate usually focuses
on estimating parameters rather than predicting future observations, 
and because some readers may be unaware of the debate,
we take the time to explain that we find the concept of confidence 
useful for prediction in spite of its limitations.

\subsection{An example of valid on-line prediction}\label{subsec:normaldistribution}

A 95\% prediction region is \textit{valid} if it
contains the truth 95\% of the time.  
To make this more precise, we must specify the set of
repetitions envisioned. 
In the on-line picture, these are successive predictions
based on accumulating information.  We make one prediction after another,
always knowing the outcome of the preceding predictions.

To make clear what validity means and how it can be 
obtained in this on-line picture, we consider prediction under
an assumption often made in a first course in statistics:
\begin{quotation}
\noindent
   Random variables $z_1,z_2,\dots$ are independently
   drawn from a normal
   distribution with unknown mean and variance.
\end{quotation}
Prediction under this assumption was discussed in 1935 by 
R.~A.\ Fisher, who explained how to give a $95\%$ prediction
interval for $z_n$ based on $z_1,\dots,z_{n-1}$
that is valid in our sense.  We will state Fisher's prediction rule, 
illustrate its application to data, and explain why it is valid
in the on-line setting. 

As we will see, the predictions given by Fisher's rule are too 
weak to be interesting from a modern machine-learning perspective.
This is not surprising, because we are predicting
$z_n$ based on old examples $z_1,\dots,z_{n-1}$ alone.  
In general, more precise prediction is possible only in the 
more favorable but more complicated set-up where we
know some features $x_n$ of the new example and can use both 
$x_n$ and the old examples to predict some other feature $y_n$.
But the simplicity of the set-up where we predict $z_n$ from
$z_1,\dots,z_{n-1}$ alone will help us make the logic of valid 
prediction clear.

\subsubsection{Fisher's prediction interval}\label{subsubsec:fisherregion}

Suppose we observe the $z_i$ in sequence.
After observing $z_1$ and $z_2$, we start predicting;
for $n=3,4,\dots$, we predict
$z_n$ after having seen $z_1,\dots,z_{n-1}$.  
The natural point predictor for $z_n$ is the average so far:
$$
      \overline{z}_{n-1} := \frac{1}{n-1} \sum_{i=1}^{n-1} z_i,
$$
but we want to give an interval that will contain $z_n$ 95\% of the time.
How can we do this?  Here is Fisher's answer \cite{fisher:1935}:
\begin{enumerate}
  \item
      In addition to calculating the average $\overline{z}_{n-1}$, 
      calculate
$$
      s_{n-1}^2 := \frac{1}{n-2} \sum_{i=1}^{n-1} (z_i - \overline{z}_{n-1})^2,
$$
      which is sometimes called the sample variance.  We can usually assume that 
      it is non-zero.
  \item
      In a table of percentiles for $t$-distributions, find $t^{0.025}_{n-2}$, the point
      that the $t$-distribution with $n-2$ degrees of freedom exceeds exactly $2.5\%$ of the time.
  \item
      Predict that $z_n$ will be in the interval
\begin{equation}\label{eq:fisherinterval}
         \overline{z}_{n-1}  \pm t^{0.025}_{n-2} \; s_{n-1} \; \sqrt{\frac{n}{n-1}}.
\end{equation}
\end{enumerate}
Fisher based this procedure on the fact that
\begin{equation}\label{eq:tratio}
       \frac{z_n - \overline{z}_{n-1}}{s_{n-1}} \sqrt{\frac{n-1}{n}}
\end{equation}
has the $t$-distribution with $n-2$ degrees of freedom, which is 
symmetric about $0$.  This implies that~(\ref{eq:fisherinterval}) 
will contain $z_n$ with probability 95\% regardless of the values of $\mu$ 
and $\sigma^2$.

\subsubsection{A numerical example}

We can illustrate~(\ref{eq:fisherinterval}) using some numbers generated
in 1900 by the students of Emanuel Czuber (1851--1925).  These numbers are 
integers, but they theoretically have a binomial distribution and are therefore
approximately normally distributed.%
\footnote{Czuber's students randomly drew balls from an urn containing six
balls, numbered $1$ to $6$.  Each time they drew a ball, they noted its label
and put it back in the urn.  After each $100$ draws, they recorded the number
of times that the ball labeled with a $1$ was drawn
(\cite{czuber:1914}, pp.~329--335).  This should have a binomial 
distribution with parameters $100$ and $1/6$, and it is therefore approximately
normal with mean $100/6 = 16.67$ and standard deviation $\sqrt{500/36}=3.79$.}

Here are Czuber's first $19$ numbers,
$z_1,\dots,z_{19}$:
\begin{equation}\label{eq:czuberdata}
   17, 20, 10, 17, 12, 15, 19, 22, 17, 19, 14, 22, 18, 17, 13, 12, 18, 15, 17.
\end{equation}
From them, we calculate
$$
    \overline{z}_{19} = 16.53,   \qquad \qquad   s_{19} = 3.32.
$$
The upper $2.5\%$ point for the $t$-distribution with $18$ degrees
of freedom, $t^{0.025}_{18}$, is $2.101$.
So the prediction interval~(\ref{eq:fisherinterval}) 
for $z_{20}$ comes out to $[9.55,23.51]$.

Taking into account our knowledge that $z_{20}$ will be an 
integer, we can say that the $95\%$ prediction is 
that $z_{20}$ will be an integer
between $10$ and $23$, inclusive.  This prediction is correct; 
$z_{20}$ is $16$.

\subsubsection{On-line validity}\label{subsubsec:validity}

Fisher did not have the on-line picture in mind.  He probably
had in mind a picture where the formula~(\ref{eq:fisherinterval}) is used
repeatedly but in entirely separate problems.  For example, we might
conduct many separate experiments that each consist 
of drawing $100$ random numbers from a
normal distribution and then predicting a 
$101$st draw using~(\ref{eq:fisherinterval}).
Each experiment might involve a different normal distribution
(a different mean and variance),
but provided the experiments are independent from each other, the law of 
large numbers will apply.  Each time the probability is $95\%$
that $z_{101}$ will be in the interval, and so this event will happen
approximately $95\%$ of the time.

The on-line story may seem more complicated, because the 
experiment involved in predicting $z_{101}$
from $z_1,\dots,z_{100}$ is not entirely
independent of the experiment involved in predicting, say,
$z_{105}$ from $z_1,\dots,z_{104}$.  The $101$ random numbers 
involved in the first experiment are all also involved in the second.
But as a master of the analytical geometry of the 
normal distribution \cite{fisher:1925applications,efron:1969}, 
Fisher would have noticed, had he thought 
about it, that this overlap does not actually matter.
As we show in Appendix~\ref{subsec:fisherind}, the events
\begin{equation}\label{eq:hitinequality}
         \overline{z}_{n-1} - t^{0.025}_{n-2} \; s_{n-1} \; \sqrt{\frac{n}{n-1}}
         \leq z_n \leq
         \overline{z}_{n-1} + t^{0.025}_{n-2} \; s_{n-1} \; \sqrt{\frac{n}{n-1}}
\end{equation}
for successive $n$ are probabilistically independent in spite of the overlap.
Because of this independence, the law of large numbers again applies.
Knowing each event has probability $95\%$, we can conclude
that approximately $95\%$ of them will happen.
We call the events~(\ref{eq:hitinequality}) \textit{hits}.  

The prediction interval~(\ref{eq:fisherinterval})
generalizes to linear regression with normally distributed
errors, and on-line hits remain independent in this general setting.
Even though formulas for these linear-regression prediction intervals appear
in textbooks, the independence of their on-line hits was not noted
prior to our work \cite{vovk/gammerman/shafer:2005}.
Like Fisher, the textbook authors 
did not have the on-line setting in mind.  They imagined
just one prediction being made in each case where data is accumulated.

We will return to the generalization to linear regression 
in \S\ref{subsubsec:gauss}.  There
we will derive the textbook intervals as conformal prediction regions
within the \textit{on-line Gaussian linear model}, an on-line compression model that 
uses slightly
weaker assumptions than the classical assumption of independent and 
normally distributed errors.

\subsection{Confidence says less than probability.}\label{subsec:less}

Neyman's notion of confidence looks at a procedure
before observations are made.
Before any of the $z_i$ are observed, the event~(\ref{eq:hitinequality})
involves multiple uncertainties:  
$\overline{z}_{n-1}$, $s_{n-1}$, and $z_n$ are all uncertain.
The probability that these three quantities
will turn out so that~(\ref{eq:hitinequality}) holds is $95\%$.

We might ask for more than this.
It is after we observe the first $n-1$ examples that we 
calculate $\overline{z}_{n-1}$ and $s_{n-1}$ and then
calculate the interval~(\ref{eq:fisherinterval}), and we would like
to be able to say at this point that there is still a $95\%$ probability
that $z_n$ will be in~(\ref{eq:fisherinterval}).  But this, it seems,
is asking for too much.  
The assumptions we have made are insufficient to enable us to find
a numerical probability for~(\ref{eq:hitinequality}) that will be 
valid at this late date.  In theory there is a 
conditional probability for~(\ref{eq:hitinequality}) 
given $z_1,\dots,z_{n-1}$, but it involves the
unknown mean and variance of the normal distribution.

Perhaps the matter is best understood from the game-theoretic 
point of view.  A probability can be thought of as an offer
to bet.  A 95\% probability, for example, is an offer to take either
side of a bet at $19$ to $1$ odds.  The probability is valid if the offer
does not put the person making it at a disadvantage, inasmuch as
a long sequence of equally reasonable offers will not allow an opponent
to multiply the capital he or she
risks by a large factor \cite{shafer/vovk:2001}.  
When we assume a probability model (such as the normal model we just
used or the on-line compression models we 
will study later), we are assuming that the model's probabilities 
are valid in this sense
before any examples are observed.  
Matters may be different afterwards.

In general, a 95\% conformal predictor is a rule for using 
the preceding examples $(x_1,y_1),\dots,(x_{n-1},y_{n-1})$ and a new 
object $x_n$ to give a set, say 
\begin{equation}\label{eq:regiontoday}
                  \Gamma^{0.05}((x_1,y_1),\dots,(x_{n-1},y_{n-1}),x_n),
\end{equation}
that we predict will contain $y_n$.  If the predictor is valid, the prediction
$$
          y_n \in  \Gamma^{0.05}((x_1,y_1),\dots,(x_{n-1},y_{n-1}),x_n)
$$
will have a 95\% probability before any of the examples are
observed, and it will be safe, at that point, to offer $19$ to $1$ odds on it.  
But after we observe
$(x_1,y_1),\dots,(x_{n-1},y_{n-1})$ and $x_n$ and calculate the set~(\ref{eq:regiontoday}),
we may want to withdraw the offer. 

Particularly striking instances of this phenomenon can arise in the case of classification,
where there are only finitely many possible labels.  We will see one such 
instance in \S\ref{subsubsec:nn}, where we consider a classification problem
in which there are only two possible labels, $\mathrm{s}$ and $\mathrm{v}$.
In this case, there are only four possibilities for the prediction region:
\begin{enumerate}
   \item
     $\Gamma^{0.05}((x_1,y_1),\dots,(x_{n-1},y_{n-1}),x_n)$ contains only $\mathrm{s}$.
   \item
     $\Gamma^{0.05}((x_1,y_1),\dots,(x_{n-1},y_{n-1}),x_n)$ contains only $\mathrm{v}$.
   \item
     $\Gamma^{0.05}((x_1,y_1),\dots,(x_{n-1},y_{n-1}),x_n)$ contains both $\mathrm{s}$ and $\mathrm{v}$.
   \item
     $\Gamma^{0.05}((x_1,y_1),\dots,(x_{n-1},y_{n-1}),x_n)$ is empty.
\end{enumerate}
The third and fourth cases can occur even though $\Gamma^{0.05}$ is valid.
When the third case happens, the prediction, though uninformative, is certain to be 
correct.  When the fourth case happens, the prediction is clearly wrong. 
These cases are consistent with the prediction being
right 95\% of the time.  But when we see
them arise, we know whether the
particular value of $n$ is one of the 95\% where we are right or the 
one of the 5\% where we are wrong, and so the $95\%$ will not 
remain valid as a probability defining betting odds.

In the case of normally distributed examples,
Fisher called the 95\% probability for $z_n$ being in 
the interval~(\ref{eq:fisherinterval})
a ``fiducial probability,''
and he seems to have believed that it would not be susceptible to 
a gambling opponent who knows the first $n-1$ examples
(see pp.~119--125 of \cite{fisher:1956}).  
But this turned out not to be the case \cite{robinson:1975}.  
For this and related reasons, most scientists who use Fisher's
methods have adopted the interpretation offered by 
Neyman, who wrote about ``confidence'' rather than
fiducial probability and emphasized that a confidence level is 
a full-fledged probability only before we acquire data.
It is the procedure or method, not the interval or region it produces 
when applied to
particular data, that has a 95\% probability of being correct.

Neyman's concept of confidence has endured in spite of its shortcomings.
It is widely taught and used in almost every branch of science.  
Perhaps it is especially useful in the on-line setting.
It is useful to know that 95\% of our predictions are correct
even if we cannot assert a full-fledged 95\% probability 
for each prediction when we make it.

\begin{figure}
\begingroup
  \begin{minipage}{.25\linewidth}
    \begin{center}
      \epsfig{file=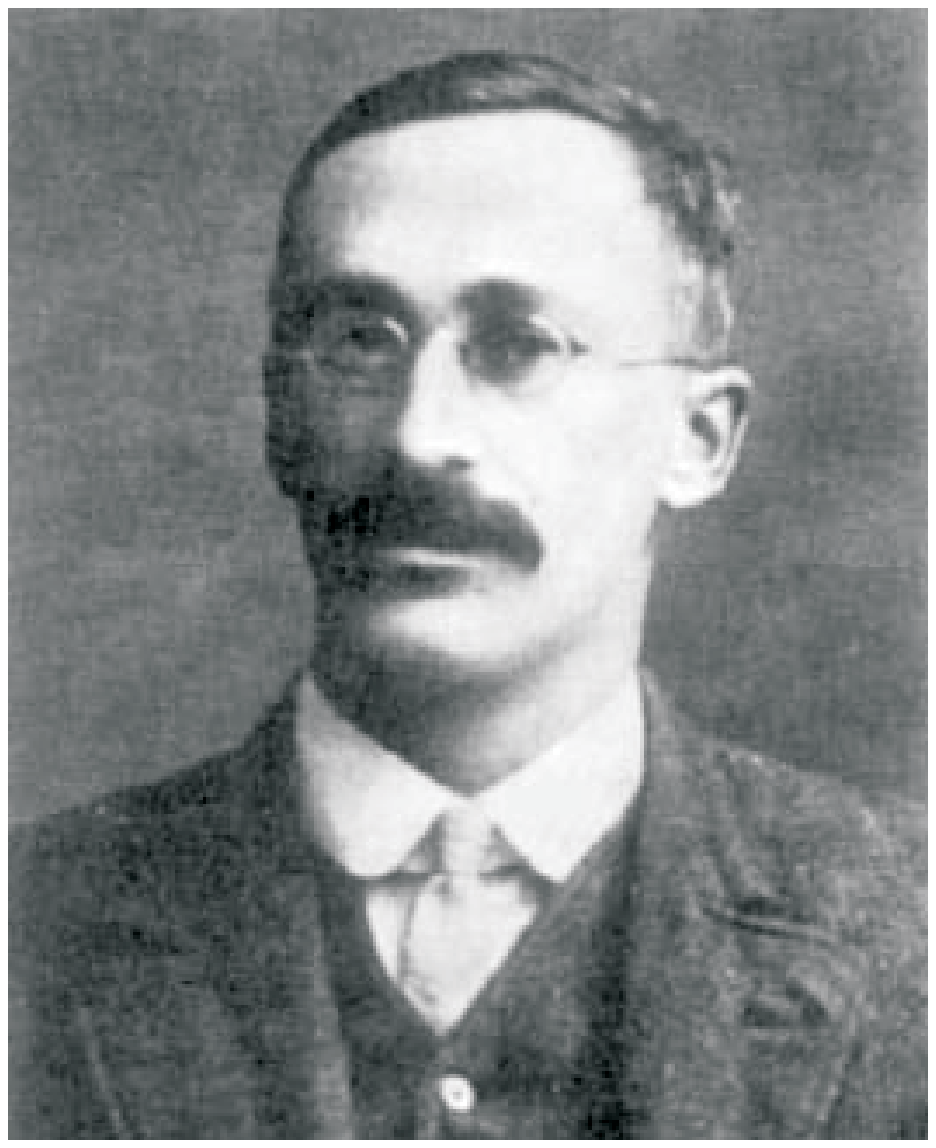,height=3cm}

    William S. Gossett\\
    1876--1937
    \end{center}
  \end{minipage}\hfil
  \begin{minipage}{.25\linewidth}
    \begin{center}
      \epsfig{file=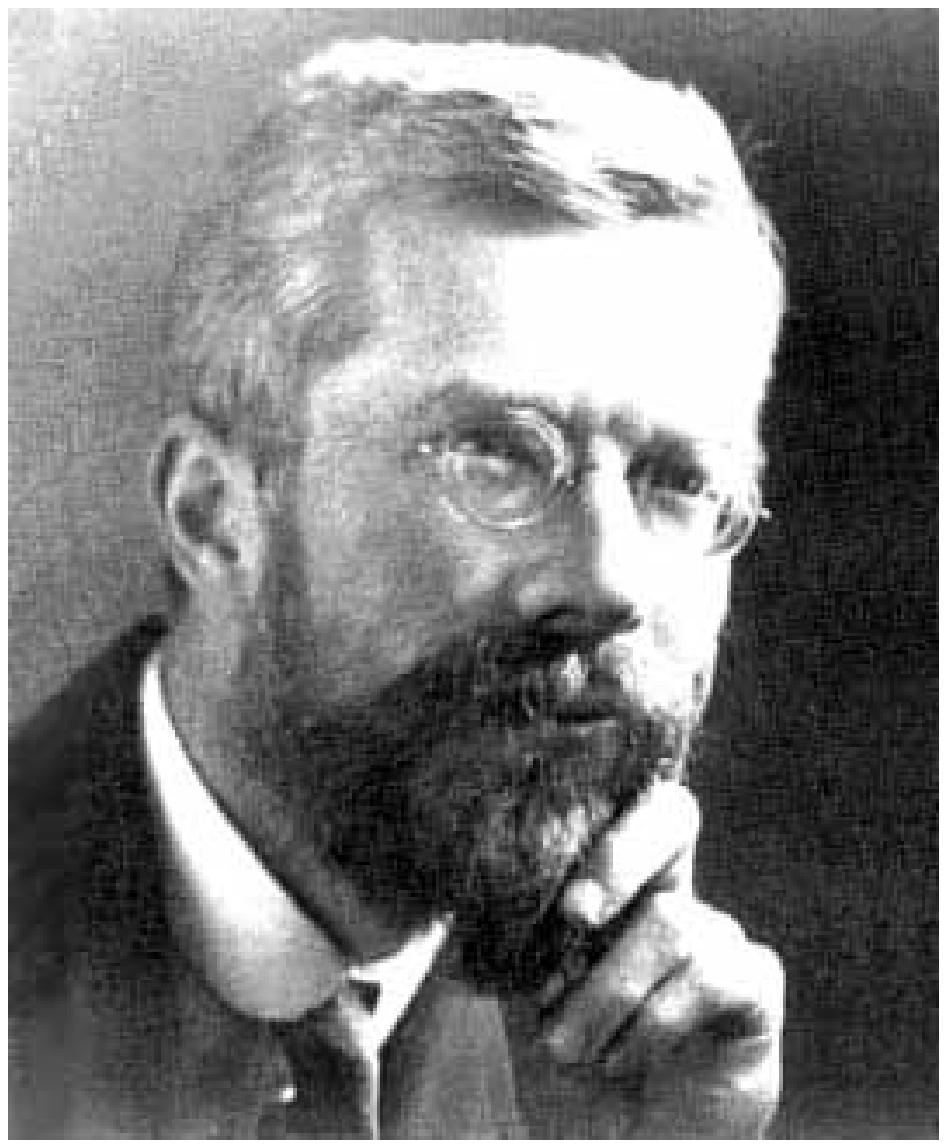,height=3cm}

    Ronald A. Fisher\\
    1890--1962
    \end{center}
  \end{minipage}\hfil
  \begin{minipage}{.25\linewidth}
    \begin{center}
      \epsfig{file=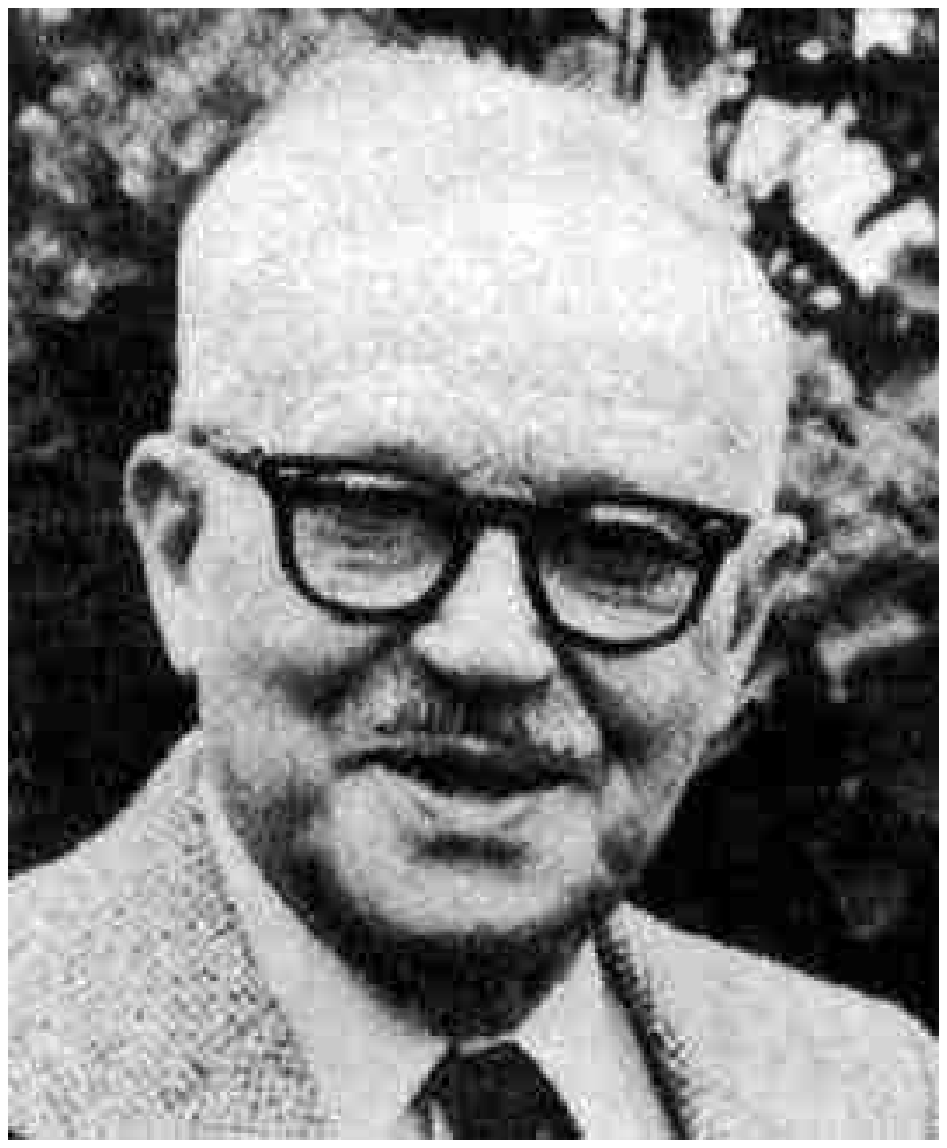,height=3cm}

    Jerzy Neyman\\
    1894--1981
    \end{center}
  \end{minipage}\hfil
  \endgroup
\caption{\textbf{Three influential statisticians.}  
         Gossett, who worked as a statistician for the Guinness brewery in Dublin,
         introduced the $t$-distribution to English-speaking statisticians
         in 1908 \cite{student:1908}.  Fisher, whose applied and theoretical
         work invigorated mathematical statistics in the 1920s and 1930s,
         refined, promoted, and extended Gossett's work.  Neyman was one
         of the most influential leaders in the subsequent movement to use
         advanced probability theory to give statistics a firmer foundation
         and further extend its applications.}
\end{figure}

\section{Exchangeability}\label{sec:exchangeable}

Consider variables $z_1,\dots,z_N$.
Suppose that for any collection of $N$ values,
the $N!$ different orderings are equally likely.
Then we say that $z_1,\dots,z_N$ are \textit{exchangeable}.  

Exchangeability is closely related to the idea that examples are drawn
independently from a probability distribution.  As we 
explain in the next section, \S\ref{sec:works}, it is the basic model for 
conformal prediction.  

In this section we look at the relationship between 
exchangeability and independence and then give a backward-looking
definition of 
exchangeability that can be understood game-theoretically.
We conclude with a law of large numbers for exchangeable sequences,
which will provide the basis for our confidence that 
our $95\%$ prediction regions are right $95\%$ of the time.

\subsection{Exchangeability and independence}\label{subsec:exind}

Although the definition of exchangeability we just gave may be clear enough
at an intuitive level,
it has two technical problems that make it inadequate as a formal
mathematical definition:  (1) in the case of continuous distributions, any specific
values for $z_1,\dots,z_N$ will have probability zero, and (2)
in the case of discrete distributions, two or more of the $z_i$ might take the same 
value, and so a list of possible values $a_1,\dots,a_N$ might contain 
fewer than $n$ distinct values.

One way of avoiding these technicalities is to
use the concept of a permutation, as follows:
\begin{quotation}
\noindent
  \textit{Definition of exchangeability using permutations.}
  The variables $z_1,\dots,z_N$ are exchangeable if for
  every permutation $\tau$ of the integers $1,\dots,N$,
  the variables $w_1,\dots,w_N$, where $w_i = z_{\tau(i)}$,
  have the same joint probability
  distribution as $z_1,\dots,z_N$.
\end{quotation}
We can extend this to a definition of exchangeability for 
an infinite sequence of variables:
$z_1,z_2,\dots$ are exchangeable if $z_1,\dots,z_N$ are
exchangeable for every $N$.\label{p:exchdef}

This definition makes it easy to see that independent and identically
distributed random variables
are exchangeable.  Suppose $z_1,\dots,z_N$ all take values from
the same example space $\bf Z$, all have the 
same probability distribution $Q$, and are independent. 
Then their joint distribution satisfies
\begin{equation}\label{eq:independent}
    \Prob \left(z_1 \in A_1 \; \& \; \dots \; \& \; z_N \in A_N \right)
       = Q(A_1) \cdots Q(A_N)
\end{equation}
for any%
\footnote{We leave aside technicalities 
          involving measurability.}
subsets $A_1,\dots,A_N$ of $\bf Z$,
where $Q(A)$ is the probability $Q$ assigns to an example
being in $A$.  Because permuting the factors $Q(A_n)$ does not
change their product, and because a joint probability distribution
for $z_1,\dots,z_N$ is determined by the probabilities it assigns
to events of the form $\{z_1 \in A_1  \; \& \; \dots \; \& \; z_N \in A_N \}$,
this makes it clear that $z_1,\dots,z_N$ are exchangeable.

\begin{table}
\renewcommand{\arraystretch}{1.5}
\begin{center}
\vbox{\hbox to\textwidth{\hfil\begin{tabular}{|c|c|}\hline
$\Prob(z_1=\mathrm{H} \; \&  \; z_2=\mathrm{H})$  & 
$\Prob(z_1=\mathrm{H} \; \&  \; z_2=\mathrm{T})$  \\ \hline
$\Prob(z_1=\mathrm{T} \; \&  \; z_2=\mathrm{H})$  & 
$\Prob(z_1=\mathrm{T} \; \&  \; z_2=\mathrm{T})$  \\ \hline
\end{tabular}\hfil}
\vspace{5mm}
\hbox to\linewidth{\hfil\begin{tabular}{|c|c|}\hline
$0.81$    &   $0.09$ \\ \hline
$0.09$     &   $0.01$ \\ \hline
\end{tabular}
\hfil
\begin{tabular}{|c|c|}\hline
$0.41$    &   $0.09$  \\ \hline
$0.09$     &   $0.41$ \\ \hline
\end{tabular}
\hfil
\begin{tabular}{|c|c|}\hline
$0.10$    &   $0.40$ \\ \hline
$0.40$     &   $0.10$ \\ \hline
\end{tabular}\hfil}}
\end{center}
\caption{\label{ta:exchange}\textbf{Examples of exchangeability.}
         We consider variables $z_1$ and $z_2$, each of which comes out
         H or T.  Exchangeability requires only that 
$
  \Prob(z_1=\mathrm{H} \; \&  \; z_2=\mathrm{T}) = \Prob(z_1=\mathrm{T} \; \&  \; z_2=\mathrm{H}).
$
         Three examples of distributions for $z_1$ and $z_2$ with this property are shown. 
         On the left, $z_1$ and $z_2$ are independent
         and identically distributed; both come out H with probability $0.9$.
         The middle example is obtained by averaging this distribution 
         with the distribution in 
         which the two variables are again independent and identically distributed
         but T's probability is $0.9$.  The distribution on the right,
         in contrast, cannot be obtained by averaging distributions under which the 
         variables are independent and identically distributed.  Examples of this 
         last type disappear as we ask for a larger and larger number of variables
         to be exchangeable.}
\renewcommand{\arraystretch}{1}
\end{table}

Exchangeability implies that variables have the same
distribution.  On the other hand, exchangeable variables 
need not be independent.  Indeed, when we average two or more distinct 
joint probability distributions
under which variables are independent, we usually get a joint 
probability distribution under which they are exchangeable 
(averaging preserves exchangeability) but not independent
(averaging usually does not preserve independence).
According to a famous theorem by de Finetti,
an exchangeable joint distribution for
an infinite sequence of distinct variables is exchangeable only
if it is a mixture of joint distributions under which the variables are independent
\cite{hewitt/savage:1955}.
As Table~\ref{ta:exchange} shows, the picture is more complicated in the finite
case.

\subsection{Backward-looking definitions of exchangeability}\label{subsec:backexc}

Another way of defining exchangeability looks backwards 
from a situation where we know the unordered values of $z_1,\dots,z_N$.

\begin{figure}
\begin{center}
      \epsfig{file=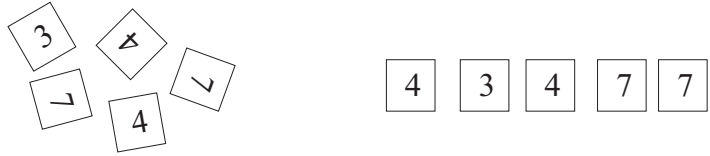,height=2cm}
\end{center}
\caption{\textbf{Ordering the tiles.}
         Joe gives Bill a bag containing five tiles, and Bill 
         arranges them to form the list $43477$.  
         Bill can calculate conditional probabilities for which
         $z_i$ had which of the five values.  His conditional
         probability for $z_5=4$, for example, is $2/5$.  
         There are $(5!)/(2!)(2!) = 30$ ways of assigning 
         the five values to the five variables;  
         $(z_1,z_2,z_3,z_4,z_5)=(4,3,4,7,7)$ is one of these,
         and they all have the same probability, $1/30$.}
\end{figure}

Suppose Joe has observed $z_1,\dots,z_N$.  He writes each value on a tile
resembling those used in $\text{Scrabble}^{\copyright}$, puts the $N$
tiles in a bag, shakes the bag, and gives it to Bill to inspect.
Bill sees the $N$ values (some possibly equal to each other) without
knowing their original order.  Bill also knows the 
joint probability distribution for $z_1,\dots,z_N$.  So he obtains
probabilities for the ordering of the tiles
by conditioning this joint distribution on his
knowledge of the bag.  The joint distribution is exchangeable if 
and only if these conditional probabilities are the same as the 
probabilities for the result of ordering the tiles by 
successively drawing them at random from the bag without replacement.

To make this into a definition of exchangeability, 
we formalize the notion of a bag.  A 
\textit{bag} (or \textit{multiset}, as it is sometimes called)
is a collection of elements in which repetition is 
allowed.  It is like a set
inasmuch as its elements are unordered but like a list inasmuch
as an element can occur more than once.  We write 
$
     \lbag a_1,\dots,a_N \rbag
$ 
for the bag obtained from the list $a_1,\dots,a_N$ by removing
information about the ordering.  

\begin{figure} 
  \begin{center}
    \unitlength=.5mm
    \begin{picture}(175,50)
      \put(-20,10){\makebox(0,0)[cc]{$\Box$}}
      \put(10,10){\makebox(0,0)[cc]{$\lbag z_1\rbag$}}
      \put(60,10){\makebox(0,0)[cc]{$\lbag z_1,z_2\rbag$}}
      \put(85,10){\makebox(0,0)[cc]{$\cdots$}}
      \put(120,10){\makebox(0,0)[cc]{$\lbag z_1,\dots,z_{N-1}\rbag$}}
      \put(170,10){\makebox(0,0)[cc]{$\lbag z_1,\dots,z_N\rbag$}}
      \put(10,40){\makebox(0,0)[cc]{$z_1$}}
      \put(60,40){\makebox(0,0)[cc]{$z_2$}}
      \put(120,40){\makebox(0,0)[cc]{$z_{N-1}$}}
      \put(170,40){\makebox(0,0)[cc]{$z_N$}}
      \put(-3,10){\vector(-1,0){10}}
      \put(38,10){\vector(-1,0){10}}
      \put(78,10){\vector(-1,0){6}}
      \put(96,10){\vector(-1,0){6}}
      \put(150,10){\vector(-1,0){6}}
      \put(10,15){\vector(0,1){20}}
      \put(60,15){\vector(0,1){20}}
      \put(120,15){\vector(0,1){20}}
      \put(170,15){\vector(0,1){20}}
    \end{picture}
  \end{center}
\caption{\label{fig:exchangeonestep}\textbf{Backward probabilities, step by step.}
         The two arrows backwards from
         each
         bag $\lbag z_1,\dots,z_n \rbag$ symbolize drawing
         an example $z_n$ out at random, leaving the 
         smaller bag $\lbag z_1,\dots,z_{n-1} \rbag$. 
         The probabilities for the result of the drawing are given by~(\ref{eq:onestepback}).
         Readers familiar with Bayes nets \cite{cowell/etal:1999} will recognize
         this diagram as an example; conditional on each variable, a joint probability distribution 
         is given for its children (the variables to which arrows from it point), and 
         given the variable, its descendants are independent of its ancestors.}
\end{figure}
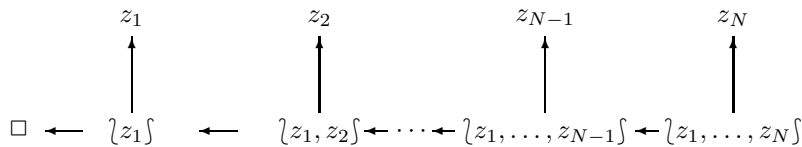

Here are three equivalent conditions on the joint distribution 
of a sequence of random variables $z_1,\dots,z_N$, any of which can be taken 
as the definition of exchangeability.
\begin{enumerate}
  \item
     For any bag $B$ of size $N$, and for any examples $a_1,\dots,a_N$,
$$
 \Prob \left(z_1 = a_1 \; \& \; \dots \; \& \; z_N = a_N \givn  \lbag z_1,\dots,z_N \rbag = B \right)
$$
     is equal to the probability that successive random drawings from the bag $B$ without 
     replacement produces first $a_N$, then $a_{N-1}$, and so on, until the last 
     element remaining in the bag is $a_1$.
   \item
     For any $n$, $1\leq n \leq N$,
     $z_n$ is independent of $z_{n+1},\dots,z_N$ given the bag
     $\lbag z_1,\dots,z_n \rbag$ and for any bag $B$ of size $n$,
\begin{equation}\label{eq:onestepback}
  \Prob \left(z_n = a \givn  \lbag z_1,\dots,z_n \rbag = B \right) 
      = 
       \frac{k}{n},  
\end{equation}
     where $k$ is the number of times $a$ occurs in $B$.
  \item
     For any bag $B$ of size $N$, and for any examples $a_1,\dots,a_N$,
\begin{multline}\label{eq:back}
 \Prob \left(z_1 = a_1 \; \& \; \dots \; \& \; z_N = a_N \givn  \lbag z_1,\dots,z_N \rbag = B \right) \\
      = 
    \begin{cases}
       \frac{n_1! \cdots n_k!}{N!}  &  \text{if } B = \lbag a_1,\dots,a_N \rbag\\
       0                            &  \text{if } B \neq \lbag a_1,\dots,a_N \rbag,
    \end{cases}
\end{multline}
     where $k$ is the number of distinct values among the $a_i$, and 
     $n_1,\dots,n_k$ are the respective numbers of times they occur.
     (If the $a_i$ are all distinct, 
     the expression $n_1! \cdots n_k!/(N!)$ reduces to $1/(N!)$.)
\end{enumerate}
We leave it to the reader to verify that these three conditions are 
equivalent to each other.
The second condition, which we will emphasize, is represented 
pictorially in Figure~\ref{fig:exchangeonestep}.

The backward-looking conditions are also equivalent to the definition of exchangeability using
permutations given on p.~\pageref{p:exchdef}.
This equivalence is elementary in the case where
every possible sequence of values $a_1,\dots,a_n$
has positive probability.  But complications arise
          when this probability is zero, because the conditional 
          probability on the left-hand side of~(\ref{eq:back}) 
          is then defined only with probability one by the 
          joint distribution.  We do not explore these
          complications here.

\subsection{The betting interpretation of exchangeability}\label{subsec:betexc}

The framework for probability developed in~\cite{shafer/vovk:2001}
formalizes classical 
results of probability theory, such as the law of large numbers,
as theorems of game theory:  a bettor can multiply the capital
he risks by a large factor if these results do not hold.  
This allows us to express the empirical interpretation of
given probabilities
in terms of betting, using what we call \textit{Cournot's principle}:  
the odds determined by the probabilities
will not allow a
bettor
to multiply the capital he or she risks by a large factor
\cite{shafer:2007}.

By applying this idea to the sequence of probabilities~(\ref{eq:onestepback}),
we obtain a betting interpretation of exchangeability.
Think of Joe and Bill as two
players in a game that moves backwards from point $N$ in 
Figure~\ref{fig:exchangeonestep}.  At each step,
Joe provides new information and Bill 
bets.  Designate by $\K_N$ the total capital Bill risks.  
He begins with this capital at $N$, and at each step 
$n$ he bets on what $z_n$ will turn out to be.  When he bets
at step $n$, he cannot risk losing more than he has at that point
(because he is not risking more than 
$\K_N$ in the whole game), but otherwise he can bet as much
as he wants for or against each possible value $a$ for $z_n$ 
at the odds
$(k/n):(1-k/n)$,
where $k$ is the number of elements
in the current bag equal to $a$.

For brevity, we write $B_n$ for the bag $\lbag z_1,\dots,z_n \rbag$, and 
for simplicity, we set
the initial capital $K_N$ equal to $\$1$.  This gives the following 
protocol:

\bigskip

\noindent
\textsc{The Backward-Looking Betting Protocol}\label{p:back}

\noindent
\textbf{Players:}  Joe, Bill

\parshape=7
\IndentI  \WidthI
\IndentI  \WidthI
\IndentI  \WidthI
\IndentII \WidthII
\IndentII \WidthII
\IndentII \WidthII
\IndentII \WidthII
\noindent
$\K_N := 1$.\\
Joe announces a bag $B_N$ of size $N$.\\
FOR $n=N,N-1,\dots,2,1$\\
  Bill bets on $z_n$ at odds set by~(\ref{eq:onestepback}).\\
  Joe announces $z_n \in B_n$.\\
  $\K_{n-1} := \K_n + \text{Bill's net gain}$.\\
  $B_{n-1} := B_n \setminus \lbag z_n \rbag$.

\noindent
\textbf{Constraint:}  Bill must move so that his 
capital $\K_n$ will be nonnegative for all $n$ no matter 
how Joe moves.

\bigskip

\noindent
Our betting interpretation of exchangeability is that Bill will not multiply
his initial capital $\K_N$ by a large factor in this game.  

The permutation definition of exchangeability
does not lead to an equally simple
betting interpretation, because the probabilities 
for $z_1,\dots,z_N$ to which the permutation definition refers are not determined by
the mere assumption of exchangeability.

\subsection{A law of large numbers for exchangeable sequences}\label{subsec:lln}

As we noted when we studied Fisher's prediction interval in~\S\ref{subsubsec:validity}, 
the validity of on-line prediction
requires more than having a high probability of a hit for each individual
prediction.  We also need a law of large numbers, so that we can 
conclude that a high proportion of the high-probability predictions will be correct.  
As we show in~\S\ref{subsec:fisherind}, the successive hits in the case of 
Fisher's region predictor are independent, so that the 
usual law of large numbers applies.  What can we say in  
the case of conformal prediction under exchangeability?

Suppose $z_1,\dots,z_N$ are exchangeable, drawn from an example 
space $\bf Z$.  In this context, we adopt the following definitions.
\begin{itemize}
  \item
    An event $E$ is an \textit{$n$-event}, where $1 \leq n \leq N$, if its happening or failing is 
    determined by the value of $z_n$ and the value of the bag $\lbag z_1,\dots,z_{n-1} \rbag$.
  \item
    An $n$-event $E$ is $\epsilon$-\textit{rare} if 
\begin{equation}\label{eq:thin}
        \Prob ( E \givn \lbag z_1,\dots,z_n \rbag  ) \leq \epsilon.
\end{equation}
\end{itemize}
The left-hand side of the inequality~(\ref{eq:thin}) is a random variable, 
because the bag $\lbag z_1,\dots,z_n \rbag$ is random.  The inequality says that 
this random variable never exceeds $\epsilon$.

As we will see in the next section,
the successive errors for a conformal predictor are $\epsilon$-rare $n$-events.
So the validity of conformal prediction follows from the following informal
proposition.
\begin{inproposition}\label{inprop:llnex}
Suppose $N$ is large, and the variables $z_1,\dots,z_N$ are exchangeable.
Suppose $E_n$ is an $\epsilon$-rare $n$-event
for $n=1,\dots,N$.
Then the law of large numbers applies;
with very high probability, no more than approximately the fraction
$\epsilon$ of the events $E_1,\dots,E_N$ will
happen. 
\end{inproposition}
In Appendix~\ref{app:proof}, we formalize this informal proposition in two ways:
classically and game-theoretically.

The classical approach appeals to the classical weak law of large numbers, which
tells us that if $E_1,\dots,E_N$ are mutually independent and each 
have probability exactly $\epsilon$, and $N$ is sufficiently large,
then there is a very high probability that the fraction of the events that
happen will be close to $\epsilon$.  We show in~\S\ref{subsec:exchind} that 
if~(\ref{eq:thin}) holds with equality, then
$E_n$ are mutually independent and each of them has unconditional
probability $\epsilon$.  Having the inequality instead of equality
means that the $E_n$ are even less likely to happen,
and this will not reverse the conclusion that few of them will happen.

The game-theoretic approach is more straightforward, because the
game-theoretic version law of large numbers does not require independence
or exact levels of probability.  In the game-theoretic framework, 
the only question is whether the probabilities specified for 
successive events are rates at which 
a bettor can place successive bets.  The 
Backward-Looking Betting Protocol says that this is the 
case for $\epsilon$-rare $n$-events.  As Bill moves through
the protocol from $N$ to $1$, he is allowed to bet 
against each error $E_n$ at a rate corresponding to its having probability
$\epsilon$ or less.  So the game-theoretic weak law of large
numbers (\cite{shafer/vovk:2001}, pp.~124--126) applies directly.
Because the game-theoretic framework is not well
known, we state and prove this law of large numbers, specialized to the
Backward-Looking Betting Protocol, in~\S\ref{subsec:gameind}.

\section{Conformal prediction under exchangeability}\label{sec:works}

We are now in a position to state the conformal algorithm under 
exchangeability and explain 
why it produces valid nested prediction regions.  

We distinguish two cases of on-line prediction.  In both cases, we observe
examples $z_1,\dots,z_N$ one after the other and repeatedly
predict what we will observe next.
But in the second case we have more to go on when we make each prediction.
\begin{enumerate}
   \item
      \textit{Prediction from old examples alone.}
      Just before observing $z_n$, we predict it based on the previous
      examples, $z_1,\dots,z_{n-1}$.  
   \item
      \textit{Prediction using features of the new object.}
      Each example $z_i$ consists of an object $x_i$ and a label $y_i$.
      In symbols:  $z_i = (x_i,y_i)$.  We observe in sequence 
      $x_1,y_1,\dots,x_N,y_N$.  Just before observing $y_n$, we predict it     
      based on what we have observed so far, $x_n$ and the previous
      examples $z_1,\dots,z_{n-1}$.
\end{enumerate}
Prediction from old examples may seem relatively uninteresting. 
It can be considered a special case of 
prediction using features $x_n$ of new examples---the case 
in which the $x_n$ provide no information, and this special 
case we may have too little information to make useful predictions.
But its simplicity makes prediction with old examples alone
advantageous as a setting
for explaining the conformal algorithm, and as we will see,
it is then straightforward to take account of the new information
$x_n$.

Conformal prediction requires that we first choose a 
nonconformity measure, which measures how different a new example
is from old examples.  In \S\ref{subsec:nonconformity}, we explain how 
nonconformity measures can be obtained from methods of point prediction.
In \S\ref{subsec:oldalone}, we state and illustrate the conformal
algorithm for predicting new examples from old examples alone.
In \S\ref{subsec:features}, we generalize to 
prediction with the help of features of a new example.  
In \S\ref{subsec:nootherway}, we explain 
why conformal prediction produces the best possible
valid nested prediction regions under exchangeability.
Finally, in~\S\ref{subsec:seldom} we discuss the implications 
of the failure of the assumption of exchangeability.

For some readers, the simplicity of the conformal algorithm may
be obscured by its generality and 
the scope of our preliminary discussion of
nonconformity measures.  We encourage such readers to look
first at~\S\ref{subsubsec:average},
\S\ref{subsubsec:nn}, and \S\ref{subsubsec:width}, which provide largely 
self-contained accounts of the algorithm as it applies to some small 
datasets.

\subsection{Nonconformity measures}\label{subsec:nonconformity}

The starting point for conformal prediction is what we call 
a \textit{nonconformity measure}, a real-valued function 
$
     A(B,z)
$
that measures how different an example $z$ is from the examples 
in a bag $B$.  The conformal algorithm
assumes that a nonconformity measure has been chosen.
The algorithm will produce valid nested prediction regions 
using any real-valued function $A(B,z)$ as the nonconformity
measure.  But the prediction regions will be efficient
(small) only if $A(B,z)$ measures well how different $z$
is from the examples in $B$.

A method $\hat{z}(B)$ for obtaining a point prediction $\hat{z}$
for a new example from a bag $B$ of old examples
usually leads naturally to a nonconformity measure $A$.  
In many cases, we only need to add a way of measuring the distance 
$d(z,z^{\prime})$ between two examples.  Then we define $A$ by 
\begin{equation}\label{eq:construct}
     A(B,z) := d(\hat{z}(B),z).
\end{equation}
The prediction regions produced by the conformal algorithm do not 
change when the nonconformity measure $A$ is transformed monotonically.
If $A$ is nonnegative, for example, replacing $A$ with $A^2$ will make no difference.
Consequently, the choice
of the distance measure $d(z,z^{\prime})$ is relatively unimportant.
The important step in determining the nonconformity measure $A$ is
choosing the point predictor $\hat{z}(B)$.

To be more concrete, suppose the examples are real numbers,
and write $\overline{z}_B$
for the average of the numbers in $B$.  If we take this average 
as our point predictor $\hat{z}(B)$, and we measure
the distance between two real numbers by the absolute value of 
their difference, then~(\ref{eq:construct}) becomes
\begin{equation}\label{eq:simplencm}
      A(B,z) := |\overline{z}_B - z|.
\end{equation}
If we use the median
of the numbers in $B$ instead of their average as $\hat{z}(B)$,
we get a different nonconformity measure, which will produce
different prediction regions when we use the conformal algorithm.
On the other hand, as we have already said, it will make 
no difference if we replace the 
absolute difference $d(z,z^{\prime}) = |z - z^{\prime}|$ with 
the squared difference $d(z,z^{\prime}) = (z - z^{\prime})^2$,
thus squaring $A$.

We can also vary~(\ref{eq:simplencm}) by including
the new example in the average:
\begin{equation}\label{eq:inclusioncm}
       A(B,z) := \left| (\text{average of $z$ and all the examples in $B$}) - z \right|.
\end{equation}
This results in the same prediction regions as~(\ref{eq:simplencm}),
because if $B$ has $n$ elements, then 
\begin{multline*}
     \left| (\text{average of $z$ and all the examples in $B$}) - z \right| \\
           = \left| \frac{n\overline{z}_B + z}{n+1} - z\right| = \frac{n}{n+1} |\overline{z}_B - z|,
\end{multline*}
and as we have said, conformal prediction regions are not changed
by a monotonic transformation of the nonconformity measure.
In the numerical example that we give in~\S\ref{subsubsec:average} below,
we use~(\ref{eq:inclusioncm}) as our nonconformity measure.

When we turn to the case where features of a new object help 
us predict a new label, we will consider, among others, 
the following two nonconformity 
measures:

\bigskip

\noindent
\textbf{Distance to the nearest neighbors for classification.}  
Suppose $B=\lbag z_1,\dots,z_{n-1}\rbag$, where each $z_i$   
consists of a number $x_i$ and a nonnumerical label $y_i$.
Again we observe $x$ but not $y$ for a new 
      example $z=(x,y)$.  The nearest-neighbor method finds the $x_i$ closest to 
      $x$ and uses its label $y_i$ as our prediction of $y$.  
If there are only two labels, or if there is no natural way to measure the
distance between labels, we cannot measure how wrong the prediction is; it
      is simply right or wrong.  But it is natural to measure the nonconformity 
      of the new example $(x,y)$ to the old examples $(x_i,y_i)$ by comparing 
      $x$'s distance to old objects with the same label to its distance to 
      old objects with a different label.  For example, we can set
\begin{equation}\label{eq:nnnm}
\begin{split}
  A(B,z)
  :&= 
  \frac
  {
    \min \{ |x_i - x| : 1 \leq i \leq n-1, y_i = y\}
  }
  {
    \min \{ |x_i - x| : 1 \leq i \leq n-1, y_i \neq y \}
  }\\
     &=  \frac{\text{distance to $z$'s nearest 
                 neighbor in $B$ with the same label}}
         {\text{distance to $z$'s nearest neighbor 
                      in $B$ with a different label}}.
\end{split}
\end{equation}

\bigskip

\noindent
\textbf{Distance to a regression line.}
      Suppose $B=\lbag (x_1,y_1),\dots,(x_l,y_l)\rbag$, where the $x_i$ and $y_i$ 
      are numbers.  The most common way 
      of fitting a line to such pairs of numbers is to calculate the averages 
$$
            \overline{x}_l := \sum_{j=1}^l x_j
     \quad \text{and} \quad
            \overline{y}_l := \sum_{j=1}^l y_j,
$$
     and then the coefficients
$$
        b_l = \frac{\sum_{j=1}^l (x_j - \overline{x}_l) y_j}{\sum_{j=1}^l (x_j-\overline{x}_l)^2} 
     \quad \text{and} \quad
        a_l = \overline{y}_l - b_l \overline{x}_l.
$$
     This gives the \textit{least-squares line}
     $y =  a_l + b_l x$.  
     The coefficients $a_l$ and $b_l$ are not affected
     if we change the order of the $z_i$; they depend only on the 
     bag $B$.

     If we observe a bag $B=\lbag z_1,\dots,z_{n-1}\rbag$ of examples
     of the form $z_i=(x_i,y_i)$
     and also $x$ but not $y$ for a new example $z=(x,y)$, 
     then the least-squares prediction of $y$ is 
\begin{equation}\label{eq:regressionline}
           \hat{y} =  a_{n-1} + b_{n-1}x.
\end{equation}
      We can use the error in this prediction as a nonconformity measure:
$$
       A(B,z) := |y - \hat{y}| = |y - (a_{n-1} + b_{n-1} x)|.
$$
      We can obtain other nonconformity measures by using other methods 
      to estimate a line.  

Alternatively, we can include the new example as one of the examples used to estimate the 
least squares line or some other regression line.  
In this case, it is natural to write $(x_n,y_n)$ for the new example.
Then $a_n$ and $b_n$ designate the coefficients calculated from all $n$ examples, and
we can use 
\begin{equation}\label{eq:lsnoncon}
           |y_i - (a_n + b_n x_i)|
\end{equation}
to measure the nonconformity of each of the $(x_i,y_i)$
with the others. In general, the inclusion of the new example 
simplifies the implementation or at least the explanation of the conformal
algorithm.  In the case of least squares,
it does not change the prediction regions.

\subsection{Conformal prediction from old examples alone}\label{subsec:oldalone}

Suppose we have chosen a nonconformity measure $A$ for 
our problem.
Given $A$, and given the assumption that the $z_i$ are exchangeable,
we now define a valid prediction region 
$$
      \gamma^{\epsilon}(z_1,\dots,z_{n-1}) \subseteq {\bf Z},
$$
where ${\bf Z}$ is the example space.  We do this by 
giving an algorithm for deciding, for each $z \in {\bf Z}$, whether
$z$ should be included in the region.  For simplicity in stating 
this algorithm, we provisionally use the symbol $z_n$ for $z$,
as if we were assuming that $z_n$ is in fact equal to $z$.

\bigskip

\noindent
\fbox{
\begin{minipage}{\linewidth}
\noindent
\textbf{The Conformal Algorithm Using Old Examples Alone}\label{p:algorithm}

\smallskip
\noindent
\textit{Input:}  
Nonconformity measure $A$, significance level $\epsilon$, examples $z_1,\dots,z_{n-1}$,  example $z$,

\smallskip
\noindent
\textit{Task:}  Decide whether to include $z$ in $\gamma^{\epsilon}(z_1,\dots,z_{n-1})$.

\smallskip
\noindent
\textit{Algorithm:}
\begin{enumerate}
  \item
     Provisionally set $z_n := z$.
  \item
     For $i=1,\dots,n$, set $\alpha_i := A(\lbag z_1,\dots,z_n \rbag \setminus \lbag z_i \rbag,z_i)$.
  \item
     Set
     $\displaystyle
        p_z := \frac{\text{number of $i$ such that $1\leq i \leq n$ and } \alpha_i \ge \alpha_n}{n}\;.
     $
   \item
     Include $z$ in $\gamma^{\epsilon}(z_1,\dots,z_{n-1})$ 
     if and only if $p_z > \epsilon$.
\end{enumerate}
\end{minipage}}

\bigskip

\noindent
If $\bf Z$ has only a few elements, this algorithm can be implemented in 
a brute-force way:  calculate $p_z$ for 
every $z\in {\bf Z}$.  If $\bf Z$ has many elements, we will need some
other way of identifying the $z$ satisfying $p_z > \epsilon$. 
 
The number $p_z$ is the fraction of the examples in $\lbag z_1,\dots,z_{n-1},z\rbag$
that are at least as different from the others as $z$ is, in the sense 
measured by $A$.  So the algorithm tells us to form a prediction
region consisting of the $z$ that are not among the fraction $\epsilon$ most
out of place when they are added to the bag of old examples.

The definition of $\gamma^{\epsilon}(z_1,\dots,z_{n-1})$ 
can be framed as an application of 
the widely accepted Neyman-Pearson theory for hypothesis testing and confidence intervals
\cite{lehmann:1959}.  In the Neyman-Pearson theory, we test a hypothesis $H$ using 
a random variable $T$ that is likely to be large if $H$ is false.  
Once we observe $T=t$, we calculate 
$
    p_H := \Prob (T \geq t \givn H).
$
We reject $H$ at level $\epsilon$ if $p_H \leq \epsilon$.
Because this happens under $H$ with probability no more than $\epsilon$, we 
can declare $1-\epsilon$ confidence that the true hypothesis $H$ is among those 
not rejected.
Our procedure makes these choices of $H$ and $T$:
\begin{itemize}
   \item
      The hypothesis $H$ says the bag of the first $n$ examples 
      is $\lbag z_1,\dots,z_{n-1},z \rbag$.
   \item
      The test statistic $T$ is the random value of $\alpha_n$.  
\end{itemize}
Under $H$---i.e., conditional on the bag $\lbag z_1,\dots,z_{n-1},z \rbag$,
$T$ is equally likely to come out equal to any of the 
$\alpha_i$.  Its observed value is $\alpha_n$.  So 
$$
    p_H 
         = \Prob( T \geq \alpha_n \givn \lbag z_1,\dots,z_{n-1},z \rbag) = p_z.
$$
Since $z_1,\dots,z_{n-1}$ are known, rejecting the bag $\lbag z_1,\dots,z_{n-1},z \rbag$
means rejecting $z_n=z$.  So our $1-\epsilon$ confidence is in the set of $z$ 
for which $p_z > \epsilon$.

The regions $\gamma^{\epsilon}(z_1,\dots,z_{n-1})$ for successive $n$ are
based on overlapping observations rather than independent observations.
But the successive errors are 
$\epsilon$-rare $n$-events.  The event that our $n$th prediction is an 
error, $z_n \notin \gamma^{\epsilon}(z_1,\dots,z_{n-1})$, is the event
$p_{z_n} \leq \epsilon$.  This is an $n$-event, because the value of $p_{z_n}$ is 
determined by $z_n$ and the bag $\lbag z_1,\dots,z_{n-1}\rbag$.
It is $\epsilon$-rare because it is the event that 
$\alpha_n$ is among a fraction $\epsilon$ or fewer of the $\alpha_i$ that
are strictly larger than all the other $\alpha_i$, and this can have probability
at most $\epsilon$ when the $\alpha_i$ are exchangeable.
So it follows from Informal Proposition~\ref{inprop:llnex} (\S\ref{subsec:lln})
that we can expect 
at least $1-\epsilon$ of the $\gamma^{\epsilon}(z_1,\dots,z_{n-1})$,
$n=1,\ldots,N$, to be correct.

\subsubsection{Example:  Predicting a number with an average}\label{subsubsec:average}

In~\S\ref{subsec:normaldistribution}, 
we discussed Fisher's $95\%$ prediction interval for
$z_n$ based on $z_1,\dots,z_{n-1}$, which is valid under the assumption that
the $z_i$ are independent and normally distributed.
We used it to predict
$z_{20}$ when the first $19$ $z_i$ are
$$
   17, 20, 10, 17, 12, 15, 19, 22, 17, 19, 14, 22, 18, 17, 13, 12, 18, 15, 17.
$$
Taking into account our knowledge that the $z_i$ are all integers,
we arrived at the $95\%$ prediction 
that $z_{20}$ is an integer between
$10$ to $23$, inclusive.

What can we predict about $z_{20}$ 
at the $95\%$ level if we drop the assumption of normality
and assume only exchangeability?
To produce a 95\% prediction interval valid under the exchangeability
assumption alone, we reason as follows.
To decide whether to include a particular value $z$
in the interval, we consider twenty numbers
that depend on $z$:
\begin{itemize}
    \item
       First, the deviation of $z$ from the average of it and
       the other $19$ numbers.  Because the sum of 
       the $19$ is $314$, this is
\begin{equation}\label{eq:lastdev}
     \left| \frac{314+z}{20} - z \right| = \frac{1}{20} \left| 314 - 19 z \right|.
\end{equation}
    \item
       Then, for $i=1,\dots,19$, the deviation of $z_i$ from this same
       average.  This is
\begin{equation}\label{eq:firstdevs}
        \left| \frac{314 + z}{20} - z_i \right| 
      =  \frac{1}{20} \left| 314 + z - 20 z_i \right| .            
\end{equation}
\end{itemize}
Under the hypothesis that $z$ is the actual value of $z_n$, these $20$ numbers
are exchangeable.  Each of them is as likely as the other to be the largest.
So there is at least a $95\%$ ($19$ in $20$) chance that~(\ref{eq:lastdev}) will not
exceed the largest of the $19$ numbers in~(\ref{eq:firstdevs}).
The largest of the $19$ $z_i$s being $22$
and the smallest $10$, we can write this condition as 
\begin{equation}\label{eq:imitation}
   \left| 314 - 19 z \right| \leq 
     \max \left\{ \left| 314 + z - (20 \times 22) \right| ,
             \left| 314 + z - (20 \times 10) \right| \right\},
\end{equation}
which reduces to 
$$
                       10 \leq z \leq \frac{214}{9} \approx 23.8.
$$
Taking into account that $z_{20}$ is an 
integer, our $95\%$ prediction is 
that it will be an integer
between $10$ and $23$, inclusive.  This is exactly the same prediction we obtained
by Fisher's method.
We have lost nothing by weakening the assumption that the $z_i$ are
independent and normally distributed to the assumption that they 
are exchangeable.  But we are still basing our prediction
region on the average of old examples, which is an optimal estimator in 
various respects under the assumption of normality.

\subsubsection{Are we complicating the story unnecessarily?}

The reader may feel that we are vacillating about whether to
include the new example in the bag with which we are comparing it.
In our statement of the conformal algorithm, we define the 
nonconformity scores by
\begin{equation}\label{eq:againscore}
     \alpha_i := A(\lbag z_1,\dots,z_n \rbag \setminus \lbag z_i \rbag,z_i),
\end{equation}
apparently signaling that we do not want to include $z_i$ in the bag
to which it is compared.  But then we use the 
nonconformity measure
$$
       A(B,z) := \left| (\text{average of $z$ and all the examples in $B$}) - z \right|,
$$
which seems to put $z$ back in the bag, reducing~(\ref{eq:againscore}) to
$$
       \alpha_i = \left| \frac{\sum_{j=1}^n z_j}{n} - z_i \right|.
$$
We could have reached this point more easily by writing 
\begin{equation}\label{eq:alal}
     \alpha_i := A(\lbag z_1,\dots,z_n \rbag,z_i)
\end{equation}
in the conformal algorithm
and using 
$
     A(B,z) := \left| \overline{z}_B - z \right|.
$

The two ways of defining nonconformity scores, (\ref{eq:againscore}) and~(\ref{eq:alal}),
are equivalent, inasmuch as whatever we can get with one of them we can get from the 
other by changing the nonconformity measure.  In this case, (\ref{eq:alal})
might be more convenient.  But we will see other cases where~(\ref{eq:againscore}) 
is more convenient.  We also have another reason for using~(\ref{eq:againscore}).
It is the form that generalizes, as we will see in~\S\ref{sec:compression}, to on-line 
compression models.

\subsection{Conformal prediction using a new object}\label{subsec:features}

Now we turn to the case where our example space $\bf Z$ is of the 
form ${\bf Z} = {\bf X} \times {\bf Y}$.  We call $\bf X$ the 
\textit{object space}, $\bf Y$ the \textit{label space}.
We observe in sequence examples $z_1,\dots,z_N$, where $z_i = (x_i,y_i)$.
At the point where we have observed
$$
    z_1,\dots,z_{n-1},x_n = (x_1,y_1),\dots,(x_{n-1},y_{n-1}), x_n,
$$
we want to predict $y_n$ by giving a prediction region 
$$
      \Gamma^{\epsilon}(z_1,\dots,z_{n-1},x_n) \subseteq {\bf Y}
$$
that is valid at the $(1-\epsilon)$ level.
As in the special case where the $x_i$ are absent, we start with a nonconformity 
measure $A(B,z)$.  

We define the prediction region by 
giving an algorithm for deciding, for each $y \in {\bf Y}$, whether
$y$ should be included in the region.  For simplicity in stating 
this algorithm, we provisionally use the symbol $z_n$ for $(x_n,y)$,
as if we were assuming that $y_n$ is in fact equal to $y$.

\bigskip

\noindent
\fbox{
\begin{minipage}{\linewidth}
\noindent
\textbf{The Conformal Algorithm}

\smallskip
\noindent
\textit{Input:}  
Nonconformity measure $A$, significance level $\epsilon$,
examples $z_1,\dots,z_{n-1}$, object $x_n$, label $y$

\smallskip
\noindent
\textit{Task:}  Decide whether to include $y$ in $\Gamma^{\epsilon}(z_1,\dots,z_{n-1},x_n)$.

\smallskip
\noindent
\textit{Algorithm:}
\begin{enumerate}
  \item
     Provisionally set $z_n :=(x_n,y)$.
  \item
     For $i=1,\dots,n$, set $\alpha_i := A(\lbag z_1,\dots,z_n \rbag \setminus \lbag z_i \rbag,z_i)$.
  \item
     Set
     $\displaystyle
        p_y := \frac{\#\{i=1,\dots,n\st \alpha_i \ge \alpha_n\}}{n}\;.
     $
   \item
     Include $y$ in $\Gamma^{\epsilon}(z_1,\dots,z_{n-1},x_n)$ 
     if and only if $p_y > \epsilon$.
\end{enumerate}
\end{minipage}}

\bigskip

\noindent
This differs only slightly from the conformal algorithm using old 
examples alone (p.~\pageref{p:algorithm}).  Now we write $p_y$
instead of $p_z$, and we say that we are including
$y$ in $\Gamma^{\epsilon}(z_1,\dots,z_{n-1},x_n)$ instead of 
saying that we are including
$z$ in $\gamma^{\epsilon}(z_1,\dots,z_{n-1})$.

To see that
this algorithm produces valid 
prediction regions, it suffices to see that it consists
of the algorithm for old examples alone together with a further
step that does not change the frequency of hits.
We know that the region the old algorithm produces,
\begin{equation}\label{eq:oldca}
     \gamma^{\epsilon}(z_1,\dots,z_{n-1}) \subseteq {\bf Z},
\end{equation}
contains the new example $z_n = (x_n,y_n)$ at least
$95\%$ of the time.  Once we know $x_n$, we can rule out all
$z=(x,y)$ in~(\ref{eq:oldca}) with $x \neq x_n$.
The $y$ not ruled out, those such that $(x_n,y)$
is in~(\ref{eq:oldca}), are precisely those in the set
\begin{equation}\label{eq:newca}
      \Gamma^{\epsilon}(z_1,\dots,z_{n-1},x_n) \subseteq {\bf Y}
\end{equation}
produced by our new algorithm.  Having
$(x_n,y_n)$ in~(\ref{eq:oldca}) $1-\epsilon$ of the time is 
equivalent to having
$y_n$ in~(\ref{eq:newca}) $1-\epsilon$ of the time.

\subsubsection{Example:  Classifying iris flowers}\label{subsubsec:nn}

\begin{figure}
  \begin{center}
      \epsfig{file=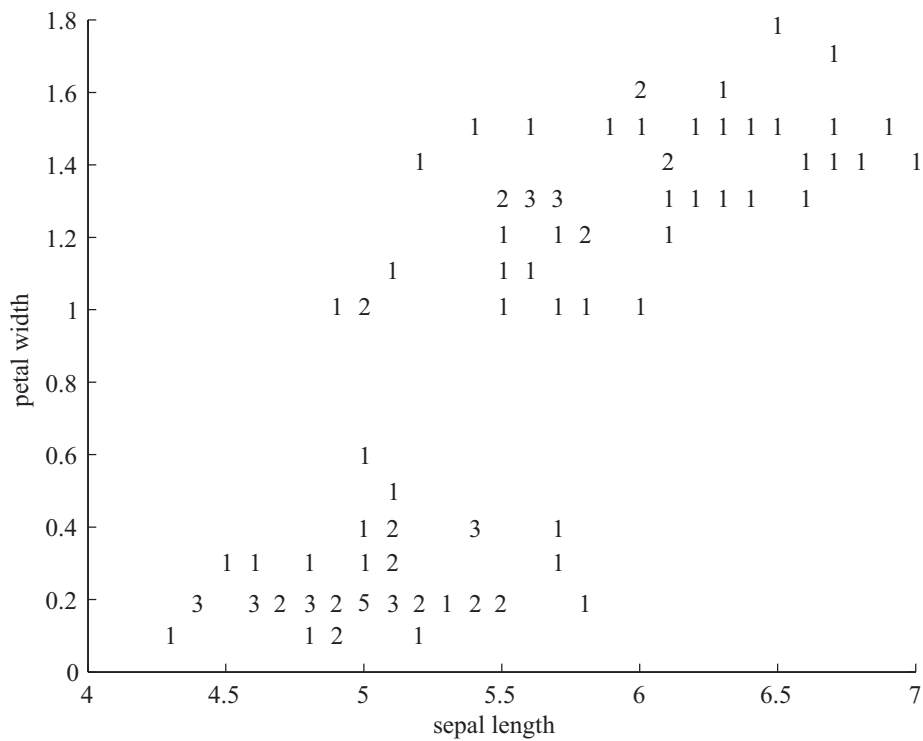,width=\linewidth}
  \end{center}
\caption{\label{fig:plot} \textbf{Sepal length, petal width, and species
                           for Edgar Anderson's $100$ flowers.}  
                           The $50$ \textit{iris setosa} are clustered at the 
                           lower left, while the $50$ \textit{iris versicolor}
                           are clustered at the upper right.  The numbers indicate
                           how many plants have exactly the same measurement; for 
                           example, there are $5$ plants that have sepals $5$ inches
                           long and petals $0.2$ inches wide.
                           Petal width separates the two species perfectly; all $50$ 
                           \textit{versicolor} petals are $1$ inch wide or wider,
                           while all \textit{setosa} petals are narrower than $1$ inch.
                           But there is substantial overlap in sepal length.}
\end{figure}

In 1936 \cite{fisher:1936}, R.~A. Fisher used discriminant analysis to 
distinguish different 
species of iris on the basis of measurements of their flowers.  
The data he used included
measurements by Edgar Anderson of flowers from 
$50$ plants each of two species, 
\textit{iris setosa} and \textit{iris versicolor}.
Two of the measurements, sepal length and petal width,
are plotted in Figure~\ref{fig:plot}.

\begin{figure}
  \begin{center}
      \epsfig{file=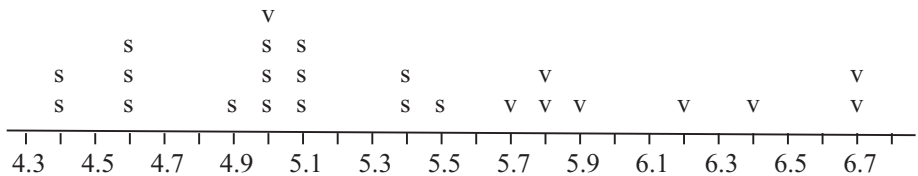,width=\linewidth}
  \end{center}
\caption{\label{fig:iris} \textbf{Sepal length and species for the first $24$ plants in
                          our random sample of size $25$.}  Except for one \textit{versicolor}
                          with sepal length $5.0$, the \textit{versicolor} in this sample
                          all have longer
                          sepals than the \textit{setosa}.  This high degree of separation 
                          is an accident of the sampling.}
\end{figure}

To illustrate how the conformal
algorithm can be used for classification, we  
have randomly chosen $25$ of the $100$ plants.  The sepal
lengths and species for the first $24$ of them are listed in 
Table~\ref{ta:iris} and plotted in Figure~\ref{fig:iris}.
The $25$th plant in the sample has sepal length $6.8$.
On the basis of this information, would you classify it
as \textit{setosa} or \textit{versicolor}, and how 
confident would you be in the classification?  
Because $6.8$ is the longest sepal in the sample,
nearly any reasonable method will classify the plant as 
\textit{versicolor}, and this is in fact the correct answer.  
But the appropriate level of confidence
is not so obvious.

\begin{table}
\begin{center}
\small
\begin{tabular}{ c  c  c  c  c  c c  c  c } 
          & \multicolumn{2}{c}{Data}& \multicolumn{6}{c}{Nonconformity scores} \\ \cline{4-9}
          & \multicolumn{2}{c}{}& \multicolumn{2}{c}{NN}&  \multicolumn{2}{c}{Species Average} & \multicolumn{2}{c}{SVM}  \\ \cline{2-9}
          & sepal  & species & $\alpha_i$ for & $\alpha_i$ for & $\alpha_i$ for  & $\alpha_i$ for & $\alpha_i$ for & $\alpha_i$ for\\
          & length &         & $y_{25}=\mathrm{s}$ & $y_{25}=\mathrm{v}$ & $y_{25}=\mathrm{s}$ & $y_{25}=\mathrm{v}$ & $y_{25}=\mathrm{s}$ & $y_{25}=\mathrm{v}$\\ \hline
 $z_1$    & 5.0     & s      & 0              & 0               & 0.06  & 0.06    & 0                   & 0\\
 $z_2$    & 4.4     & s      & 0              & 0               & 0.66  & 0.54    & 0                   & 0\\
 $z_3$    & 4.9     & s      & 1              & 1               & 0.16  & 0.04    & 0                   & 0\\
 $z_4$    & 4.4     & s      & 0              & 0               & 0.66  & 0.54    & 0                   & 0\\
 $z_5$    & 5.1     & s      & 0              & 0               & 0.04  & 0.16    & 0                   & 0\\ \hline
 $z_6$    & 5.9     & v      & 0.25           & 0.25            & 0.12  & 0.20    & 0                   & 0\\
 $z_7$    & 5.0     & s      & 0              & 0               & 0.06  & 0.06    & 0                   & 0\\
 $z_8$    & 6.4     & v      & 0.50           & 0.22            & 0.38  & 0.30    & 0                   & 0\\
 $z_9$    & 6.7     & v      & 0              & 0               & 0.68  & 0.60    & 0                   & 0\\
 $z_{10}$ & 6.2     & v      & 0.33           & 0.29            & 0.18  & 0.10    & 0                   & 0\\ \hline
 $z_{11}$ & 5.1     & s      & 0              & 0               & 0.04  & 0.16    & 0                   & 0\\
 $z_{12}$ & 4.6     & s      & 0              & 0               & 0.46  & 0.34    & 0                   & 0\\
 $z_{13}$ & 5.0     & s      & 0              & 0               & 0.06  & 0.06    & 0                   & 0\\
 $z_{14}$ & 5.4     & s      & 0              & 0               & 0.34  & 0.46    & 0                   & 0\\
 $z_{15}$ & 5.0     & v      & $\infty$       & $\infty$        & 1.02  & 1.10    & $\infty$            & $\infty$\\ \hline
 $z_{16}$ & 6.7     & v      & 0              & 0               & 0.68  & 0.60    & 0                   & 0\\
 $z_{17}$ & 5.8     & v      & 0              & 0               & 0.22  & 0.30    & 0                   & 0\\
 $z_{18}$ & 5.5     & s      & 0.50           & 0.50            & 0.44  & 0.56    & 0                   & 0\\
 $z_{19}$ & 5.8     & v      & 0              & 0               & 0.22  & 0.30    & 0                   & 0\\
 $z_{20}$ & 5.4     & s      & 0              & 0               & 0.34  & 0.46    & 0                   & 0\\ \hline
 $z_{21}$ & 5.1     & s      & 0              & 0               & 0.04  & 0.16    & 0                   & 0\\
 $z_{22}$ & 5.7     & v      & 0.50           & 0.50            & 0.32  & 0.40    & 0                   & 0\\
 $z_{23}$ & 4.6     & s      & 0              & 0               & 0.46  & 0.34    & 0                   & 0\\
 $z_{24}$ & 4.6     & s      & 0              & 0               & 0.46  & 0.34    & 0                   & 0\\ \hline
 $z_{25}$ & 6.8       & $\mathrm{s}$ & 13     &                 & 1.74  &         & $\infty$            & \\
 $z_{25}$ & 6.8       & $\mathrm{v}$ &        & 0.077           &       & 0.7     &                     & 0\\ \hline
$p_{\mathrm{s}}$&  & & $0.08$ &        & $0.04$ &           & $0.08$ & \\ 
$p_{\mathrm{v}}$&  & &        & $0.32$ &        & $0.08$    &       & $1$
\end{tabular}
\normalsize
\end{center}
\caption{\label{ta:iris}
         \textbf{Conformal prediction of iris species from sepal length, using three
         different nonconformity measures.}  The data used are 
         sepal length and species for a random 
         sample of $25$ of the 
         $100$ plants measured by Edgar Anderson.  The second column gives $x_i$, the sepal
         length.  The third column gives $y_i$, the species.  The $25$th plant 
         has sepal length $x_{25} = 6.8$, and our task is to predict its species $y_{25}$. 
         For each nonconformity measure, we calculate nonconformity scores under each hypothesis, 
         $y_{25} = \mathrm{s}$ and $y_{25} = \mathrm{v}$.  The $p$-value in each column
         is computed from the $25$ nonconformity scores in that column; it is the fraction 
         of them equal to or larger than the $25$th.  The results from the three nonconformity 
         measures are consistent, inasmuch as the $p$-value for $\mathrm{v}$ is always 
         larger than the $p$-value for $\mathrm{s}$.}
\end{table}

We calculate conformal prediction regions using three different
nonconformity measures:  one based on distance to the nearest 
neighbors, one based on distance to the species average, and 
one based on a support-vector machine.  
Because our evidence is relatively weak,
we do not achieve the high precision with high confidence
that can be achieved in many applications of machine learning
(see, e.g., \S\ref{subsec:seldom}).
But we get a clear view 
of the details of the calculations and the interpretation of 
the results.

\bigskip

\noindent
\textbf{Distance to the nearest neighbor belonging to each species.}
Here we use the 
nonconformity measure~(\ref{eq:nnnm}).  The fourth and fifth
columns of Table~\ref{ta:iris} (labeled NN for nearest neighbor)
give nonconformity scores
$\alpha_i$ obtained
with $y_{25} =\mathrm{s}$ and $y_{25}=\mathrm{v}$, respectively.
In both cases, these scores are given by
\begin{equation}\label{eq:scorenn}
\begin{split}
  \alpha_i &= A(\lbag z_1,\dots,z_{25}\rbag \setminus \lbag z_i \rbag,z_i) \\
  &= 
  \frac
  {
    \min \{
    |x_j - x_i|  :  1 \leq j \leq 25 \; \& \; j \neq i \; \& \; y_j = y_i  \}
  }
  {
    \min \{
    |x_j - x_i|  :  1 \leq j \leq 25 \; \& \; j \neq i \; \& \; y_j \neq y_i  \}
  },
\end{split}
\end{equation}
but for the fourth
column $z_{25} = (6.8,\mathrm{s})$, while for the fifth 
column $z_{25} = (6.8,\mathrm{v})$.

If both the numerator and the denominator in~(\ref{eq:scorenn}) are equal to zero,
we take the ratio also to be zero.  This happens in the case of the first plant, 
for example.  It has the same sepal length, $5.0$, as the $7$th and
$13$th plants, which are \textit{setosa}, and the $15$th plant, which is 
\textit{versicolor}.

Step~3 of the conformal algorithm yields $p_{\mathrm{s}} = 0.08$ and $p_{\mathrm{v}} = 0.32$.
Step~4 tells us that
\begin{itemize}
    \item
      $\mathrm{s}$ is in the $1-\epsilon$ prediction region when $1 - \epsilon > 0.92$, and
    \item
      $\mathrm{v}$ is in the $1-\epsilon$ prediction region when $1 - \epsilon > 0.68$.
\end{itemize}
Here are prediction regions for a few levels of $\epsilon$.
\begin{itemize}
    \item
       $\Gamma^{0.08} = \{\mathrm{v}\}$.
       With $92\%$ confidence, we predict that $y_{25} = \mathrm{v}$. 
    \item
       $\Gamma^{0.05} = \{\mathrm{s},\mathrm{v}\}$.
       If we raise the confidence with which we want to predict $y_{25}$ to $95\%$, 
       the prediction is completely uninformative.
    \item
       $\Gamma^{1/3} = \emptyset$.
       If we lower the confidence to $2/3$, we get a prediction we know is false:
       $y_{25}$ will be in the empty set.
\end{itemize}
In fact, $y_{25} =\mathrm{v}$.  Our $92\%$ prediction is correct.

The fact that we are making a known-to-be-false prediction with $2/3$ confidence
is a signal that the $25$th sepal length, $6.8$, is unusual for either species.
A close look at the nonconformity scores reveals that it is 
being perceived as unusual simply because $2/3$ of the plants have other plants
in the sample with exactly the same sepal length, whereas there is no other 
plant with the sepal length $6.8$.  

In classification problems, it is natural to report the greatest $1-\epsilon$ 
for which $\Gamma^{\epsilon}$ is a single label.
In our example, this produces the statement that we are $92\%$ confident
that $y_{25}$ is $\mathrm{v}$.  
But in order to avoid overconfidence when the object $x_n$ is
unusual, it is wise to report also the largest $\epsilon$ for which 
$\Gamma^{\epsilon}$ is empty.  We call this the \textit{credibility} of 
the prediction (\cite{vovk/gammerman/shafer:2005}, p.~96).
In our example, the prediction 
that $y_{25}$ will be $\mathrm{v}$ has credibility of
only $32\%$.%
\footnote{This notion of credibility is one of the novelties of the 
          theory of conformal prediction.  It is not found in the prior literature on 
          confidence and prediction regions.}

\bigskip

\noindent
\textbf{Distance to the average of each species.}
The nearest-neighbor nonconformity measure, because
it considers only nearby sepal lengths, does not take full advantage of the
fact that a \textit{versicolor} flower typically has longer sepals than
a \textit{setosa} flower.  
We can expect to obtain a more efficient conformal predictor 
(one that produces smaller regions for a given level of confidence) if we 
use a nonconformity measure that takes account of 
average sepal length for the two species.

We use the nonconformity measure $A$ defined by 
\begin{equation}\label{eq:disav}
   A(B,(x,y)) = | \overline{x}_{B \cup \lbag (x,y) \rbag,y} - x |,
\end{equation}
where $\overline{x}_{B,y}$ denotes the average sepal length of all 
plants of species $y$ in the bag $B$, and
$B \cup \lbag z \rbag$ denotes the bag obtained by adding $z$ to $B$.
To test $y_{25} = \mathrm{s}$, we consider the bag
consisting of the $24$ old examples together with $(6.8,\mathrm{s})$, and
we calculate the average sepal lengths for the 
two species in this bag:  $5.06$ for \textit{setosa} and $6.02$ for 
\textit{versicolor}.  Then we use~(\ref{eq:disav}) to calculate the nonconformity scores
shown in the sixth column of Table~\ref{ta:iris}:
$$
   \alpha_i = 
 \begin{cases} 
    |5.06 - x_i|  & \text{if } y_i = \mathrm{s}\\
    |6.02 - x_i|  & \text{if } y_i = \mathrm{v}
 \end{cases}
$$
for $i = 1,\dots,25$, where we take $y_{25}$ to be $\mathrm{s}$.
To test $y_{25} = \mathrm{v}$, we consider the bag
consisting of the $24$ old examples together with $(6.8,\mathrm{v})$, and
we calculate the average sepal lengths for the  
two species in this bag:  $4.94$ for \textit{setosa} and $6.1$ for 
\textit{versicolor}.   Then we use~(\ref{eq:disav}) to calculate the nonconformity scores
shown in the seventh column of Table~\ref{ta:iris}:
$$
   \alpha_i = 
 \begin{cases} 
    |4.94 - x_i| & \text{if } y_i = \mathrm{s}\\
    |6.1  - x_i| & \text{if } y_i = \mathrm{v}
 \end{cases}
$$
for $i = 1,\dots,25$, where we take $y_{25}$ to be $\mathrm{v}$.

We obtain $p_{\mathrm{s}} = 0.04$ and $p_{\mathrm{v}} = 0.08$,
so that
\begin{itemize}
    \item
      $\mathrm{s}$ is in the $1-\epsilon$ prediction region when $1 - \epsilon > 0.96$, and
    \item
      $\mathrm{v}$ is in the $1-\epsilon$ prediction region when $1 - \epsilon > 0.92$.
\end{itemize}
Here are the prediction regions for some different levels of $\epsilon$.
\begin{itemize}
    \item
       $\Gamma^{0.04} = \{\mathrm{v}\}$.
       With $96\%$ confidence, we predict that $y_{25} = \mathrm{v}$. 
    \item
       $\Gamma^{0.03} = \{\mathrm{s},\mathrm{v}\}$.
       If we raise the confidence with which we want to predict $y_{25}$ to $97\%$, 
       the prediction is completely uninformative.
    \item
       $\Gamma^{0.08} = \emptyset$.
       If we lower the confidence to $92\%$, we get a prediction we know is false:
       $y_{25}$ will be in the empty set.
\end{itemize}
In this case, we predict $y_{25} = \mathrm{v}$ with confidence $96\%$  but 
credibility only $8\%$.  The credibility is lower with this nonconformity measure
because it perceives $6.8$ as being even more unusual than the nearest-neighbor 
measure did.  It is unusually far from the average sepal lengths for both species.

\bigskip

\noindent
\textbf{A support-vector machine.}
As Vladimir Vapnik explains on pp.~408--410 of his \textit{Statistical 
Learning Theory} \cite{vapnik:1998}, support-vector machines
grew out of the idea of separating two groups of examples with a hyperplane in a
way that makes as few mistakes as possible---i.e., puts as few examples as possible on the wrong side.
This idea springs to mind when we look at Figure~\ref{fig:iris}.  In this one-dimensional picture,
a hyperplane is a point.  We 
are tempted to separate the \textit{setosa} from the \textit{versicolor} 
with a point between $5.5$ and $5.7$.

\begin{figure}
  \begin{center}
      \epsfig{file=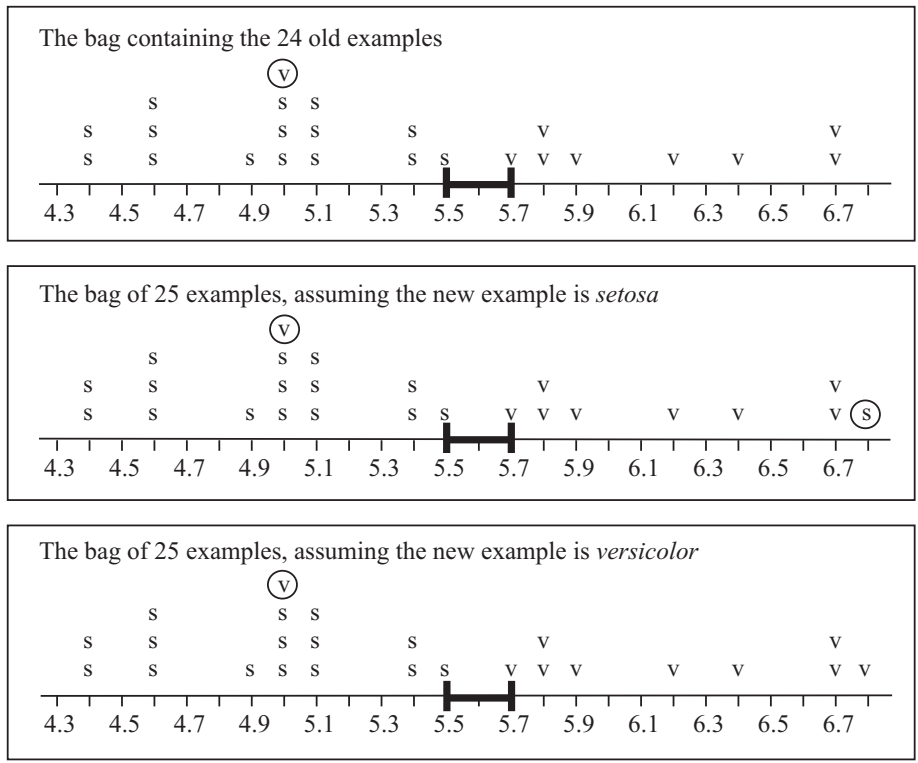,width=\linewidth}
  \end{center}
\caption{\label{fig:iris2} \textbf{Separation for three bags.}  In each case,
                           the separating band is the interval $[5.5,5.7]$.  Examples 
                           on the wrong side of the interval are considered strange and are circled.}
\end{figure}

Vapnik proposed to separate two groups not with a single hyperplane but with a band:  two
hyperplanes with few or no examples between them that separate the two groups as well as possible.
Examples on the wrong side of both hyperplanes
would be considered very strange; those between the hyperplanes
would also be considered strange but less so.  In our one-dimensional example, the obvious separating band is 
the interval from $5.5$ to $5.7$.  The only strange example is the \textit{versicolor} with
sepal length $5.0$.

Here is one way of making Vapnik's idea into an algorithm for calculating 
nonconformity scores for all the examples in a bag $\lbag (x_1,y_1),\dots(x_n,y_n)\rbag$.  
First plot all the examples as in Figure~\ref{fig:iris}.  
Then find numbers $a$ and $b$ such that $a \leq b$ and the interval $[a,b]$
separates the two groups
with the fewest mistakes---i.e., minimizes%
\footnote{Here we are implicitly assuming that the \textit{setosa} flowers will
          be on the left, with shorter sepal lengths.  A general algorithm should also 
          check the possibility
          of a separation with the \textit{versicolor} flowers on the left.}
\begin{equation*}
     \# \{ i | 1\leq i \leq n, x_i < b, \text{ and } y_i = \mathrm{v}\}\\
                          + \# \{ i | 1\leq i \leq n, x_i > a, \text{ and } y_i = \mathrm{s}\}.
\end{equation*}
There may be many intervals that minimize this count; choose one that is widest.
Then give the $i$th example the score
$$ 
    \alpha_i = \left\{
\begin{array}{lll}
     \infty & \text{if  $y_i = \mathrm{v}$  and $x_i < a$}  &\text{or  
                             $y_i = \mathrm{s}$ and $b < x_i$} \\
     1      & \text{if  $y_i = \mathrm{v}$ and $a \leq x_i < b$} &\text{or  
                             $y_i = \mathrm{s}$ and $a < x_i \leq b$}\\
     0      & \text{if  $y_i = \mathrm{v}$ and $b \leq x_i$} &\text{or  
                             $y_i = \mathrm{s}$ and $a \leq x_i$}.
\end{array}
\right.
$$
When applied to the bags in Figure~\ref{fig:iris2}, 
this algorithm gives the circled examples the score $\infty$ and all the others
the score $0$.  These scores are listed
in the last two columns of Table~\ref{ta:iris}.

As we see from the table, the resulting $p$-values 
are $p_{\mathrm{s}}= 0.08$
and $p_{\mathrm{v}}= 1$.  So this time we obtain $92\%$ confidence in 
$y_{25} = \mathrm{v}$, with $100\%$ credibility.

The algorithm just described is too complex to implement when there are 
thousands of examples.  For this reason, Vapnik and his collaborators
proposed instead a quadratic minimization that balances the width
of the separating band against the number and size of the mistakes it makes.  
Support-vector machines of this type have been widely used.  They 
usually solve the dual optimization problem, and 
the Lagrange multipliers they calculate can serve as 
nonconformity scores.  Implementations sometimes fail to treat 
the old examples symmetrically
because they make various uses of the order in which examples are presented,
but this difficulty can be overcome by a preliminary randomization
(\cite{vovk/gammerman/shafer:2005}, p.~58).

\begin{table}
\small
\begin{center}
\begin{tabular}{lccc}
                  & NN             & Species average & SVM            \\ \cline{2-4}
singleton hits    &  164           &  441            &   195          \\
uncertain         &  795           &  477            &   762          \\
total hits        &  959           &  918            &   957          \\ \hline
empty             &    9           &   49            &     1          \\ 
singleton errors  &   32           &   33            &    42          \\
total errors      &   41           &   82            &    43          \\ \hline
total examples    & 1000           & 1000            &  1000          \\ 
\% hits           & $96\%$ & $92\%$  & $96\%$ \\ \hline
total singletons  & 196            & 474             &  237           \\
\% hits           & $84\%$ & $93\%$  & $82\%$ \\ \hline
total errors      & 41             &  82             &  43            \\
\% empty          & $22\%$ & $60\%$  & $2\%$ 
\end{tabular}
\normalsize
\end{center}
\caption{\label{ta:comparison}\textbf{Performance of $92\%$ prediction regions
         based on three nonconformity measures.}
         For each nonconformity measure, we have found $1,000$ prediction regions at the $92\%$
         level, using each time a different 
         random sample of $25$ from Anderson's $100$ flowers.  The ``uncertain'' regions
         are those equal to the whole label space, ${\bf Y}=\{\mathrm{s},\mathrm{v}\}$.}
\end{table}

\bigskip

\noindent
\textbf{A systematic comparison.}
The random sample of $25$ plants we have considered is odd in two ways:  (1) except for 
the one \textit{versicolor} with sepal length of only $5.0$, the two species do not 
overlap in sepal length, and (2) the flower whose species we
are trying to predict has a sepal that is unusually long for either species.

In order to get a fuller picture of how the three nonconformity measures perform
in general on the iris data, we have applied each of them to $1,000$ different 
samples of size $25$ selected from the population of Anderson's $100$ plants.
The results are shown in Table~\ref{ta:comparison}.

The $92\%$ regions based on the species average were correct about $92\%$ of the time
($918$ times out of $1000$),
as advertised.  The regions based on the other two measures were correct more often, 
about $96\%$ of the time.  The reason for this difference is visible in Table~\ref{ta:iris};
the nonconformity scores based on the species average take a greater variety of 
values and therefore produce ties less often.  The regions based
on the species averages are also more efficient (smaller); $447$
of its hits were informative, as opposed to fewer than $200$ for each of the other
two nonconformity measures.  This efficiency also shows up in more empty
regions among the errors.  The species average produced an empty $92\%$ prediction
region for the random sample used in Table~\ref{ta:iris}, and Table~\ref{ta:comparison}
shows that this happens $5\%$ of the time.

As a practical matter, the uncertain prediction regions ($\Gamma^{0.08}=\{\mathrm{s},\mathrm{v}\}$)
and the empty ones ($\Gamma^{0.08}=\emptyset$) are equally uninformative.
The only errors that mislead are the singletons that are wrong, 
and the three methods all produce these at about the same rate---$3$ or $4\%$.

\subsubsection{Example:  Predicting petal width from sepal length}\label{subsubsec:width}

We now turn to the use of the conformal algorithm to predict a number.  
We use the same $25$ plants, 
but now we use the data in the second and third columns
of Table~\ref{ta:iris3}:  the sepal length and petal width for the first
$24$ plants, and the sepal length for the $25$th.  Our task is to 
predict the petal width for the $25$th.

\begin{table}
\begin{center}
\small
\begin{tabular}{ c  c  c  c  c } 
          & sepal length & petal width & Nearest neighbor & Linear regression \\ \cline{2-5}
 $z_1$    & 5.0          & 0.3         &       0.3        & $|0.003y_{25} - 0.149|$   \\
 $z_2$    & 4.4          & 0.2         &       0          & $|0.069y_{25} + 0.050|$   \\
 $z_3$    & 4.9          & 0.2         &       0.25       & $|0.014y_{25} - 0.199|$   \\
 $z_4$    & 4.4          & 0.2         &       0          & $|0.069y_{25} + 0.050|$   \\
 $z_5$    & 5.1          & 0.4         &       0.15       & $|0.008y_{25} + 0.099|$    \\ \hline
 $z_6$    & 5.9          & 1.5         &       0.3        & $|0.096y_{25} - 0.603|$     \\
 $z_7$    & 5.0          & 0.2         &       0.4        & $|0.003y_{25} - 0.249|$      \\
 $z_8$    & 6.4          & 1.3         &       0.2        & $|0.151y_{25} - 0.154|$       \\
 $z_9$    & 6.7          & 1.4         &       0.3        & $|0.184y_{25} - 0.104|$       \\
 $z_{10}$ & 6.2          & 1.5         &       0.2        & $|0.129y_{25} - 0.453|$       \\ \hline
 $z_{11}$ & 5.1          & 0.2         &       0.15       & $|0.008y_{25} + 0.299|$       \\
 $z_{12}$ & 4.6          & 0.2         &       0.05       & $|0.047y_{25} - 0.050|$       \\
 $z_{13}$ & 5.0          & 0.6         &       0.3        & $|0.003y_{25} + 0.151|$       \\
 $z_{14}$ & 5.4          & 0.4         &       0          & $|0.041y_{25} + 0.248|$     \\
 $z_{15}$ & 5.0          & 1.0         &       0.75       & $|0.003y_{25} + 0.551|$      \\ \hline
 $z_{16}$ & 6.7          & 1.7         &       0.3        & $|0.184y_{25} - 0.404|$      \\
 $z_{17}$ & 5.8          & 1.2         &       0.2        & $|0.085y_{25} - 0.353|$      \\
 $z_{18}$ & 5.5          & 0.2         &       0.2        & $|0.052y_{25} + 0.498|$       \\
 $z_{19}$ & 5.8          & 1.0         &       0.2        & $|0.085y_{25} - 0.153|$       \\
 $z_{20}$ & 5.4          & 0.4         &       0          & $|0.041y_{25} + 0.248|$       \\ \hline
 $z_{21}$ & 5.1          & 0.3         &       0          & $|0.008y_{25} + 0.199|$       \\
 $z_{22}$ & 5.7          & 1.3         &       0.2        & $|0.074y_{25} - 0.502|$        \\
 $z_{23}$ & 4.6          & 0.3         &       0.1        & $|0.047y_{25} + 0.050|$        \\
 $z_{24}$ & 4.6          & 0.2         &       0.05       & $|0.047y_{25} - 0.050|$        \\ 
 $z_{25}$ & 6.8          & $y_{25}$    & $|y_{25}-1.55|$  & $|0.805y_{25} - 1.345|$   
\end{tabular}
\normalsize
\end{center}
\caption{\label{ta:iris3}
         \textbf{Conformal prediction of petal width from sepal length.}  We use
         the same random $25$ plants that we used for predicting the species.
         The actual value of $y_{25}$ is $1.4$.}
\end{table}
 
The most conventional
way of analyzing this data is to calculate the least-squares 
line~(\ref{eq:regressionline}):
$$
           \hat{y} = a_{24} + b_{24} x = -2.96 + 0.68 x.
$$
The sepal length for the $25$th plant being $x_{25}=6.8$, the line predicts
that $y_{25}$ should be near $-2.96 + 0.68\times 6.8 = 1.66$.  Under the 
textbook assumption 
that the $y_i$ are all independent and normally distributed with means on the line
and a common variance, 
we estimate the common variance by 
$$
   s_{24}^2 = \frac{\sum_{i=1}^{24} (y_i - (a_{24}+b_{24}x_i))^2}{22} = 0.0780.
$$
The textbook $1-\epsilon$ interval for $y_{25}$
based on $\lbag(x_1,y_1),\dots,(x_{24},y_{24})\rbag$ and $x_{25}$ is
\begin{equation}\label{eq:conventional}
  1.66 \pm t^{\epsilon/2}_{22} s_{24} 
    \sqrt{1 + \frac{1}{24} + \frac{(x_{25}-\overline{x}_{24})^2}{\sum_{j=1}^{24}(x_j - \overline{x}_{24})^2}} 
        =1.66 \pm 0.311 t^{\epsilon/2}_{22} 
\end{equation}
(\cite{draper/smith:1998}, p.~82; \cite{ryan:1997}, pp.~21--22; \cite{seber/lee:2003}, p.~145). 
Taking into account the fact $y_{25}$ is measured to only one decimal place,
we obtain $[1.0,2.3]$ for the $96\%$ interval and $[1.1,2.2]$ for the $92\%$ interval.

The prediction interval~(\ref{eq:conventional}) is analogous to Fisher's interval
for a new example from the same normally distributed population as a bag of old examples
(\S\ref{subsubsec:fisherregion}).
In~\S\ref{subsubsec:gauss} we will review the general model of which
both are special cases.

As we will now see, the conformal algorithm under exchangeability gives confidence intervals
comparable to~(\ref{eq:conventional}), without the assumption that the errors are normal.
We use two different nonconformity measures:  
one based on the nearest neighbor, and one based on the least-squares line.

\bigskip

\noindent
\textbf{Conformal prediction using the nearest neighbor.}
Suppose $B$ is a bag of old examples 
and $(x,y)$ is a new example, for which we know 
the sepal length $x$ but not the petal width $y$.  We can predict $y$
using the nearest neighbor in an obvious way:  We find the $z^{\prime} \in B$
for which the sepal length $x^{\prime}$ is closest to $x$, 
and we predict that $y$ will be the same as the petal width $y^{\prime}$.
If there are several examples in the bag with sepal length equally close 
to $x$, then we take the median of their petal widths as 
our predictor $\hat{y}$. 
The associated nonconformity measure is $|y - \hat{y}|$.

The fourth column of Table~\ref{ta:iris3} gives the nonconformity scores
for our sample using this nonconformity measure.  We see that
$\alpha_{25} = |y_{25} - 1.55|$.  The other nonconformity scores
do not involve $y_{25}$; the largest is $0.75$, and the second largest
is $0.40$.  So we obtain these prediction regions $y_{25}$:
\begin{itemize}
   \item
     The $96\%$ prediction region consists of all the 
     $y$ for which $p_y > 0.04$, which requires that at least one of the 
     other $\alpha_i$ be as large as $\alpha_{25}$, or that 
$
              0.75 \geq |y - 1.55|.
$
     This is the interval $[0.8,2.3]$.  
   \item
     The $92\%$ prediction region consists of all the 
     $y$ for which $p_y > 0.08$, which requires that at least two of the 
     other $\alpha_i$ be as large as $\alpha_{25}$, or that 
$
              0.40 \geq |y - 1.55|.
$
     This is the interval $[1.2,1.9]$.
\end{itemize}

\bigskip

\noindent
\textbf{Conformal prediction using least-squares.}
Now we use the least-squares nonconformity measure with inclusion, 
given by~(\ref{eq:lsnoncon}). In our case, $n=25$, so our nonconformity
scores are
\begin{equation*}
\begin{split}
  \alpha_i   &=    |y_i - (a_{25} + b_{25} x_i)| \\
             &= \left| y_i -  \frac{\sum_{j=1}^{25} y_j}{25}
     - \frac{\sum_{j=1}^{25} (x_j -\overline{x}_{25})y_j}{\sum_{j=1}^{25} (x_j-\overline{x}_{25})^2} 
                           \left(x_i - \frac{\sum_{j=1}^{25} x_j}{25}\right)\right|
\end{split}
\end{equation*}
When we substitute values of 
$\sum_{j=1}^{24}y_j$, $\sum_{j=1}^{24} (x_j-\overline{x}_{25})y_j$, 
$\sum_{j=1}^{25} (x_j-\overline{x}_{25})^2$, 
and $\sum_{j=1}^{25} x_j$ calculated from Table~\ref{ta:iris3}, 
this becomes
$$
         \alpha_i =\left| y_i + \left(0.553-0.110 x_i \right)y_{25} - 0.498x_i + 2.04 \right|. 
$$
For $i=1,\dots,24$, we can further evaluate this by substituting the values 
of $x_i$ and $y_i$.  For $i=25$, we can substitute $6.8$ for
$x_{25}$.  These substitutions produce the expressions 
of the form $|c_i y _{25} + d_i|$ listed
in the last column of Table~\ref{ta:iris3}.  We have made sure
that $c_i$ is always positive by multiplying by $-1$ within the 
absolute value when need be.

Table~\ref{ta:iris4} shows calculations required to find the conformal 
prediction region.  The task is to identify, for $i=1,\dots,24$, 
the $y$ for which
$|c_i y + d_i| \geq |0.805y - 1.345|$.  We first find the 
solutions of the equation $|c_i y + d_i| = |0.805y - 1.345|$, 
which are 
$$
    -\frac{d_i+ 1.345}{c_i-0.805}  \qquad \text{and}  \qquad  -\frac{d_i- 1.345}{c_i+0.805}.
$$
As it happens, $c_i < 0.805$ for $i=1,\dots,24$, and in this case the $y$
satisfying $|c_i y + d_i| \geq |0.805 - 1.345|$ form the interval
between these two points.  This interval is shown in the last column of the table.

In order to be in the $96\%$ interval, $y$ must be in at least one of the $24$
intervals in the table; in order to be in the $92\%$ interval, it must be in at least
two of them.  So the $96\%$ interval is $[1.0,2.4]$, and the 
$92\%$ interval is $[1.0,2.3]$. 

An algorithm for finding conformal prediction intervals using 
a least-squares or ridge-regression nonconformity measure with an object
space of any finite dimension is 
spelled out on pp.~32--33 of \cite{vovk/gammerman/shafer:2005}.

\begin{table}
\begin{center}
\small
\begin{tabular}{ c  c  c  c  c } 
          & & & & \\
          & $\alpha_i = |c_i y_{25} + d_i|$ 
               & $\displaystyle -\frac{d_i+ 1.345}{c_i-0.805}$
                   & $\displaystyle -\frac{d_i- 1.345}{c_i+0.805}$
                       & 
\begin{minipage}{1in}
\begin{center}
    $y$ satisfying 
    $|c_i y + d_i| \geq$
    $ |0.805 - 1.345|$ 
\end{center}
\end{minipage}                                               \\     \hline
 $z_1$    & $|0.003y_{25} - 0.149|$ &1.49 &1.85 & [1.49,1.85] \\
 $z_2$    & $|0.069y_{25} + 0.050|$ &1.90 &1.48 & [1.48,1.90] \\
 $z_3$    & $|0.014y_{25} - 0.199|$ &1.45 &1.89 & [1.45,1.89] \\
 $z_4$    & $|0.069y_{25} + 0.050|$ &1.90 &1.48 & [1.48,1.90] \\
 $z_5$    & $|0.008y_{25} + 0.099|$ &1.81 &1.53 & [1.53,1.81]  \\ \hline
 $z_6$    & $|0.096y_{25} - 0.603|$ &1.05 &2.16 & [1.05,2.16]   \\
 $z_7$    & $|0.003y_{25} - 0.249|$ &1.37 &1.97 & [1.37,1.97]    \\
 $z_8$    & $|0.151y_{25} - 0.154|$ &1.82 &1.57 & [1.57,1.82]     \\
 $z_9$    & $|0.184y_{25} - 0.104|$ &2.00 &1.47 & [1.47,2.00]     \\
 $z_{10}$ & $|0.129y_{25} - 0.453|$ &1.32 &1.93 & [1.32,1.93]     \\ \hline
 $z_{11}$ & $|0.008y_{25} + 0.299|$ &2.06 &1.29 & [1.29,2.06]     \\
 $z_{12}$ & $|0.047y_{25} - 0.050|$ &1.71 &1.64 & [1.64,1.71]     \\
 $z_{13}$ & $|0.003y_{25} + 0.151|$ &1.87 &1.48 & [1.48,1.87]     \\
 $z_{14}$ & $|0.041y_{25} + 0.248|$ &2.09 &1.30 & [1.30,2.09]   \\
 $z_{15}$ & $|0.003y_{25} + 0.551|$ &2.36 &0.98 & [0.98,2.36]    \\ \hline
 $z_{16}$ & $|0.184y_{25} - 0.404|$ &1.52 &1.77 & [1.52,1.77]    \\
 $z_{17}$ & $|0.085y_{25} - 0.353|$ &1.38 &1.91 & [1.38,1.91]    \\
 $z_{18}$ & $|0.052y_{25} + 0.498|$ &2.45 &0.99 & [0.99,2.45]     \\
 $z_{19}$ & $|0.085y_{25} - 0.153|$ &1.66 &1.68 & [1.66,1.68]     \\
 $z_{20}$ & $|0.041y_{25} + 0.248|$ &2.09 &1.30 & [1.30,2.09]     \\ \hline
 $z_{21}$ & $|0.008y_{25} + 0.199|$ &1.94 &1.41 & [1.41,1.94]     \\
 $z_{22}$ & $|0.074y_{25} - 0.502|$ &1.15 &2.10 & [1.15,2.10]      \\
 $z_{23}$ & $|0.047y_{25} + 0.050|$ &1.84 &1.52 & [1.52,1.84]      \\
 $z_{24}$ & $|0.047y_{25} - 0.050|$ &1.71 &1.64 & [1.64,1.71]      \\ \hline
 $z_{25}$ & $|0.805y_{25} - 1.345|$  &  &  &  
\end{tabular}
\normalsize
\end{center}
\caption{\label{ta:iris4}
         \textbf{Calculations with least-squares nonconformity scores.}  The
                 column on the right gives the values of $y$ for which the 
                 example's nonconformity score will exceed that of the $25$th example.}
\end{table}

\bigskip

\begin{table}
\begin{center}
\begin{tabular}{cccc}
                  & Least-squares       &  \multicolumn{2}{c}{Conformal prediction with two}      \\ 
                  & prediction with     &  \multicolumn{2}{c}{different nonconformity measures}   \\ \cline{3-4}
                  & normal errors       &  NN             & Least squares         \\ \hline
$96\%$            &  $[1.0,2.3]$      &  $[0.8,2.3]$  &   $[1.0,2.4]$       \\
$92\%$            &  $[1.1,2.2]$      &  $[1.2,1.9]$  &   $[1.0,2.3]$       
\end{tabular}
\end{center}
\caption{\label{ta:comparison2}\textbf{Prediction intervals for the $25$th plant's
         petal width, calculated by three different methods.}  The conformal prediction 
         intervals using the least-squares nonconformity measure are quite close to 
         the standard intervals based on least-squares with normal errors.  All the 
         intervals contain the actual value, $1.4$.}
\end{table}

\subsection{Optimality}\label{subsec:nootherway}

The predictions produced by the conformal algorithm are 
invariant with respect to the old examples, correct with the 
advertised probability, and nested.
As we now show, they are optimal among all region predictors with these properties.

Here is more precise statement of the three properties:
\begin{enumerate}
    \item
       \textit{The predictions are invariant with respect to the 
       ordering of the old examples.}  Formally, this means that the
       predictor $\gamma$ is a function of two variables, the significance
       level $\epsilon$ and the bag $B$ of old examples.
       We write $\gamma^{\epsilon}(B)$ for the prediction, 
       which is a subset of the example space $\bf Z$.
    \item
       \textit{The probability of a hit is always at least the advertised
       confidence level.}
       For every positive integer $n$ and every
       probability distribution under which 
       $z_1,\dots,z_n$ are exchangeable, 
$$
  \Prob \{ z_n \in \gamma^{\epsilon} (\lbag z_1,\dots,z_{n-1} \rbag) \} \geq 1-\epsilon.
$$
    \item
       \textit{The prediction regions are nested.}  
       If $\epsilon_1 \geq \epsilon_2$, then
       $\gamma^{\epsilon_1}(B) \subseteq \gamma^{\epsilon_2}(B)$.
\end{enumerate}
Conformal predictors satisfy these three conditions.
Other region predictors can also satisfy them.  
But as we now demonstrate, any $\gamma$ satisfying them
can be improved on by a conformal predictor:  there always exists 
a nonconformity measure $A$ such that the predictor $\gamma_A$
constructed from $A$ by the conformal algorithm satisfies
$\gamma_A^{\epsilon}(B) \subseteq \gamma^{\epsilon}(B)$ for all $B$
and $\epsilon$.

The key to the demonstration is the following lemma:
\begin{lemma}
Suppose $\gamma$ is a region predictor
satisfying the three conditions, $\lbag a_1,\dots,a_n \rbag$ 
is a bag of examples,
and $0< \epsilon \leq 1$.  Then $n\epsilon$ or fewer of the $n$ elements
of the bag satisfy
\begin{equation}\label{eq:mistake}
     a_i \notin \gamma^{\epsilon}(\lbag a_1,\dots,a_n  \rbag \setminus \lbag a_i \rbag).
\end{equation}
\end{lemma}
\begin{proof}
Consider the unique exchangeable probability distribution for 
$z_1,\dots,z_n$ that gives probability $1$ to 
$\lbag z_1,\dots,z_n \rbag = \lbag a_1,\dots,a_n \rbag$.
Under this distribution, each element of $\lbag a_1,\dots,a_n \rbag$ 
has an equal probability of being $z_n$, and in this case, 
(\ref{eq:mistake}) is a mistake.  By the second condition, the 
probability of a mistake is $\epsilon$ or less.  So the fraction of the
bag's elements for which~(\ref{eq:mistake}) holds is $\epsilon$ or less.
\qedtext
\end{proof}

Given the region predictor $\gamma$, what nonconformity measure will 
give us a conformal predictor that improves on it?  If 
\begin{equation}\label{eq:exclude}
    z \notin \gamma^{\delta} (B),
\end{equation}
then $\gamma$ is asserting confidence $1-\delta$ that $z$ should 
not appear next because it is so different from $B$.  So the 
largest $1-\delta$ for which~(\ref{eq:exclude}) holds
is a natural nonconformity measure:
$$
   A(B,z) = \sup \{1- \delta \; | \; z \notin \gamma^{\delta} (B) \}.
$$
The conformal predictor $\gamma_A$ obtained from this nonconformity
measure, though it agrees with $\gamma$ on how to rank different $z$ with 
respect to their nonconformity with $B$, may produce tighter
prediction regions if $\gamma$ is too conservative in the levels
of confidence it asserts.

To show that $\gamma_A^{\epsilon}(B) \subseteq \gamma^{\epsilon}(B)$ for 
every $\epsilon$ and every $B$, we assume that 
\begin{equation}\label{eq:in}
z\in \gamma_A^{\epsilon}(\lbag z_1,\dots,z_{n-1} \rbag)
\end{equation}
and show that $z\in \gamma^{\epsilon}(\lbag z_1,\dots,z_{n-1} \rbag)$.
According to the conformal algorithm, (\ref{eq:in}) means that when 
we provisionally set $z_n$ equal to $z$ and calculate the nonconformity scores
$$
    \alpha_i = \sup \{1- \delta \; | \; z_i \notin 
        \gamma^{\delta} (\lbag z_1,\dots,z_n \rbag \setminus \lbag z_i \rbag) \}
$$
for $i=1,\dots,n$, 
we find that strictly more than $n\epsilon$ of these scores
are greater than or equal to $\alpha_n$.  Because $\gamma$'s prediction
regions are nested (condition 3), it follows that if 
$z_n \notin \gamma^{\epsilon}(\lbag z_1,\dots,z_{n-1}\rbag)$, then
$z_i \notin \gamma^{\epsilon} (\lbag z_1,\dots,z_n \rbag \setminus \lbag z_i \rbag)$
for strictly more than $n\epsilon$ of the $z_i$.  But by Lemma~1, 
$n\epsilon$ or fewer of the $z_i$ can satisfy this condition.
So $z_n \in \gamma^{\epsilon}(\lbag z_1,\dots,z_{n-1}\rbag)$.

There are sensible reasons to use
region predictors that are not invariant.
We may want to exploit possible departures from exchangeability 
even while insisting on validity under exchangeability.
Or it may simply be more practical to use a predictor that is
not invariant.  But invariance is a natural condition when we want
to rely only on exchangeability, and in this case 
our optimality result is persuasive.
For further discussion, see~\S2.4 of \cite{vovk/gammerman/shafer:2005}.

\subsection{Examples are seldom exactly exchangeable.}\label{subsec:seldom}

Although the assumption of exchangeability is weak compared to 
the assumptions embodied in most statistical models, it is still an 
idealization, seldom matched exactly by what we see in the world.
So we should not expect conclusions derived from this assumption to be
exactly true.  In particular, we should not be surprised if
a $95\%$ conformal predictor is wrong more than $5\%$ of 
the time.

We can make this point with the USPS dataset so often used to
illustrate machine learning methods.  This dataset consists of 9298 examples of the form $(x,y)$,
where $x$ is a $16 \times 16$ gray-scale matrix and $y$ is one of the ten digits
$0,1,\dots,9$.  It has been used in hundreds of books and articles.  In
\cite{vovk/gammerman/shafer:2005}, it is used to illustrate conformal prediction with
a number of different nonconformity measures.  It is well known that the examples
in this dataset are not perfectly exchangeable.
In particular, the first 7291 examples, which are often 
treated as a training set, are systematically different in some respects from the 
remaining 2007 examples, which are usually treated as a test set.

\begin{figure}
  \begin{center}
      \epsfig{file=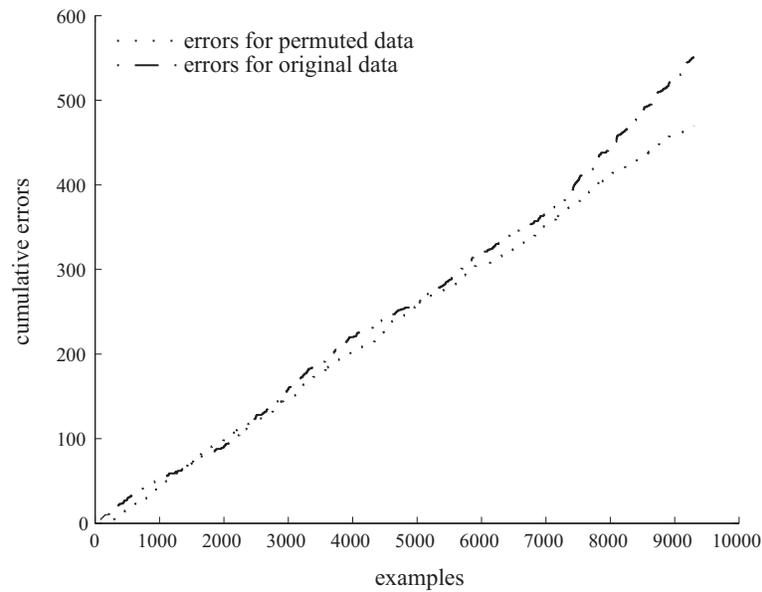,height=8cm}
  \end{center}
\caption{\label{fig:usps}
         \textbf{Errors in $95\%$ nearest-neighbor conformal prediction 
         on the classical USPS dataset.}  When the $9298$ examples are predicted in a
         randomly chosen order, so that the exchangeability assumption is satisfied 
         for sure, the error rate is approximately $5\%$ as 
         advertised.  When they are taken in their original
         order, first the $7291$ in the training set, and then the $2007$
         in the test set, the error rate is higher, especially in the test set.}
\end{figure}

\begin{table}
\small
\begin{center}
\begin{tabular}{lcccccc}
                  & \multicolumn{3}{c}{Original data}        &  \multicolumn{3}{c}{Permuted data}     \\ \cline{2-7}
                  & Training       & Test         & Total      & Training       & Test           & Total \\ \hline
singleton hits    & 6798           & 1838         & 8636       & 6800           & 1905           &  8705 \\
uncertain hits    &  111           &   0          & 111        &   123          &  0             & 123   \\ 
total hits        & 6909           & 1838         &  8747      &    6923        &   1905         &  8828 \\ \hline
empty             & 265            & 142          & 407        & 205            & 81             & 286   \\
singleton errors  & 102            & 27           & 129        & 160            & 21             & 181   \\
uncertain errors   &  15            &   0          &  15        &   3            &  0             &  3    \\
total errors      & 382            & 169          & 551        & 368            & 102            & 470   \\ \hline
total examples    & 7291           & 2007         &  9298      & 7291           & 2007           & 9298  \\ 
\% hits           & $95\%$         & $92\%$       & $94\%$     & $95\%$         & $95\%$         & $95\%$ \\ \hline
total singletons  &  6900          &   1865       & 8765       &   6960         &  1926          & 8880  \\ 
\% hits           & 99\%           &   99\%       & 99\%       &   98\%         &  99\%          & 98\%  \\ \hline
total uncertain   &  126           &   0          & 126        &   126          &  0             & 126   \\ 
\% hits           &  82\%          &              & 82\%       &  98\%          &                & 98\%  \\ \hline
total errors      & 382            & 169          & 551        & 368            & 102            & 470   \\
\% empty          &  69\%          &   85\%       &  74\%      &  57\%          &  79\%          & 61\%  
\end{tabular}
\normalsize
\end{center}
\caption{\label{ta:postal}
         \textbf{Details of the performance of $95\%$ nearest-neighbor conformal prediction 
         on the classical USPS dataset.}  
         Because there are $10$ labels, the uncertain predictions, those containing
         more than one label, can be hits or errors.}
\end{table}

Figure~\ref{fig:usps} illustrates how the non-exchangeability of the USPS data affects
conformal prediction.  The figure records the performance of the $95\%$ conformal 
predictor using the nearest-neighbor nonconformity measure~(\ref{eq:nnnm}),
applied to the USPS data in two ways.  
First we use the 9298 examples in the order in which they are given in the dataset.
(We ignore the distinction between training and test examples, but since the 
training examples are given first we do go through them first.)  Working through the
examples in this order, we predict each $y_n$ using the previous examples and $x_n$.
Second, we randomly permute all 9298 examples, thus producing an order with respect to which the 
examples are necessarily exchangeable.  The law of large numbers works when we go 
through the examples in the permuted order:  we make mistakes at a steady rate,
about equal to the expected $5\%$.
But when we go through the examples in the original order,
the fraction of mistakes is less stable,
and it worsens as we move into the test set.
As Table~\ref{ta:postal} shows, the fraction of mistakes is $5\%$, as desired,
in the first 7291 examples (the training set)
but jumps to $8\%$ in the last 2007 examples.

Non-exchangeability can be tested statistically, using
conventional or game-theoretic methods
(see \S7.1 of \cite{vovk/gammerman/shafer:2005}).  In the case
of this data, any reasonable test will reject exchangeability 
decisively.  Whether the deviation from exchangeability is of practical
importance for prediction depends, of course, on circumstances.
An error rate of $8\%$ when $5\%$ has been promised may or may not 
be acceptable.

\section{On-line compression models}\label{sec:compression}

In this section, we generalize conformal prediction
from the exchangeability model to a whole class of models,
which we call on-line compression models. 

In the exchangeability model, we compress or summarize examples
by omitting information about their order.  We then look backwards
from the summary (the bag of unordered examples) and give
probabilities for the different orderings that could have produced it.
The compression can be done on-line:  each time we see a new
example, we add it to the bag. The backward-looking probabilities
can also be given step by step.
Other on-line compression models compress more or less drastically but 
have a similar structure.

On-line compression models
were studied in the 1970s and 1980s, under various names, by 
Per Martin-L\"of \cite{martin-lof:1974}, 
Steffen Lauritzen \cite{lauritzen:1988},
and Eugene Asarin \cite{asarin:1987,asarin:1988}.
Different authors had different motivations.
Lauritzen and Martin-L\"of started from statistical mechanics, whereas
Asarin started from Kolmogorov's thinking about the meaning of randomness.
But the models they studied all summarize past examples
using statistics that contain 
all the information useful for predicting
future examples.  The summary is
updated each time one observes a new example, and the 
probabilistic content of the structure is 
expressed by Markov kernels that give probabilities for 
summarized examples conditional on the summaries.

In general, a \text{Markov kernel} is a mapping that 
specifies, as a function of one variable,
a probability distribution for some other variable or variables.  
A Markov kernel for $w$ given $u$,
for example, gives a probability distribution for $w$
for each value of $u$.  It is conventional to write $P(w|u)$ for this
distribution.  We are interested in 
Markov kernels of the form $P(z_1,\dots,z_n | \sigma_n)$,
where $\sigma_n$ summarizes the examples 
$z_1,\dots,z_n$.  Such a kernel 
gives probabilities for the different $z_1,\dots,z_n$ that
could have produced $\sigma_n$.

Martin-L\"of, Lauritzen, and Asarin were interested
in justifying widely used statistical models from 
principles that seem less arbitrary than 
the models themselves.  On-line compression
models offer an opportunity to do this, because they typically
limit their use of probability to representing ignorance
with a uniform distribution but 
lead to statistical models that seem to say something more.
Suppose, for example, that Joe
summarizes numbers $z_1,\dots,z_n$ by
$$
     \overline{z} =\frac1n \sum_{i=1}^n z_i  
                 \qquad \text{and} \qquad  
     r^2 = \sum_{i=1}^n (z_i - \overline{z})^2
$$
and gives these summaries to Bill, who 
does not know $z_1,\dots,z_n$.  Bill might adopt a probability
distribution for $z_1,\dots,z_n$ that is uniform over the 
possibilities, which form the 
surface of the $n$-dimensional sphere of radius $r$ 
centered around $(\overline{z},\dots,\overline{z})$.  
As we will see in \S\ref{subsubsec:gauss}, this is an on-line
compression model.  It was shown, by Freedman and Smith
(\cite{vovk/gammerman/shafer:2005}, p.~217)
and then by Lauritzen (\cite{lauritzen:1988}, pp.~238--247),
that if we assume this model
is valid for all $n$, then the distribution of $z_1,z_2,\dots$ must be a mixture
of distributions under which $z_1,z_2,\dots$ are independent and normal
with a common mean and variance.  This is analogous to de Finetti's
theorem, which says that if $z_1,\dots,z_n$ are exchangeable for all $n$,
then the distribution of $z_1,z_2,\dots$ must be a mixture
of distributions under which $z_1,z_2,\dots$ are independent.

For our own part, we are interested in using an on-line compression model
directly for prediction rather than as a step towards 
a model that specifies probabilities 
for examples more fully.  We have already seen how the exchangeability model 
can be used directly for prediction:  we establish a law of large numbers
for backward-looking probabilities (\S\ref{subsec:lln}), and we use it to justify 
confidence in conformal prediction regions (\S\ref{subsec:oldalone}).  
The argument extends to on-line compression models
in general.

For the exchangeability model, conformal prediction is
optimal for obtaining prediction regions (\S\ref{subsec:nootherway}).
No such statement can be made for on-line compression models
in general.  In fact, there are other on-line 
compression models in which conformal prediction
is very inefficient
(\cite{vovk/gammerman/shafer:2005}, p.~220).

After developing the general theory of conformal prediction for on-line
compression models (\S\ref{subsec:definitions} and \S\ref{subsec:cpcompress}),
we consider two examples: the exchangeability-within-label model (\S\ref{subsubsec:mondrian})
and on-line Gaussian linear model (\S\ref{subsubsec:gauss}).

\subsection{Definitions}\label{subsec:definitions}

A more formal look at the exchangeability model will suffice to bring the general
notion of an on-line compression model into focus.  

In the exchangeability model, we summarize examples 
simply by omitting information
about their ordering; the ordered examples are summarized by a
bag containing them.  The backward-looking
probabilities are equally simple; given the bag, 
the different possible orderings all have equal probability, as
if the ordering resulted from drawing the examples 
successively at random from the bag without replacement.
Although this picture is very simple, we can distinguish
four distinct mathematical operations within it:
\begin{enumerate}
  \item
\textit{Summarizing.}  
The examples $z_1,\dots,z_n$ are summarized by the bag 
$\lbag z_1,\dots,z_n \rbag$.  
We can say that the summarization is accomplished by a 
\textit{summarizing function}
$\Sigma_n$ that maps an $n$-tuple of examples $(z_1,\dots,z_n)$ to 
the bag containing these examples:
$$
    \Sigma_n(z_1,\dots,z_n) := \lbag z_1,\dots,z_n\rbag.
$$ 
We write $\sigma_n$ for the summary---i.e., the bag $\lbag z_1,\dots,z_n \rbag$.
  \item
\textit{Updating.}  
The summary can be formed step by step as the examples are observed.  
Once you have the bag containing the
first $n-1$ examples, you just add the $n$th.  This defines
an \textit{updating function} $U_n(\sigma,z)$ that satisfies
$$
    \Sigma_n(z_1,\dots,z_n) = U_n(\Sigma_{n-1}(z_1,\dots,z_{n-1}),z_n).
$$
The top panel in Figure~\ref{fig:stepwise} depicts how the summary $\sigma_n$ is 
built up step by step from $z_1,\dots,z_n$ using the updating functions
$U_1,\dots,U_n$.  First $\sigma_1 = U_1(\Box,z_1)$, where $\Box$ is the 
empty bag.  Then $\sigma_2 = U_2(\sigma_1,z_2)$, and
so on.
  \item
\textit{Looking back all the way.}  Given the bag
$\sigma_n$, the $n!$ different orderings of the elements of the bag
are equally likely, just as they would be if we ordered the contents
of the bag randomly.  As we learned in \S\ref{subsec:backexc},
we can say this with a formula that takes explicit account of the possibility
of repetitions in the bag:  the probability of the event
$
        \{ z_1= a_1,\dots,z_n = a_n\}
$
is 
\begin{equation}\label{eq:markovker}
      P_n(a_1,\dots,a_n | \sigma_n) = 
\begin{cases}
     \frac{n_1! \cdots n_k!}{n!}  & \text{if } \lbag a_1,\dots,a_n \rbag = \sigma_n\\
               0                  & \text{if } \lbag a_1,\dots,a_n \rbag \neq \sigma_n,
\end{cases}
\end{equation}
where $k$ is the number of distinct elements in $\sigma_n$, and $n_1,\dots,n_k$ 
are the numbers of times these distinct elements occur.  We call $P_1,P_2,\dots$
the \textit{full kernels}. 
  \item
\textit{Looking back one step.}  
We can also look back one step.  Given the bag $\sigma_n$, what are the probabilities
for $z_n$ and $\sigma_{n-1}$?  They are the 
same as if we drew $z_n$ out of $\sigma_n$ at random.  In other words, for each $z$
that appears in $\sigma_n$, there is a probability $k/n$, 
where $k$ is the number of times $z$
appears in $\sigma_n$, that (1) $z_n = z$ and (2) $\sigma_{n-1}$ is the bag obtained by removing
one instance of $z$ from $\sigma_n$.  The kernel defined in this way is represented
by the two arrows backward from $\sigma_n$ in the bottom panel of 
Figure~\ref{fig:stepwise}.  Let us designate it by $R_n$.
We similarly obtain a kernel $R_{n-1}$ backward from $\sigma_{n-1}$ and so on. 
These are the \textit{one-step kernels} for the model.
We can obtain the full kernel $P_n$ by combining the 
one-step kernels $R_n,R_{n-1},\dots,R_1$.  This is most readily understood not
in terms of formulas but in terms of a sequence of drawings whose outcomes have
the probability distributions given by the kernels.  The drawing from $\sigma_n$
(which goes by the probabilities given by $R_n(\cdot|\sigma_n)$)
gives us $z_n$ and $\sigma_{n-1}$, the drawing from $\sigma_{n-1}$
(which goes by the probabilities given by $R_{n-1}(\cdot|\sigma_{n-1})$) gives us 
$z_{n-1}$ and $\sigma_{n-2}$, and so on; we finally obtain the whole random sequence
$z_1,\dots,z_n$, which has the distribution $P_n(\cdot|\sigma_n)$.
This is the meaning of the bottom panel in Figure~\ref{fig:stepwise}.
\end{enumerate}
All four operations are important.  The second and fourth, 
updating and looking back one step, can be thought of as the most 
fundamental, because we can derive the other two from them.  Summarization
can be carried out by composing updates, and looking back all the way can be 
carried out by composing one-step look-backs.  Moreover, the
conformal algorithm uses the one-step back probabilities.  
But when we turn to particular on-line compression models, we
will find it initially most convenient to describe them in 
terms of their summarizing functions and full kernels.

\begin{figure} 
\fbox{
\begin{minipage}{\linewidth}
  \begin{center}
    \unitlength=.6mm
    \begin{picture}(170,50)
      \put(10,10){\makebox(0,0)[cc]{$\Box$}}
      \put(40,10){\makebox(0,0)[cc]{$\sigma_1$}}
      \put(70,10){\makebox(0,0)[cc]{$\sigma_2$}}
      \put(100,10){\makebox(0,0)[cc]{$\cdots$}}
      \put(130,10){\makebox(0,0)[cc]{$\sigma_{n-1}$}}
      \put(160,10){\makebox(0,0)[cc]{$\sigma_n$}}
      \put(40,40){\makebox(0,0)[cc]{$z_1$}}
      \put(70,40){\makebox(0,0)[cc]{$z_2$}}
      \put(130,40){\makebox(0,0)[cc]{$z_{n-1}$}}
      \put(160,40){\makebox(0,0)[cc]{$z_n$}}
      \put(15,10){\vector(1,0){20}}
      \put(45,10){\vector(1,0){20}}
      \put(75,10){\vector(1,0){18}}
      \put(107,10){\vector(1,0){14}}
      \put(138,10){\vector(1,0){16}}
      \put(40,35){\vector(0,-1){20}}
      \put(70,35){\vector(0,-1){20}}
      \put(130,35){\vector(0,-1){20}}
      \put(160,35){\vector(0,-1){20}}
    \end{picture}
  \end{center}
\textit{Updating.}
         We speak of ``on-line'' compression models because the summary
         can be updated with each new example.  In the case of the exchangeability
         model, we obtain the bag $\sigma_i$ by adding the new example $z_i$
         to the old bag $\sigma_{i-1}$.
\end{minipage}}

\vspace{0.3cm}

\fbox{
\begin{minipage}{\linewidth}
  \begin{center}
    \unitlength=.6mm
    \begin{picture}(170,50)
      \put(10,10){\makebox(0,0)[cc]{$\Box$}}
      \put(40,10){\makebox(0,0)[cc]{$\sigma_1$}}
      \put(70,10){\makebox(0,0)[cc]{$\sigma_2$}}
      \put(100,10){\makebox(0,0)[cc]{$\cdots$}}
      \put(130,10){\makebox(0,0)[cc]{$\sigma_{n-1}$}}
      \put(160,10){\makebox(0,0)[cc]{$\sigma_n$}}
      \put(40,40){\makebox(0,0)[cc]{$z_1$}}
      \put(70,40){\makebox(0,0)[cc]{$z_2$}}
      \put(130,40){\makebox(0,0)[cc]{$z_{n-1}$}}
      \put(160,40){\makebox(0,0)[cc]{$z_n$}}
      \put(35,10){\vector(-1,0){20}}
      \put(65,10){\vector(-1,0){20}}
      \put(93,10){\vector(-1,0){18}}
      \put(121,10){\vector(-1,0){14}}
      \put(154,10){\vector(-1,0){16}}
      \put(40,15){\vector(0,1){20}}
      \put(70,15){\vector(0,1){20}}
      \put(130,15){\vector(0,1){20}}
      \put(160,15){\vector(0,1){20}}
    \end{picture}
  \end{center}
\textit{Backward probabilities.}
         The two arrows backwards from $\sigma_i$ symbolize our probabilities,
         conditional on $\sigma_i$, for what example $z_i$ and what previous summary $\sigma_{i-1}$
         were combined to produce $\sigma_i$.  
         Like the diagram in Figure~\ref{fig:exchangeonestep} that it generalizes, this diagram is 
         a Bayes net.
\end{minipage}}
\caption{\label{fig:stepwise}\textbf{Elements of an on-line compression model.}
         The top diagram represents the updating functions $U_1,\dots,U_n$.  The bottom
         diagram represents the one-step kernels $R_1,\dots,R_n$.}
\end{figure}

In general, an \textit{on-line compression model} for an example 
space $\bf Z$ consists of a space $\bf S$, whose elements
we call \textit{summaries}, and two sequences of mappings:
\begin{itemize}
  \item
     Updating functions $U_1,U_2,\dots$.  The function $U_n$ maps
     a summary $s$ and an example $z$ to a new summary $U_n(s,z)$.
  \item
     One-step kernels $R_1,R_2,\dots$.  For each summary $s$,
     the kernel $R_n$ gives a joint probability distribution 
     $R_n(s^{\prime},z|s)$ for an unknown summary $s^{\prime}$ and unknown example $z$.
     We require that $R_n(\cdot|s)$ give probability one to the set of 
     pairs $(s^{\prime},z)$ such that $U_{n-1}(s^{\prime},z)=s$.  
\end{itemize}
We also require that the summary space $\bf S$ include the empty
summary $\Box$.

The recipes for constructing the
summarizing functions $\Sigma_1,\Sigma_2,\dots$ and the full 
kernels $P_1,P_2,\dots$ are the same in general as in the
exchangeability model:
\begin{itemize}
  \item
     The summary $\sigma_n=\Sigma_n(z_1,\dots,z_n)$ is 
     built up step by step from $z_1,\dots,z_n$ using the updating functions.
     First $\sigma_1 = U_1(\Box,z_1)$, then $\sigma_2 = U_2(\sigma_1,z_2)$, and
     so on.
  \item
     We obtain the full kernel $P_n$ by combining, backwards
     from $\sigma_n$, the random experiments represented by the 
     one-step kernels $R_n,R_{n-1},\dots,R_1$.  First we 
     draw  $z_n$ and $\sigma_{n-1}$ from $R_n(\cdot|\sigma_n)$, then we draw
     $z_{n-1}$ and $\sigma_{n-2}$ from $R_{n-1}(\cdot|\sigma_{n-1})$, and so on.
     The sequence $z_1,\dots,z_n$ obtained in this way has the distribution
     $P_n(\cdot|\sigma_n)$.
\end{itemize}

On-line compression models are usually initially specified in terms of 
their summarizing functions $\Sigma_n$ and their
full kernels $P_n$, because these are usually easy to describe.
One must then verify that these easily described objects do define an on-line
compression model.  This requires verifying two points:
\begin{enumerate}
  \item
     $\Sigma_1,\Sigma_2,\dots$
     can be defined successively by means of updating functions:
\begin{equation}\label{eq:updating}
    \Sigma_n(z_1,\dots,z_n) = U_n(\Sigma_{n-1}(z_1,\dots,z_{n-1}),z_n).
\end{equation}
     In words: $\sigma_n$ depends on $z_1,\dots,z_{n-1}$ only through 
     the earlier summary $\sigma_{n-1}$.  
  \item
     Each $P_n$ can be obtained as required
     using one-step kernels.  One way to verify this is to exhibit 
     the one-step kernels $R_1,\dots,R_n$
     and then to check that drawing $z_n$ and $\sigma_{n-1}$ from $R_n(\cdot|\sigma_n)$, 
     then drawing
     $z_{n-1}$ and $\sigma_{n-2}$ from $R_{n-1}(\cdot|\sigma_{n-1})$, and so on
     produces a 
     sequence $z_1,\dots,z_n$ with the distribution
     $P_n(\cdot|\sigma_n)$.
     Another way to verify it, without necessarily exhibiting the one-step 
     kernels, is to verify the conditional independence relations 
     represented by Figure~\ref{fig:stepwise}:
     $z_n$ (and hence also $\sigma_n$) is probabilistically independent of 
     $z_1,\dots,z_{n-1}$ given $\sigma_{n-1}$.
\end{enumerate}

\subsection{Conformal prediction}\label{subsec:cpcompress}

In the context of an on-line compression model, a \textit{nonconformity measure}
is an arbitrary real-valued function
$
    A(\sigma,z),
$
where $\sigma$ is a summary and $z$ is an example.
We choose $A$ so that $A(\sigma,z)$ is large
when $z$ seems very different from the examples that might be summarized by $\sigma$.

In order to state the conformal algorithm, we write $\tilde{\sigma}_{n-1}$ and $\tilde{z}_n$ for random 
variables with a joint probability distribution given by the one-step kernel
$R(\cdot|\sigma_n)$.  The algorithm using old examples alone can then be stated as follows:

\bigskip

\noindent
\fbox{
\begin{minipage}{\linewidth}
\noindent
\textbf{The Conformal Algorithm Using Old Examples Alone}

\smallskip
\noindent
\textit{Input:}  
Nonconformity measure $A$, significance level $\epsilon$,
examples $z_1,\dots,z_{n-1}$, example $z$

\smallskip
\noindent
\textit{Task:}  Decide whether to include $z$ in $\gamma^{\epsilon}(z_1,\dots,z_{n-1})$.

\smallskip
\noindent
\textit{Algorithm:}
\begin{enumerate}
  \item
     Provisionally set $z_n := z$.
  \item
     Set
     $\displaystyle
        p_z := R_n(A(\tilde{\sigma}_{n-1},\tilde{z}_n) \ge A(\sigma_{n-1},z_n) | \sigma_n).
     $
   \item
     Include $z$ in $\gamma^{\epsilon}(z_1,\dots,z_{n-1})$ 
     if and only if $p_z > \epsilon$.
\end{enumerate}
\end{minipage}}

\bigskip

To see that this reduces to the algorithm we gave for the exchangeability
model on p.~\pageref{p:algorithm}, recall that 
$\sigma_n = \lbag z_,\dots,z_n \rbag$ and
$\tilde{\sigma}_{n-1} = \lbag z_1,\dots,z_n \rbag \setminus \lbag \tilde{z}_n \rbag$
in that model, so that
\begin{equation}\label{eq:nonconformspecial}
   A(\tilde{\sigma}_{n-1},\tilde{z}_n) 
     = A(\lbag z_1,\dots,z_n \rbag \setminus \lbag \tilde{z}_n \rbag,\tilde{z}_n) 
\end{equation}
and 
\begin{equation}\label{eq:nonconformspecial2}
   A(\sigma_{n-1},z_n) 
     = A(\lbag z_1,\dots,z_{n-1} \rbag,z_n)
\end{equation}
Under $R_n(\cdot \givn \lbag z_1,\dots,z_n \rbag)$, the random variable $\tilde{z}_n$
has equal chances of being any of the $z_i$, so that the 
probability of~(\ref{eq:nonconformspecial}) being greater than or 
equal to~(\ref{eq:nonconformspecial2}) 
is simply the 
fraction of the $z_i$ for which
$$
   A(\lbag z_1,\dots,z_n \rbag \setminus \lbag z_i \rbag,z_i)
       \geq A(\lbag z_1,\dots,z_{n-1} \rbag,z_n),
$$
and this is how $p_z$ is defined on p.~\pageref{p:algorithm}.

Our arguments
for the validity of the regions $\gamma^{\epsilon}(z_1,\dots,z_{n-1})$
in the exchangeability model generalize readily.
The definitions of $n$-event and $\epsilon$-rare generalize in 
an obvious way:
\begin{itemize}
  \item
    An event $E$ is an \textit{$n$-event} if its happening or failing is 
    determined by the value of $z_n$ and the value of the 
    summary $\sigma_{n-1}$.\label{p:compressrare}
  \item
    An $n$-event $E$ is $\epsilon$-\textit{rare} if
    $R_n ( E \givn \sigma_n) \leq \epsilon$.
\end{itemize}
The event $z_n \notin \gamma^{\epsilon}(z_1,\dots,z_{n-1})$
is an $n$-event, and it is 
$\epsilon$-rare (the probability is $\epsilon$ or less
that a random variable will take a value that it equals or exceeds
with a probability of $\epsilon$ or less). 
So working backwards from the summary
$\sigma_N$ for a large value of $N$, Bill can still bet against the errors successively
at rates corresponding to their probabilities under $\sigma_n$,
which are always $\epsilon$ or less.
This produces an exact analog to
Informal Proposition~\ref{inprop:llnex}:
\begin{inproposition}\label{inprop:second}
Suppose $N$ is large, and the variables $z_1,\dots,z_N$ obey an on-line compression model.
Suppose $E_n$ is an $\epsilon$-rare $n$-event
for $n=1,\dots,N$.
Then the law of large numbers applies;
with very high probability, no more than approximately the fraction
$\epsilon$ of the events $E_1,\dots,E_N$ will
happen. 
\end{inproposition}

The conformal algorithm using features of the new example generalizes similarly:

\bigskip

\noindent
\fbox{
\begin{minipage}{\linewidth}
\noindent
\textbf{The Conformal Algorithm}

\smallskip
\noindent
\textit{Input:}  
Nonconformity measure $A$, significance level $\epsilon$
examples $z_1,\dots,z_{n-1}$, object $x_n$, label $y$

\smallskip
\noindent
\textit{Task:}  Decide whether to include $y$ in $\Gamma^{\epsilon}(z_1,\dots,z_{n-1},x_n)$.

\smallskip
\noindent
\textit{Algorithm:}
\begin{enumerate}
  \item
     Provisionally set $z_n :=(x_n,y)$.
  \item
     Set
     $\displaystyle
        p_y := R_n(A(\tilde{\sigma}_{n-1},\tilde{z}_n) \ge A(\sigma_{n-1},z_n) | \sigma_n).
     $
   \item
     Include $y$ in $\Gamma^{\epsilon}(z_1,\dots,z_{n-1},x_n)$ 
     if and only if $p_y > \epsilon$.
\end{enumerate}
\end{minipage}}

\bigskip

\noindent
The validity of this algorithm follows from the validity of the algorithm using 
old examples alone by the same argument as in the case of exchangeability.

\subsection{Examples}\label{subsec:compressexamples}

We now look at two on-line compression 
models:  the exchangeability-within-label model
and the on-line Gaussian linear model.  

The exchangeability-within-label model was first introduced in work leading up
to our monograph~\cite{vovk/gammerman/shafer:2005}.  It weakens the assumption of
exchangeability.

The on-line Gaussian linear model, as we have already mentioned, has been widely
studied.  It overlaps the exchangeability model, in the sense that 
the assumptions for both of the models can hold at the same time, but the assumptions
for one of them can hold without the assumptions for the other holding.
It is closely related to the classical Gaussian linear model.
Conformal prediction in the on-line model leads to the same prediction regions 
that are usually used for the classical model.
But the conformal prediction theory adds the new information
that these intervals are valid in the sense of this article:  they are right 
$1-\epsilon$ of the time when used on accumulating data.

\subsubsection{The exchangeability-within-label model}\label{subsubsec:mondrian}

The assumption of exchangeability can be weakened in many ways.  
In the case of classification, one interesting
possibility is to assume only that the 
examples for each label are exchangeable with each other.  
For each label, the objects with that label are as likely to appear in one
order as in another.  This assumption leaves open the possibility that the 
appearance of one label might change the probabilities for the next label.

Suppose the label space has $k$ elements, say ${\bf Y} = \{1,\dots,k\}$.
Then we can define the \textit{exchangeability-within-label model}
as follows:
\begin{description}
  \item[Summarizing Functions]
     The $n$th summarizing function is
\begin{equation}\label{eq:within}
  \Sigma_n(z_1,\dots,z_n)
  :=
  \left(
    y_1,\dots,y_n,B_1^n,\dots,B_k^n
  \right),
\end{equation}
      where $B_j^n$ is the bag consisting of the objects in the list
      $x_1,\dots,x_n$ that have the label $j$.
   \item[Full Kernels]
      The full kernel $P_n(z_1,\dots,z_n\givn y_1,\dots,y_n,B_1^n,\dots,B_k^n)$ 
      is most easily described
      in terms the random action for which it gives the probabilities:  
      independently for each label $j$, 
      distribute the objects in the bag $B_j^n$ randomly among the positions $i$ for
      which $y_i$ is equal to $j$.
\end{description}

To check that this is an on-line compression model, we exhibit the updating 
function and the one-step kernels:
\begin{description}
  \item[Updating]
     When $(x_n,y_n)$ is observed, the summary 
$$
        (y_1,\dots,y_{n-1},B_1^{n-1},\dots,B_k^{n-1})
$$ 
     is updated
     by inserting $y_n$ after $y_{n-1}$ and adding $x_n$ to $B_{y_n}^{n-1}$.  
  \item[One step back]
     The one-step kernel $R_n$ is given by
$$
 R_n(\text{summary},(x,y) \givn y_1,\dots,y_n,B_1^n,\dots,B_k^n) =
\begin{cases}
   \frac{k}{|B_{y_n}^n|} & \text{if } y=y_n\\
               0         & \text{otherwise},
\end{cases}
$$
     where $k$ is the number of $x$s in $B_{y_n}^n$.  
     This is the same as the probability the one-step kernel for the 
     exchangeability model without objects would give for $x$
     on the basis of a bag of size $|B_{y_n}^n|$ that includes $k$ $x$s.
\end{description}

\begin{figure}
  \begin{center}
      \epsfig{file=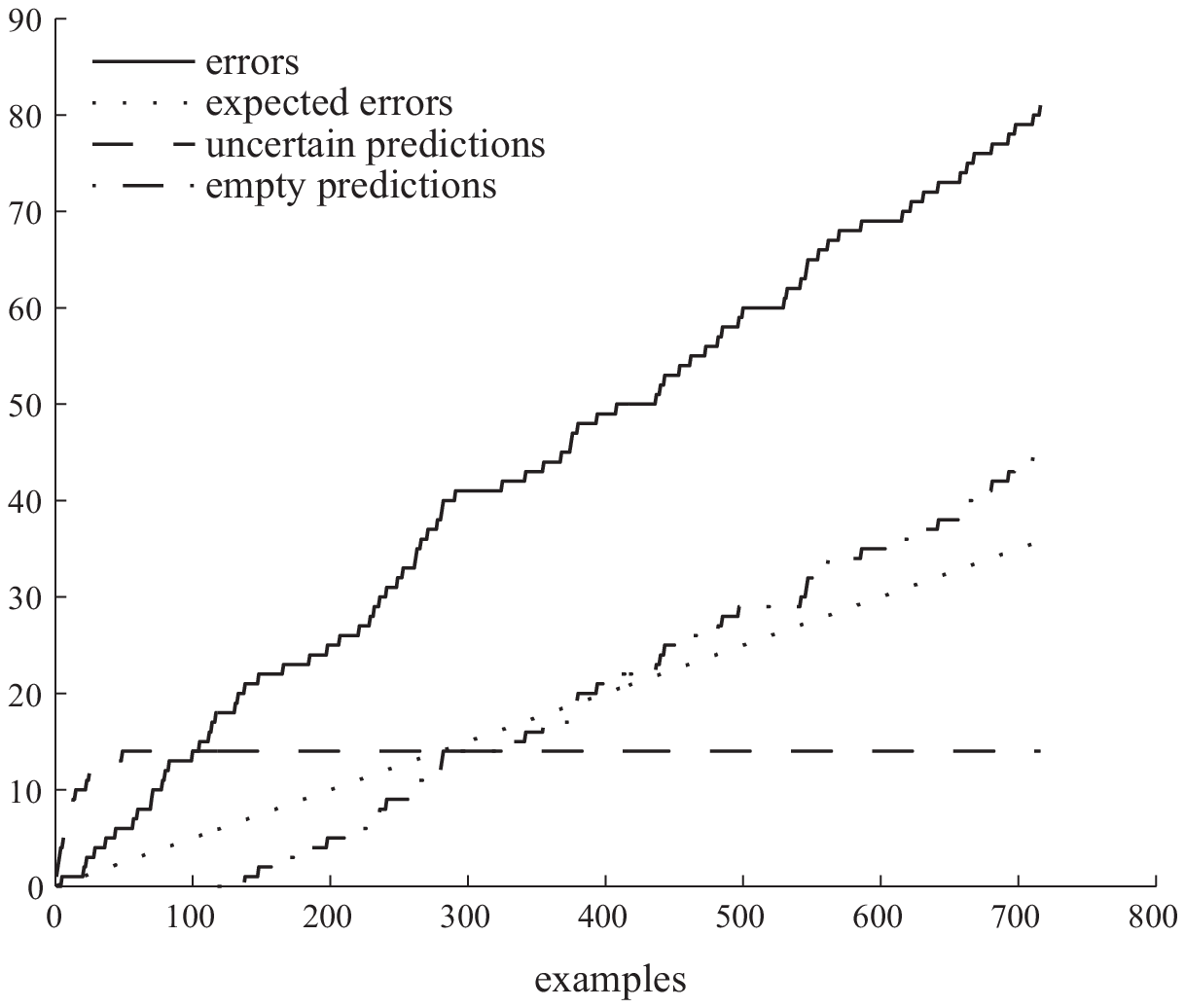,height=7cm}

      \textit{Exchangeability model}
  \end{center}

\vspace{0.2cm}

  \begin{center}
      \epsfig{file=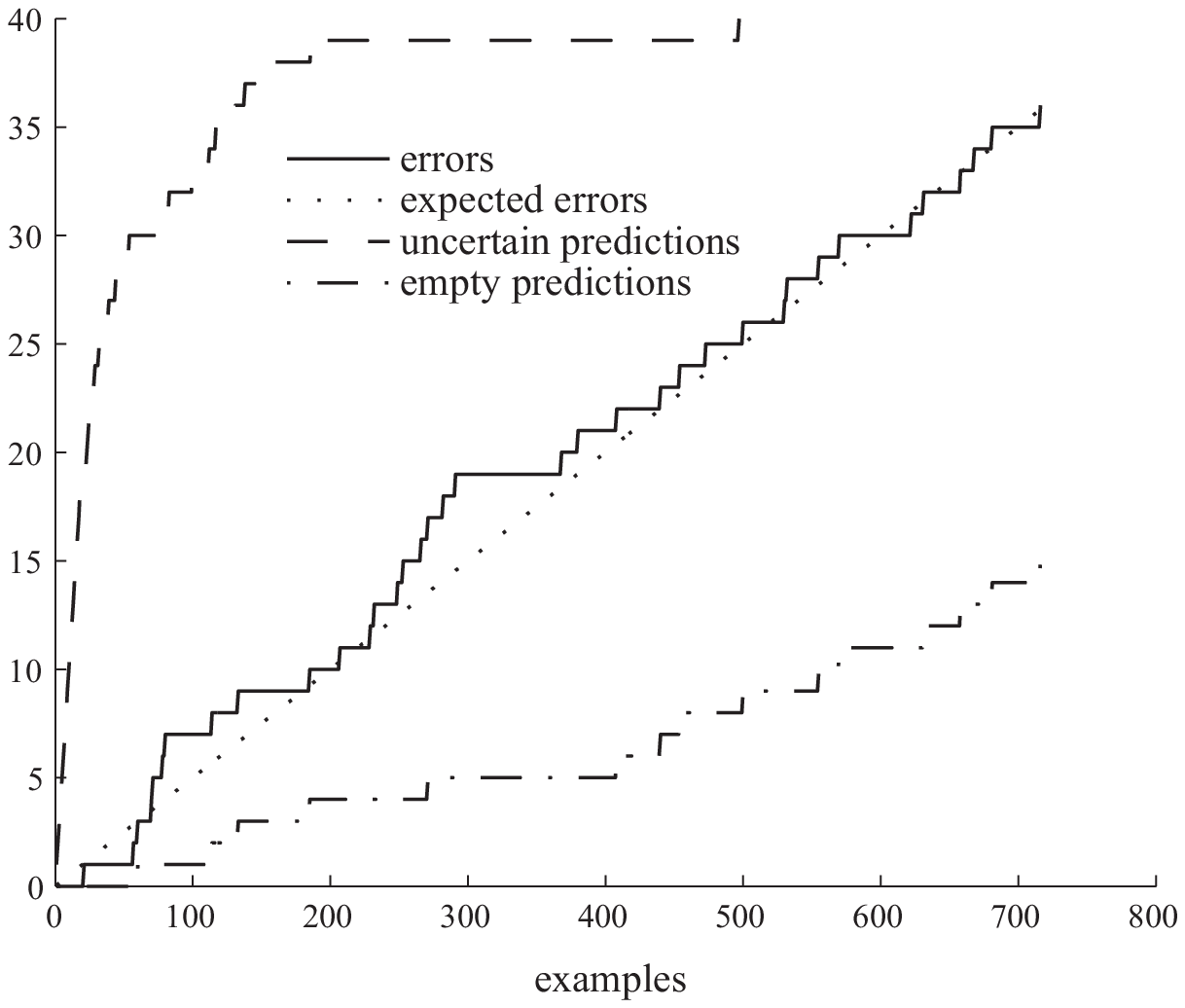,height=7cm}

      \textit{Exchangeability-within-label model}  
  \end{center}

\caption{\label{fig:withinlabel}\textbf{Errors for $95\%$ conformal prediction
        using nearest neighbors
        in the permuted USPS data when the true label is $5$.}  In both figures,
        the dotted line represents the overall expected error rate of $5\%$.  The 
        actual error rate for $5$s with the exchangeability-within-label
        model tracks this line, 
        but with the exchangeability model it is much higher. The 
        exchangeability-within-label predictor keeps its error rate down by issuing more 
        prediction regions containing more than one digit (``uncertain predictions'').}
\end{figure}

Because the true labels are part of the summary, our imaginary bettor Bill can
choose to bet just on those rounds of his game with Joe where the label has 
a particular value, and this implies that a $95\%$ conformal predictor
under the exchangeability-within-label model will make errors at no more than
a $5\%$ rate for examples with that label.
This is not necessarily true for a $95\%$ conformal predictor under the exchangeability model;
although it can make errors
no more than about $5\%$ of the time overall, its error rate may be
higher for some labels and lower for others.  
As Figure~\ref{fig:withinlabel} shows, this happens in the case of 
the USPS dataset.  The graph in the top panel of the figure shows the cumulative errors
for examples with the label 5, which is particularly easy to confuse with other digits, 
when the nearest-neighbor conformal predictor 
is applied to that data in permuted form.  The error rate for 5 is over $11\%$.
The graph in the bottom panel shows the results of the exchangeability-within-label
conformal predictor using the same nearest-neighbor nonconformity measure;
here the error rate stays close to $5\%$.  As this graph makes clear, the 
predictor holds the error rate down to $5\%$ in this case by producing many 
prediction regions containing more than one label (``uncertain predictions'').

As we explain in \S 4.5 and \S 8.4 of \cite{vovk/gammerman/shafer:2005},
the exchangeability-within-label model is a \textit{Mondrian model}.
In general, a Mondrian model decomposes the space ${\bf Z} \times \bbbn$, 
where $\bbbn$ is set of the natural numbers, into non-overlapping rectangles,
and it asks for exchangeability only within these rectangles.
For each example $z_i$, it then records,
as part of the summary, the rectangle into which $(z_i,i)$ falls.  
Mondrian models can be useful when we need to weaken
the assumption of exchangeability.  They can also be 
attractive even if we are willing to assume exchangeability across the categories,
because the conformal predictions they produce will be calibrated within 
categories.

\subsubsection{The on-line Gaussian linear model}\label{subsubsec:gauss}

Consider
examples $z_1,\dots,z_N$, of the form $z_n = (x_n,y_n)$, where $y_n$ is
a number and $x_n$ is a row vector consisting of $p$ numbers.  
For each $n$ between $1$ and $N$, set
$$
        X_n := \left[
    \begin{array}{c}
        x_1\\
       \vdots  \\
        x_n
    \end{array}  \right]
                        \qquad \text{and} \qquad
        Y_n := \left[
    \begin{array}{c}
        y_1\\
       \vdots  \\
        y_n
    \end{array}  \right].
$$
Thus $X_n$ is an $n \times p$ matrix, and $Y_n$ is a column vector of length $n$.

In this context, the \textit{on-line Gaussian linear model} is the on-line compression model
defined by the following summarizing functions and full kernels:
\begin{description}
  \item[Summarizing Functions]
     The $n$th summarizing function is
\begin{equation}\label{eq:gausssum}
\begin{split}
  \Sigma_n(z_1,\dots,z_n)
  :&=
  \left(
    x_1,\dots,x_n,
    \sum_{i=1}^n y_ix_i,
    \sum_{i=1}^n y_i^2
  \right)\\
   &=
  \left(
    X_n,
    X_n^{\prime}Y_n,
    Y_n^{\prime}Y_n
  \right).
\end{split}
\end{equation}
   \item[Full Kernels]
      The full kernel $P_n(z_1,\dots,z_n\givn\sigma_n)$ distributes its
      probability uniformly over the space of vectors $(y_1,\dots,y_n)$
      consistent with the summary $\sigma_n$.  (We consider probabilities
      only for $y_1,\dots,y_n$, because $x_1,\dots,x_n$ are fixed 
      by $\sigma_n$.)
\end{description}
We can write
$
     \sigma_n = (X_n,C,r^2),
$
where $C$ is a column vector of length $p$, and $r$ is a nonnegative number.
A vector $(y_1,\dots,y_n)$ is consistent with $\sigma_n$ if
$$
  \sum_{j=1}^n y_jx_j = C   \qquad \text{and} \qquad  \sum_{j=1}^n y_j^2 = r^2.
$$
This is the intersection of a hyperplane with the surface of a sphere.
Not being empty, the intersection is either a point (in the exceptional case where the
hyperplane is tangent to the sphere) or the surface of a lower-dimensional sphere.
(Imagine intersecting a plane and the surface of a $3$-dimensional sphere;
the result is a circle, the surface of a $2$-dimensional sphere.)
The kernel $P_n(\cdot\givn\sigma_n)$ puts all its probability on the point
or distributes it uniformly over the surface of the lower-dimensional sphere.

To see that the summarizing functions and full kernels define an on-line
compression model, we must check that the summaries can be updated and that
the full kernels have the required conditional independence property:
conditioning $P_n(\cdot\givn\sigma_n)$
on $z_{i+1},\dots,z_n$ gives $P_i(\cdot\givn\sigma_i)$.
(We do not condition on $\sigma_i$
since it can be computed from $z_{i+1},\dots,z_n$ and $\sigma_n$.)
Updating is straightforward; when we observe $(x_n,y_n)$, we update the summary 
$$
   \left( x_1,\dots,x_{n-1}, \sum_{i=1}^{n-1} y_ix_i, \sum_{i=1}^{n-1} y_i^2 \right)
$$
by inserting $x_n$ after $x_{n-1}$ and adding a term to each of the sums.
To see that conditioning $P_n(\cdot\givn\sigma_n)$
on $z_{i+1},\dots,z_n$ gives $P_i(\cdot\givn\sigma_i)$, we note
that conditioning the uniform distribution on the surface of a sphere 
on values $y_{i+1}=a_{i+1},\dots,y_n=a_n$ involves intersecting the surface
with the hyperplanes
defined by these $n-i$ equations.  This produces
the uniform distribution on the surface of the possibly 
lower-dimensional sphere defined by 
$$
   \sum_{j=1}^i y_j^2 = r^2 - \sum_{j=i+1}^n y_j^2   
      \qquad \text{and} \qquad 
   \sum_{j=1}^i y_jx_j = C -  \sum_{j=i+1}^n y_jx_j;
$$
this is indeed $P_i(y_1,\dots,y_i\givn\sigma_i)$.

The on-line Gaussian linear model is closely related to the 
\textit{classical Gaussian linear model}.
In the classical model,%
\footnote{There are many names for the classical model.  The name
          ``classical Gaussian linear model'' is used by Bickel 
          and Doksum \cite{bickel/doksum:2001}, p.~366.}
\begin{equation}\label{eq:classicalglm}
     y_i = x_i \beta + e_i,
\end{equation}
where the $x_i$ are row vectors of known numbers,
$\beta$ is a column vector of unknown
numbers (the regression coefficients), and the $e_i$ are independent 
of each other and normally
distributed with mean zero and a common variance.
When $n -1 > p$ and $\mathrm{Rank}(X_{n-1}) = p$, the theory of the classical 
model tells us the following:
\begin{itemize}
   \item
     After observing examples $(x_1,y_1,\dots,x_{n-1},y_{n-1})$, 
     estimate the vector of coefficients $\beta$ by 
$$
      \hat{\beta}_{n-1} := (X_{n-1}^{\prime}X_{n-1})^{-1}X_{n-1}^{\prime}Y_{n-1} 
$$
     and after further observing $x_n$, predict $y_n$ by
$$
       \hat{y}_n :=  x_n \hat{\beta}_{n-1} 
              = x_n (X_{n-1}^{\prime}X_{n-1})^{-1}X_{n-1}^{\prime}Y_{n-1}.
$$
   \item
      Estimate the variance of the $e_i$ by 
$$
        s_{n-1}^2 := \frac{Y_{n-1}^{\prime}Y_{n-1} - \beta_{n-1}^{\prime} X_{n-1}^{\prime}Y_{n-1}}{n-p-1}.
$$
   \item
      The random variable
\begin{equation}\label{eq:definitiont}
         t_n := \frac{y_n - \hat{y}_n}{s_{n-1} \sqrt{1+ x_n^{\prime} (X_{n-1}^{\prime}X_{n-1})^{-1}x_n}}
\end{equation}
      has a $t$-distribution with $n-p-1$ degrees of freedom, and so 
\begin{equation}\label{eq:generalt}
    \hat{y}_n \pm t^{\epsilon/2}_{n-p-1} s_{n-1} \sqrt{1+ x_n^{\prime} (X_{n-1}^{\prime}X_{n-1})^{-1}x_n}
\end{equation}
      has probability $1-\epsilon$ of containing $y_n$
      (\cite{ryan:1997}, p.~127; \cite{seber/lee:2003}, p.~132).  
\end{itemize}
The assumption $\mathrm{Rank}(X_{n-1}) = p$ can be relaxed, at the price of complicating
the formulas involving $(X_{n-1}^{\prime}X_{n-1})^{-1}$.  But the assumption
$n-1 > \mathrm{Rank}(X_{n-1})$ is essential to finding a prediction 
interval of the type~(\ref{eq:generalt}); when it fails there are values for the
coefficients $\beta$ such that $y_{n-1}=X_{n-1}\beta$, and consequently there is
no residual variance with which to estimate the variance of the $e_i$.

We have already used two special cases of~(\ref{eq:generalt}) in this article.  
Formula~(\ref{eq:fisherinterval}) 
in \S\ref{subsubsec:fisherregion} is the special 
case with $p=1$
and each $x_i$ equal to $1$, and formula~(\ref{eq:conventional}) at the 
beginning of \S\ref{subsubsec:width} is the special case with $p=2$ and the first entry of 
each $x_i$ equal to $1$.

The relation between the classical and on-line models, 
fully understood in the theoretical literature since the 1980s, 
can be summarized as follows:
\begin{itemize}
   \item
      If $z_1,\dots,z_N$ satisfy the assumptions of the classical
      Gaussian linear model, then
      they satisfy the assumptions of the on-line Gaussian linear 
      model.  In other words, the 
      assumption that the errors $e_i$ in~(\ref{eq:classicalglm}) are independent
      and normal with mean zero and a common variance implies that conditional
      on $X_n^{\prime}Y_n = C$ and $Y_n^{\prime}Y_n = r^2$, the vector
      $Y_n$ is distributed uniformly over the surface of the sphere
      defined by $C$ and $r^2$.  This was already noted by
      R.~A. Fisher in 1925 \cite{fisher:1925applications}.
   \item
      The assumption of the on-line Gaussian linear model, that conditional
      on $X_n^{\prime}Y_n = C$ and $Y_n^{\prime}Y_n = r^2$, 
      the vector $Y_n$ is distributed uniformly over the surface of the sphere
      defined by $C$ and $r^2$, is sufficient to guarantee that~(\ref{eq:definitiont})
      has the $t$-distribution with $n-p-1$ degrees of freedom
      \cite{dempster:1969,efron:1969}. 
   \item
      Suppose $z_1,z_2,\dots$ is an infinite sequence of random variables.
      Then $z_1,\dots,z_N$ satisfy the assumptions of the on-line Gaussian linear 
      model
      for every integer $N$ if and only if the joint distribution of $z_1,z_2,\dots$ 
      is a mixture of distributions given by the classical Gaussian linear 
      model, each model in the mixture possibly having a different 
      $\beta$ and a different variance for the $e_i$ \cite{lauritzen:1988}.
\end{itemize}

A natural nonconformity measure $A$ for the on-line Gaussian linear model
is given, for
$\sigma = (X,X^{\prime}Y,Y^{\prime}Y)$
and $z=(x,y)$, by
\begin{equation}\label{eq:glmnm}
   A(\sigma,z):=|y - \hat{y}|,
\end{equation}
where $\hat{y} = x(X^{\prime}X)^{-1}X^{\prime}Y$.  
\begin{proposition}\label{prop:final}
When~(\ref{eq:glmnm}) is used as the nonconformity measure, the
$1-\epsilon$ conformal prediction region for $y_n$ is~(\ref{eq:generalt}),
the interval given by the $t$-distribution in the classical theory.
\end{proposition}
\begin{proof}
When~(\ref{eq:glmnm}) is used as the nonconformity measure,
the test statistic $A(\sigma_{n-1},z_n)$ used in the conformal algorithm 
becomes $|y_n - \hat{y}_n|$.  The conformal algorithm considers the
distribution of this statistic under $R_n(\cdot \givn \sigma_n)$.
But when $\sigma_n$ is fixed and $t_n$ is given by~(\ref{eq:definitiont}),
$|t_n|$ is a monotonically increasing function of $|y_n - \hat{y}_n|$
(see pp.~202--203 of \cite{vovk/gammerman/shafer:2005} for details).
So the conformal prediction region is the interval of values of $y_n$
for which $|t_n|$ does not take its most extreme values. 
Since $t_n$ has the $t$-distribution with $n-p-1$
degrees of freedom under $R_n(\cdot \givn \sigma_n)$, 
this is the interval~(\ref{eq:generalt}).
\qedtext
\end{proof}

Together with Informal Proposition~\ref{inprop:second},
Proposition~\ref{prop:final} implies
that when we use~(\ref{eq:generalt}) for a large number of successive
values of $n$, $y_n$ will be in the interval $1-\epsilon$ of
the time.  In fact, because the probability of error each time is exactly $\epsilon$,
we can say simply that the errors are independent and for this reason the classical
law of large numbers applies.

In our example involving the prediction of petal width from sepal length,
the exchangeability and Gaussian linear 
models gave roughly comparable results
(see Table~\ref{ta:comparison2} in~\S\ref{subsubsec:width}).  This will often be the case.  
Each model makes an assumption, however, that the other does not make.
The exchangeability model assumes that the $x$s, as well as the $y$s, 
are exchangeable.  The Gaussian linear model assumes that given the $x$s, 
the $y$s are normally distributed.

\addcontentsline{toc}{section}{Acknowledgements}

\section*{Acknowledgements}

This article is based on a tutorial lecture by Glenn Shafer at the Workshop
on Information Theory and Applications at the University of 
California at San Diego on February 1, 2007.  Glenn Shafer
would like to thank Yoav Freund and Alon Orlitsky for the
invitation to participate in this workshop.  We thank Wei Wu for
his assistance in computations.  Many other intellectual
debts are acknowledged in the preface of \cite{vovk/gammerman/shafer:2005}.

\addcontentsline{toc}{section}{References}

\bibliographystyle{plain}

\begin{thebibliography}{10}
\bibitem{asarin:1987}
Eugene~A. Asarin.
\newblock Some properties of {K}olmogorov $\delta$-random finite sequences.
\newblock {\em Theory of Probability and its Applications}, 32:507--508, 1987.

\bibitem{asarin:1988}
Eugene~A. Asarin.
\newblock On some properties of finite objects random in the algorithmic sense.
\newblock {\em Soviet Mathematics Doklady}, 36:109--112, 1988.

\bibitem{bickel/doksum:2001}
Peter~J. Bickel and Kjell~A. Doksum.
\newblock {\em Mathematical Statistics: Basic Ideas and Selected Topics. Volume
  I}.
\newblock Prentice Hall, Upper Saddle River, New Jersey, second edition, 2001.

\bibitem{cowell/etal:1999}
Robert~G. Cowell, A.~Philip Dawid, Steffen~L. Lauritzen, and David~J.
  Spiegelhalter.
\newblock {\em Probabilistic Networks and Expert Systems}.
\newblock Springer, New York, 1999.

\bibitem{czuber:1914}
Emanuel Czuber.
\newblock {\em Wahrscheinlichkeitsrechnung und ihre {A}nwendung auf
  {F}ehlerausgleichung, {S}tatistik, und {L}ebensversicherung}, volume I.
  {W}ahrscheinlichkeitstheorie, {F}ehlerausgleichung, {K}ollektivmasslehre.
\newblock Teubner, Leipzig and Berlin, third edition, 2000.

\bibitem{dempster:1969}
A.~P. Dempster.
\newblock {\em Elements of Continuous Multivariate Analysis}.
\newblock Addison-Wesley, Reading, Massachusetts, 1969.

\bibitem{draper/smith:1998}
Norman~R. Draper and Harry Smith.
\newblock {\em Applied Regression Analysis}.
\newblock Wiley, New York, third edition, 1998.

\bibitem{efron:1969}
Bradley Efron.
\newblock Student's t-test under symmetry conditions.
\newblock {\em Journal of the American Statistical Association}, 64:1278--1302,
  1969.

\bibitem{fisher:1925applications}
Ronald~A. Fisher.
\newblock Applications of {S}tudent's distribution.
\newblock {\em Metron}, 5:90--104, 1925.

\bibitem{fisher:1935}
Ronald~A. Fisher.
\newblock The fiducial argument in statistical inference.
\newblock {\em Annals of Eugenics}, 6:391--398, 1935.

\bibitem{fisher:1936}
Ronald~A. Fisher.
\newblock The use of multiple measurements in taxonomic problems.
\newblock {\em Annals of Eugenics}, 7:179--188, 1936.

\bibitem{fisher:1956}
Ronald~A. Fisher.
\newblock {\em Statistical Methods and Scientific Inference}.
\newblock Hafner, New York, third edition, 1973.
\newblock First edition, 1956.

\bibitem{gammerman/vovk:2007}
Alex Gammerman and Vladimir Vovk.
\newblock Hedging predictions in machine learning.
\newblock {\em Computer Journal}, 50:151--163, 2007.

\bibitem{student:1908}
Student (William~S. Gossett).
\newblock On the probable error of a mean.
\newblock {\em Biometrika}, 6:1--25, 1908.

\bibitem{hewitt/savage:1955}
Edwin Hewitt and Leonard~J. Savage.
\newblock Symmetric measures on {C}artesian products.
\newblock {\em Transactions of the American Mathematical Society}, 80:470--501,
  1955.

\bibitem{lauritzen:1988}
Steffen~L. Lauritzen.
\newblock {\em Extremal Families and Systems of Sufficient Statistics},
  volume~49 of {\em Lecture Notes in Statistics}.
\newblock Springer, New York, 1988.

\bibitem{lehmann:1959}
Erich~L. Lehmann.
\newblock {\em Testing Statistical Hypotheses}.
\newblock Wiley, New York, second edition, 1986.
\newblock First edition, 1959.

\bibitem{martin-lof:1974}
Per Martin-L\"of.
\newblock Repetitive structures and the relation between canonical and
  microcanonical distributions in statistics and statistical mechanics.
\newblock In Ole~E. Barndorff-Nielsen, Preben Bl\ae{}sild, and Geert Schou,
  editors, {\em Proceedings of Conference on Foundational Questions in
  Statistical Inference}, pages 271--294, Aarhus, 1974.
\newblock Memoirs~1.

\bibitem{neyman:1937}
Jerzy Neyman.
\newblock Outline of a theory of statistical estimation based on the classical
  theory of probability.
\newblock {\em Philosophical Transactions of the Royal Society, Series A},
  237:333--380, 1937.

\bibitem{robinson:1975}
G.~K. Robinson.
\newblock Some counterexamples to the theory of confidence intervals.
\newblock {\em Biometrika}, 62:155--161, 1975.

\bibitem{ryan:1997}
Thomas~P. Ryan.
\newblock {\em Modern Regression Methods}.
\newblock Wiley, New York, 1997.

\bibitem{seber/lee:2003}
G.~A.~F. Seber and A.~J. Lee.
\newblock {\em Linear Regression Analysis}.
\newblock Wiley, Hoboken, NJ, second edition, 2003.

\bibitem{shafer:2007}
Glenn Shafer.
\newblock From {C}ournot's principle to market efficiency.
\newblock In Jean-Philippe Touffut, editor, {\em Augustin Cournot, Economic
  Models and Rationality}. Edward Elgar, Cheltenham, 2007.

\bibitem{shafer/vovk:2001}
Glenn Shafer and Vladimir Vovk.
\newblock {\em Probability and Finance: It's only a Game!}
\newblock Wiley, New York, 2001.

\bibitem{small/etal:2006}
Dylan~S.\ Small, Joseph~L.\ Gastwirth, Abba~M.\ Krieger, and Paul~R. Rosenbaum.
\newblock R-estimates vs.\ {GMM}: {A} theoretical case study of validity and
  efficiency.
\newblock {\em Statistical Science}, 21:363--375, 2006.

\bibitem{tukey:1986}
John~W. Tukey.
\newblock Sunset salvo.
\newblock {\em American Statistician}, 40:72--76, 1986.

\bibitem{vapnik:1998}
Vladimir~N. Vapnik.
\newblock {\em Statistical Learning Theory}.
\newblock Wiley, New York, 1998.

\bibitem{vovk/gammerman/shafer:2005}
Vladimir Vovk, Alex Gammerman, and Glenn Shafer.
\newblock {\em Algorithmic Learning in a Random World}.
\newblock Springer, New York, 2006.
\end{thebibliography}

\appendix

\section{Validity}\label{app:proof}

The main purpose of this appendix is to formalize and prove the following
informal proposition:
\setcounter{inproposition}{0}
\begin{inproposition} 
Suppose $N$ is large, and the variables $z_1,\dots,z_N$ are exchangeable.
Suppose $E_n$ is an $\epsilon$-rare $n$-event
for $n=1,\dots,N$.
Then the law of large numbers applies;
with very high probability, no more than approximately the fraction
$\epsilon$ of the events $E_1,\dots,E_N$ will
happen. 
\end{inproposition}
We used this informal proposition in~\S\ref{subsec:lln} to establish
the validity of conformal prediction in the exchangeability model.
As we promised then, we will discuss two different approaches
to formalizing it:  a classical approach and a game-theoretical approach.
The classical approach shows that the $E_n$ are mutually independent in the 
case where they are exactly $\epsilon$-rare and then 
appeals to the classical weak law of large numbers for independent events.
The game-theoretic approach appeals directly to the more flexible 
game-theoretic weak law of large numbers.

Our proofs will also establish the analogous Informal Proposition~\ref{inprop:second},
which we used to establish the validity of conformal prediction in on-line
compression models in general.

In~\S\ref{subsec:fisherind}, we return to R.~A. Fisher's prediction interval
for a normal random variable, which we discussed in~\S\ref{subsubsec:fisherregion}.
We show that this prediction interval's successive hits are independent, so that validity
follows from the usual law of large numbers.  
Fisher's prediction interval is a special case of conformal prediction 
for the Gaussian linear model, and so it is covered by the general result for 
on-line compression models.  But the proof in~\S\ref{subsubsec:fisherregion},
being self-contained and elementary and making no reference to conformal prediction,
may be especially informative for many readers.

\subsection{A classical argument for independence}\label{subsec:exchind}

Recall the definitions we gave in~\S\ref{subsec:lln} in the 
case where $z_1,\dots,z_N$ are exchangeable:  
An event $E$ is an \textit{$n$-event} if its happening or failing is 
determined by the value of $z_n$ and the value of the bag $\lbag z_1,\dots,z_{n-1} \rbag$,
and an $n$-event $E$ is $\epsilon$-\textit{rare} if 
$\Prob ( E \givn \lbag z_1,\dots,z_n \rbag  ) \leq \epsilon$.
Let us further say that $n$-event $E$ is \textit{exactly} $\epsilon$-\textit{rare} if 
\begin{equation}\label{eq:rareapp}
   \Prob ( E \givn \lbag z_1,\dots,z_n \rbag  ) = \epsilon.
\end{equation}
The conditional probability in this equation is a random variable,
depending on the random bag $\lbag z_1,\dots,z_n \rbag$, but the equation
says that it is not really random, for it is always equal to $\epsilon$.
Its expected value, the unconditional probability of $E$, 
is therefore also equal to $\epsilon$.

\begin{proposition}\label{prop:preciseind}
Suppose $E_n$ is an exactly $\epsilon$-rare $n$-event for $n=1,\dots,N$.
Then $E_1,\dots,E_N$ are mutually independent.
\end{proposition}
\begin{proof}
Consider~(\ref{eq:rareapp}) for $n=N-1$:
\begin{equation}\label{eq:penultapp}
           \Prob ( E_{N-1} \givn \lbag z_1,\dots,z_{N-1} \rbag  ) = \epsilon.
\end{equation}
Given $\lbag z_1,\dots,z_{N-1} \rbag$, knowledge of $z_N$ does not change
the probabilities for $z_{N-1}$ and $\lbag z_1,\dots,z_{N-2} \rbag$,
and $z_{N-1}$ and $\lbag z_1,\dots,z_{N-2} \rbag$
determine the $(N-1)$-event $E_{N-1}$.
So adding knowledge of $z_N$ will not change the probability in~(\ref{eq:penultapp}):
$$
       \Prob ( E_{N-1} \givn \lbag z_1,\dots,z_{N-1} \rbag \; \& \;z_N ) = \epsilon.
$$
Because $E_N$ is determined by $z_N$ once $\lbag z_1,\dots,z_{N-1} \rbag$ is given,
it follows that 
$$
           \Prob ( E_{N-1} \givn \lbag z_1,\dots,z_{N-1} \rbag \; \& \; E_N ) = \epsilon,
$$
and from this it follows that 
$
           \Prob ( E_{N-1} \givn E_N ) = \epsilon.
$
The unconditional probability of $E_{N-1}$ is also $\epsilon$.
So $E_N$ and $E_{N-1}$ are independent.  Continuing the reasoning backwards to $E_1$,
we find that the $E_n$ are all mutually independent.
\qedtext
\end{proof}
This proof generalizes immediately to the general case of on-line compression 
models (see p.~\pageref{p:compressrare}); 
we simply replace $\lbag z_1,\dots,z_n \rbag$ with $\sigma_n$.

If $N$ is sufficiently large, and $E_n$ is an exactly $\epsilon$-rare $n$-event 
for $n=1,\dots,N$, then the law of large numbers applies;
with very high probability, no more than approximately the fraction
$\epsilon$ of the $N$ events will
happen.  It is intuitively clear that this conclusion will also hold if we have
an inequality instead of an equality in~(\ref{eq:rareapp}), 
because making the $E_n$ even less likely to happen
cannot reverse the conclusion that few of them will happen.

The preceding argument is less than rigorous on two counts.  First, the proof
of~Proposition~\ref{prop:preciseind} does not consider
the existence of the conditional probabilities it uses.  Second, 
the argument from the case where~(\ref{eq:rareapp}) is an equality
to that where it is merely an inequality, though entirely convincing,
is only intuitive.   A fully rigorous proof, which uses Doob's measure-theoretic
framework to deal with the conditional probabilities and uses a randomization
to bring the inequality up to an equality, is provided on pp.~211--213
of \cite{vovk/gammerman/shafer:2005}.

\subsection{A game-theoretic law of large numbers}\label{subsec:gameind}

As we explained in~\S\ref{subsec:betexc}, the game-theoretic interpretation
of exchangeability involves a backward-looking protocol, in which Bill 
observes first the bag $\lbag z_1,\dots,z_N \rbag$ and then successively
$z_N$, $z_{N-1}$, and so on, finally observing $z_1$.  Just before he observes
$z_n$, he knows the bag $\lbag z_1,\dots,z_n \rbag$ and 
can bet on the value of $z_n$ 
at odds corresponding to the probabilities the bag determines:
\begin{equation}\label{eq:onestepback1}
  \Prob \left(z_n = a \givn  \lbag z_1,\dots,z_n \rbag = B \right) 
      = 
       \frac{k}{n},  
\end{equation}
where $k$ is the number of times $a$ occurs in $B$.

\bigskip

\noindent
\textsc{The Backward-Looking Betting Protocol}\nopagebreak

\noindent
\textbf{Players:}  Joe, Bill

\parshape=7
\IndentI  \WidthI
\IndentI  \WidthI
\IndentI  \WidthI
\IndentII \WidthII
\IndentII \WidthII
\IndentII \WidthII
\IndentII \WidthII
\noindent
$\K_N := 1$.\\
Joe announces a bag $B_N$ of size $N$.\\
FOR $n=N,N-1,\dots,2,1$\\
  Bill bets on $z_n$ at odds set by~(\ref{eq:onestepback1}).\\
  Joe announces $z_n \in B_n$.\\
  $\K_{n-1} := \K_n + \text{Bill's net gain}$.\\
  $B_{n-1} := B_n \setminus \lbag z_n \rbag$.

\bigskip

\noindent
Bill's initial capital $\K_N$ is $1$.  His final capital is $\K_0$.

Given an event $E$, set 
$$
        e :=
\begin{cases}
    1 & \text{if $E$ happens}\\
    0 & \text{if $E$ fails}.
\end{cases}
$$
Given events $E_1,\dots,E_N$, set 
$$
   {\rm Freq}_N := \frac{1}{N} \sum_{j=1}^N e_j.
$$
This is the fraction of the events that happen---the frequency with which they happen.
Our game-theoretic law of large numbers will say
that if each $E_n$ is an $\epsilon$-rare $n$-event, then
it is very unlikely that ${\rm Freq}_N$ will substantially
exceed $\epsilon$.

In game-theoretic probability, what do we mean when we say 
an event $E$ is ``very unlikely''?
We mean that the bettor, Bill in this protocol, 
has a betting strategy that guarantees 
\begin{equation}\label{eq:cases}
  \K_0 \geq 
\begin{cases}
      C & \text{if $E$ happens}\\
      0 & \text{if $E$ fails},
\end{cases}
\end{equation}
where $C$ is a large positive number.  Cournot's principle, which says that 
Bill will not multiply his initial unit capital
by a large factor without risking bankruptcy, 
justifies our thinking $E$ unlikely.  The larger $C$, the more
unlikely $E$.  We call the quantity
\begin{equation}\label{eq:updef}
  \UP E := \inf \left\{ \frac{1}{C}  \biggm|  \text{ Bill can
                             guarantee~(\ref{eq:cases})} \right\}
\end{equation}
$E$'s \textit{upper probability}.
An unlikely event is one with small upper probability.

\begin{proposition}[Game-theoretic weak law of large numbers]\label{prop:gamellnprecise}
Suppose $E_n$ is an $\epsilon$-rare $n$-event,
for $n=1,\dots,N$.
Suppose $\epsilon<1/2$,
$\delta_1 > 0$, $\delta_2 > 0$,
and $N \geq 1/\delta_1 \delta_2^2$.  Then
$$
     \UP ( {\rm Freq}_N \geq   \epsilon  + \delta_2 ) \leq \delta_1.
$$
\end{proposition}
In words:  If $N$ is sufficiently large, there is a small (less than $\delta_1$)
upper probability that the frequency will exceed $\epsilon$ substantially
(by more than $\delta_2$).

Readers familiar with game-theoretic probability will recognize 
Proposition~\ref{prop:gamellnprecise} as a form of the game-theoretic 
weak law of large
numbers stated and proven on pp.~124--126 of \cite{shafer/vovk:2001}. 
The bound it gives for the upper probability of the event
${\rm Freq}_N \geq   \epsilon  + \delta_2$ is the same as 
the bound that Chebyshev's inequality gives
for the probability of this event in classical probability theory
when the $E_n$ are independent and all have probability $\epsilon$.

For the benefit of those not familiar with the concepts 
used on pp.~124--126 of \cite{shafer/vovk:2001} (after being introduced earlier in 
the book), we conclude with 
an elementary and self-contained proof of
Proposition~\ref{prop:gamellnprecise}.  

\begin{lemma}\label{lem:llngame}
Suppose, for $n=1,\dots,N$, that $E_n$ is an $\epsilon$-rare $n$-event.
Then Bill has a strategy that guarantees that 
his capital $\K_n$ will satisfy
\begin{equation}\label{eq:capineq}
    \K_n \geq
    \frac{n}{N} +
    \frac{1}{N}
    \left(\left(
      \sum_{j=n+1}^{N}
      (e_j-\epsilon)
    \right)^+\right)^2
\end{equation}
for $n=1,\dots,N$, where $t^+:=\max(t,0)$.
\end{lemma}
\begin{proof}
When $n=N$, (\ref{eq:capineq}) reduces to $\K_N \geq 1$, and this certainly holds;
Bill's initial capital $\K_N$ is equal to $1$.  So it suffices to show that if~(\ref{eq:capineq})
hold for $n$, then Bill can bet on $E_n$ in such a way that the corresponding
inequality for $n-1$, 
\begin{equation}\label{eq:capineq2}
    \K_{n-1} \geq
    \frac{n-1}{N} +
    \frac{1}{N}
    \left(\left(
      \sum_{j=n}^{N}
      (e_j-\epsilon)
    \right)^+\right)^2,
\end{equation}
also holds.  Here is how Bill bets.  
\begin{itemize}
  \item
If
  $
    \sum_{j=n+1}^N
    (e_j-\epsilon)
    \ge\epsilon
  $,
then Bill buys
$
    (2/N)
      \sum_{j=n+1}^N
      (e_j-\epsilon)
$
units of $e_n$.  By assumption, he pays no more than
$\epsilon$ for each unit.  So we have a lower bound on his 
net gain, $\K_{n-1} - \K_n$:
\begin{multline}\label{eq:forpos}
 \K_{n-1} - \K_n 
   \geq
    \frac{2}{N}
    \left(
    \sum_{j=n+1}^N
      (e_j-\epsilon)
    \right)
    (e_n-\epsilon)\\
  \ge
\frac{1}{N} 
    \left(
    \sum_{j=n}^N
      (e_j-\epsilon)
    \right)^2
    -
\frac{1}{N} 
    \left(
    \sum_{j=n+1}^N
      (e_j-\epsilon)
    \right)^2
    -\frac1N\\
=
\frac{1}{N} 
    \left(\left(
      \sum_{j=n}^N
      (e_j-\epsilon)
    \right)^+\right)^2
    -
\frac{1}{N} 
    \left(\left(
      \sum_{j=n+1}^N
      (e_j-\epsilon)
    \right)^+\right)^2
    -\frac1N .\\
\end{multline}
Adding~(\ref{eq:forpos}) and~(\ref{eq:capineq}), we obtain~(\ref{eq:capineq2}).
  \item
If
  $
    \sum_{j=n+1}^N
    (e_j-\epsilon)
    <
    \epsilon
  $,
then Bill does not bet at all, and $\K_{n-1}=\K_n$.  Because 
$$
    \left(\left(
      \sum_{j=n}^N
      (e_j-\epsilon)
    \right)^+\right)^2
  -
    \left(\left(
      \sum_{j=n+1}^N
      (e_j-\epsilon)
    \right)^+\right)^2
  \leq
    (\epsilon + (e_n - \epsilon))^2 \leq 1,
$$
we again obtain~(\ref{eq:capineq2}) from~(\ref{eq:capineq}).
\end{itemize}
  \qedtext
\end{proof}

\begin{proof} \hspace{-.1cm} \textbf{of Proposition~\ref{prop:gamellnprecise}}
The inequality ${\rm Freq}_N \geq   \epsilon  + \delta_2$ is equivalent to
\begin{equation}\label{eq:eqivevent}
       \frac{1}{N}
    \left(\left(
      \sum_{j=1}^{N}
      (e_j-\epsilon)
    \right)^+\right)^2
       \geq N\delta_2^2.
\end{equation}
Bill's strategy in Lemma~\ref{lem:llngame} does not risk bankruptcy
(it is obvious that $\K_n \geq 0$ for all $n$
when $\epsilon<1/2$),
and~(\ref{eq:capineq})
says
\begin{equation}\label{eq:capineq3}
    \K_0 \geq
    \frac{1}{N}
    \left(\left(
      \sum_{j=1}^{N}
      (e_j-\epsilon)
    \right)^+\right)^2.
\end{equation}
Combining~(\ref{eq:eqivevent}) and ~(\ref{eq:capineq3}) with the assumption 
that $N \geq 1/\delta_1\delta_2^2$, we see that when 
the event ${\rm Freq}_N \geq   \epsilon  + \delta_2$ happens,
$\K_0 \geq 1/\delta_1$.  So by~(\ref{eq:cases}) and~(\ref{eq:updef}),
$\UP ( {\rm Freq}_N \geq   \epsilon  + \delta_2 ) \leq \delta_1$.
  \qedtext
\end{proof}

\subsection{The independence of hits for Fisher's interval}\label{subsec:fisherind}

Recall that if $z_1,\dots,z_n,z_{n+1}$ are independent normal random variables
with mean $0$ and standard deviation $1$, the distribution of the ratio 
\begin{equation}\label{eq:normtratioapp}
       \frac{z_{n+1}}{\sqrt{\sum_{i=1}^n z_i^2/n}}
\end{equation}
is called the $t$-\textit{distribution with $n$ degrees of freedom}.
The upper percentile points for this distribution, the points
$
      t_n^{\epsilon}
$
exceeded by~(\ref{eq:normtratioapp}) with probability exactly $\epsilon$, are readily 
available from textbooks and standard computer programs.

Given a sequence of numbers $z_1,\dots,z_l$, where $l\ge 2$, we set
$$
         \overline{z}_l := \frac{1}{l} \sum_{i=1}^l z_i     
   \qquad \text{and}  \qquad  
        s_l^2 := \frac{1}{l-1} \sum_{i=1}^l (z_i - \overline{z}_l)^2.
$$
As we recalled in~\S\ref{subsubsec:fisherregion}, R.~A. Fisher proved 
that if $n\geq 3$ and $z_1,\dots,z_n$ are independent and normal 
with a common mean and standard deviation, then the ratio
$t_n$ given by 
\begin{equation}\label{eq:tratioapp}
     t_n :=  \frac{z_n - \overline{z}_{n-1}}{s_{n-1}} \sqrt{\frac{n-1}{n}}
\end{equation}
has the $t$-distribution with $n-2$ degrees of freedom \cite{fisher:1935}.
It follows that the event 
\begin{equation}\label{eq:fisherintervalapp}
  \overline{z}_{n-1} - t^{\epsilon/2}_{n-2} \; s_{n-1} \; \sqrt{\frac{n}{n-1}}
     \leq z_n \leq
  \overline{z}_{n-1} + t^{\epsilon/2}_{n-2} \; s_{n-1} \; \sqrt{\frac{n}{n-1}}
\end{equation}
has probability $1-\epsilon$.
We will now prove that the $t_n$ for successive $n$ are 
independent.  This implies that the events~(\ref{eq:fisherintervalapp})
for successive values of $n$ are independent, so that the law of large numbers applies:
with very high probability approximately $1-\epsilon$ of these events will happen.
This independence was overlooked by Fisher and subsequent authors.

We begin with two purely arithmetic lemmas, which do not rely on any assumption 
about the probability distribution of $z_1,\dots,z_n$.
\begin{lemma}\label{lem:invariant}
The ratio $t_n$ given by~(\ref{eq:tratioapp}) depends on $z_1,\dots,z_n$ only through
the ratios among themselves of the differences
\begin{equation}\label{eq:differences}
    z_1 - \overline{z}_n,\dots,z_n - \overline{z}_n.
\end{equation}
\end{lemma}
\begin{proof}
It is straightforward to verify that
\begin{equation}\label{eq:equivlence1}
   z_n - \overline{z}_{n-1} = \frac{n}{n-1}\left(z_n-\overline{z}_n\right)
\end{equation}
and
\begin{equation}\label{eq:equivlence2}
   s_{n-1}^2 = \frac{(n-1)s_n^2}{n-2} - \frac{n(z_n - \overline{z}_n)^2}{(n-1)(n-2)}.
\end{equation}
Substituting~(\ref{eq:equivlence1}) and~(\ref{eq:equivlence2}) in~(\ref{eq:tratioapp}) produces
\begin{equation}\label{eq:tmono1}
       t_n = 
   \frac{\sqrt{n(n-2)}(z_n - \overline{z}_n)}
        {\sqrt{(n-1)^2 s_n^2 - n (z_n - \overline{z}_n)^2}}
\end{equation}
or
\begin{equation}\label{eq:tmono}
       t_n = 
   \frac{\sqrt{n(n-2)}(z_n - \overline{z}_n)}
        {\sqrt{(n-1) \sum_{i=1}^n (z_i - \overline{z}_n)^2 - n (z_n - \overline{z}_n)^2}}.
\end{equation}
The value of~(\ref{eq:tmono}) is unaffected if all the 
$z_i - \overline{z}_n$ are multiplied by a nonzero constant.
\qedtext
\end{proof}

\begin{lemma}\label{lem:infosequence}
Suppose $\overline{z}_n$ and $s_n$ are known.  Then the following three
additional items of information are equivalent, inasmuch as the other
two can be calculated from any of the three:
\begin{enumerate}
   \item
      $z_n$
   \item
      $\overline{z}_{n-1}$ and $s_{n-1}$ 
   \item
      $t_n$
\end{enumerate} 
\end{lemma}
\begin{proof}
Given $z_n$, we can calculate $\overline{z}_{n-1}$ and $s_{n-1}$ 
from~(\ref{eq:equivlence1}) and~(\ref{eq:equivlence2}) 
and then calculate $t_n$ from~(\ref{eq:tratioapp}).
Given $\overline{z}_{n-1}$ and $s_{n-1}$, we can calculate $z_n$
from~(\ref{eq:equivlence1}) or~(\ref{eq:equivlence2}) 
and then $t_n$ from~(\ref{eq:tratioapp}).
Given $t_n$, we can invert~(\ref{eq:tmono1}) to find $z_n$
(when $\overline{z}_n$ and $s_n$ are fixed, this equation expresses
$t_n$ as a monotonically increasing function of $z_n$) and then 
calculate $\overline{z}_{n-1}$ and $s_{n-1}$ 
from~(\ref{eq:equivlence1}) and~(\ref{eq:equivlence2}). 
\qedtext
\end{proof}

Now we consider probability distributions for $z_1,\dots,z_n$.
\begin{lemma}\label{lem:uniform}
If $z_1,\dots,z_n$ are independent and
normal with a common mean and standard deviation, then 
conditional on $\overline{z}_n = w$ 
and $\sum_{i=1}^n (z_i - \overline{z}_n)^2 = r^2$, the vector
$(z_1,\dots,z_n)$ is distributed uniformly over the surface of
the $n$-dimensional
sphere of radius $r$ centered on the point $(w,\dots,w)$ in $\bbbr^n$.
\end{lemma}
\begin{proof}
The logarithm of the joint density of $z_1,\dots,z_n$ is
\begin{multline*}
  - \frac{n}{2}\log (2\pi \sigma^2)
-\frac{1}{2\sigma^2} \sum_{i=1}^n (z_i - \mu)^2 \\
   = 
  - \frac{n}{2}\log (2\pi \sigma^2)
-\frac{1}{2\sigma^2} \left( \sum_{i=1}^n (z_i - \overline{z}_n)^2 +n(\overline{z}_n - \mu)^2\right),
\end{multline*}
where $\mu$ and $\sigma$ are the mean and standard deviation, respectively.
Because this depends on $(z_1,\dots,z_n)$ only through 
$\overline{z}_n$ 
and $\sum_{i=1}^n (z_i - \overline{z}_n)^2$, the distribution of
$(z_1,\dots,z_n)$ conditional on $\overline{z}_n = w$ 
and $\sum_{i=1}^n (z_i - \overline{z}_n)^2 = r^2$ is
uniform over the set of vectors satisfying these conditions.
\qedtext
\end{proof}

\begin{lemma}\label{lem:uniformt}
If the vector $(z_1,\dots,z_n)$ is distributed uniformly 
over the surface of the $n$-dimensional sphere of radius $r$ around $(w,\dots,w)$ in 
$\bbbr^n$, then $t_n$
has the $t$-distribution with $n-2$ degrees of freedom.
\end{lemma}
\begin{proof}
The distribution of $t_n$ does not 
depend on $w$ or $r$.  This is because we can transform the uniform
distribution over one $n$-dimensional sphere into a uniform distribution over another
by adding a constant to all the $z_i$ and then multiplying the 
differences $z_i - \overline{z}_n$ by a constant, and 
by Lemma~\ref{lem:invariant}, this will not change $t_n$.

Now suppose $z_1,\dots,z_n$ are independent and
normal with a common mean and standard deviation.  
Lemma~\ref{lem:uniform} says that conditional on
$\overline{z}_n = w$ 
and $(n-1)s_n^2 =r^2$,
the vector $(z_1,\dots,z_n)$ is distributed uniformly 
over the surface of the sphere of radius $r$ centered on $w,\dots,w$.
Since the resulting distribution for $t_n$ does not depend on $w$
or $r$, it must be the same as the unconditional distribution.
\qedtext
\end{proof}

\begin{lemma}\label{lem:uniformcase}
Suppose $(z_1,\dots,z_n)$ is distributed uniformly over the surface of the
$N$-dimensional 
sphere of radius $r$ around $(w,\dots,w)$ in $\bbbr^N$.  
Then $t_3,\dots,t_N$ are mutually independent.
\end{lemma}
\begin{proof}
It suffices to show that $t_n$ still has the 
$t$-distribution with $n-2$ degrees of freedom conditional on $t_{n+1},\dots,t_N$.  
This will
imply that $t_n$ is independent of $t_{n+1},\dots,t_N$ and hence that 
all the $t_n$ are mutually independent.

We start knowing $\overline{z}_N$ and $s_N$.
So by Lemma~\ref{lem:infosequence}, learning $t_{n+1},\dots,t_N$ 
is the same as learning $z_{n+1},\dots,z_N$.  Geometrically, when we 
learn $z_N$ we intersect our $N$-dimensional 
sphere in $\bbbr^N$
with a hyperplane, reducing it to an $(N-1)$-dimensional
sphere in $\bbbr^{N-1}$.  (Imagine, for example, intersecting
a $3$-dimensional sphere with a plane: the result is a disc.)
When we learn $z_{N-1}$, we reduce the dimension again, and so on.
In each case, we obtain a uniform distribution on the surface 
of the lower-dimensional sphere for the remaining $z_i$.  
In the end, we find that $(z_1,\dots,z_n)$ is 
distributed uniformly over the surface of an
$n$-dimensional sphere in $\bbbr^n$, and 
so $t_n$ has the $t$-distribution with $n-2$ degrees of freedom 
by Lemma~\ref{lem:uniformt}.
\qedtext
\end{proof}

\begin{proposition}\label{prop:fishercase}
Suppose $z_1,\dots,z_N$ are independent and
normal with a common mean and standard deviation.
Then $t_3,\dots,t_N$ are mutually independent.
\end{proposition}
\begin{proof}
By Lemma~\ref{lem:uniformcase}, 
$t_3,\dots,t_N$ are mutually independent 
conditional on $\overline{z}_N = w$ and $s_N = r$, 
each $t_n$ having the $t$-distribution with $n-2$ degrees of freedom.
Because this joint distribution for $t_3,\dots,t_N$ does not depend
on $w$ or $r$, it is also their 
unconditional joint distribution.\qedtext
\end{proof}
\label{lastpage}
\ifWP
  \lastpage
\fi
\end{document}